\documentclass{article}
\PassOptionsToPackage{numbers, compress}{natbib}
\usepackage[final]{neurips_2021}

\usepackage[utf8]{inputenc} 
\usepackage[T1]{fontenc}    
\usepackage{hyperref}
\usepackage{url}            
\usepackage{booktabs}       
\usepackage{amsfonts}       
\usepackage{nicefrac}       
\usepackage{microtype}      
\usepackage{xcolor}         
\usepackage{amsmath}
\usepackage{bbm}
\usepackage{graphicx}
\usepackage{subcaption}
\usepackage{soul}
\usepackage{wrapfig}

\usepackage{algorithm}
\usepackage{algorithmic}

\usepackage{xcolor}

\definecolor{nyupurple}{RGB}{87, 6, 140}

\renewcommand{\vec}[1]{\mathbf{#1}}

\newcommand{\expval}{\mathop{\mathbb{E}}}
\newcommand{\argmax}{\mathop{\mathrm{argmax}}}

\title{Conditioning Sparse Variational Gaussian Processes for Online Decision-making}

\author{
  Wesley J. Maddox 
    \\
  New York University \\
  \texttt{wjm363@nyu.edu} \\
  \And
  Samuel Stanton \\
  New York University \\
  \texttt{ss13641@nyu.edu} \\
  \And
  Andrew Gordon Wilson \\
  New York University \\
  \texttt{andrewgw@cims.nyu.edu}
}

\begin{document}

\maketitle

\begin{abstract}
With a principled representation of uncertainty and closed form posterior updates, Gaussian processes (GPs) are a natural choice for online decision making. However, Gaussian processes typically require at least $\mathcal{O}(n^2)$ computations for $n$ training points, limiting their general applicability.
Stochastic variational Gaussian processes (SVGPs) can provide scalable inference for a dataset of fixed size, but are difficult to efficiently condition on new data. 
We propose online variational conditioning (OVC), a procedure for efficiently conditioning SVGPs in an online setting
that does not require 
re-training through the evidence lower bound with the addition of new data. 
OVC enables the pairing of SVGPs with advanced look-ahead acquisition functions for black-box optimization, even with non-Gaussian likelihoods. 
We show OVC provides compelling performance in a range of applications including 
active learning of malaria incidence,
and reinforcement learning on MuJoCo simulated robotic control tasks.
\end{abstract}

\section{Introduction}
Intelligent systems should be able to quickly and efficiently adapt to new data, adjusting their prior beliefs in response to the most recent events.
These characteristics are desirable whether the system in question is controlling the actuators of a robot, tuning the power output of a laser, or monitoring the changing preferences of users on an online platform. 
What these applications share in common is a constant stream of new information.
In this paper, we are interested in efficient \emph{conditioning}, meaning that we wish to efficiently update a posterior distribution after receiving new data.

The ability of Gaussian process (GP) regression models to condition on new data in closed form has made them a popular choice for Bayesian optimization (BO), active learning, and control \citep{frazier2018tutorial}.
All of these settings share similar characteristics: there is an ``outer loop'', where new data is acquired from the real world (e.g. an expensive simulator), interleaved with an ``inner loop'', which chooses where to collect data.
In BO, for example, the ``inner loop'' is the optimization of an acquisition function evaluated using a surrogate model of the true objective.
Simple acquisition functions, e.g. expected improvement (EI), consider only the current state of the surrogate, while more sophisticated acquisition functions ``look ahead'' to consider the effect of hypothetical observations on future surrogate states.
One such acquisition function, batch knowledge gradient (qKG), defines the one-step Bayes-optimal data batch as the batch that maximizes the expected surrogate maximum \emph{after} the batch has been acquired \citep{balandat_botorch_2020, wu_parallel_2016}.
Advanced acquisition functions like qKG require the surrogate to have both efficient posterior sampling and efficient conditioning on new data.

\begin{figure*}[h!]
	\begin{subfigure}{\textwidth}
		\centering
		\includegraphics[width=\textwidth,clip,clip,trim=0cm 0cm 0cm 14cm]{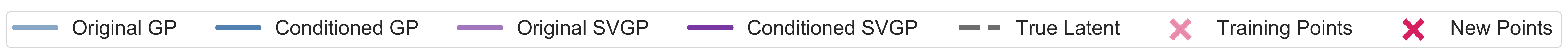}
	\end{subfigure}
\centering
\begin{subfigure}{0.32\textwidth}
\centering
\includegraphics[width=\linewidth]{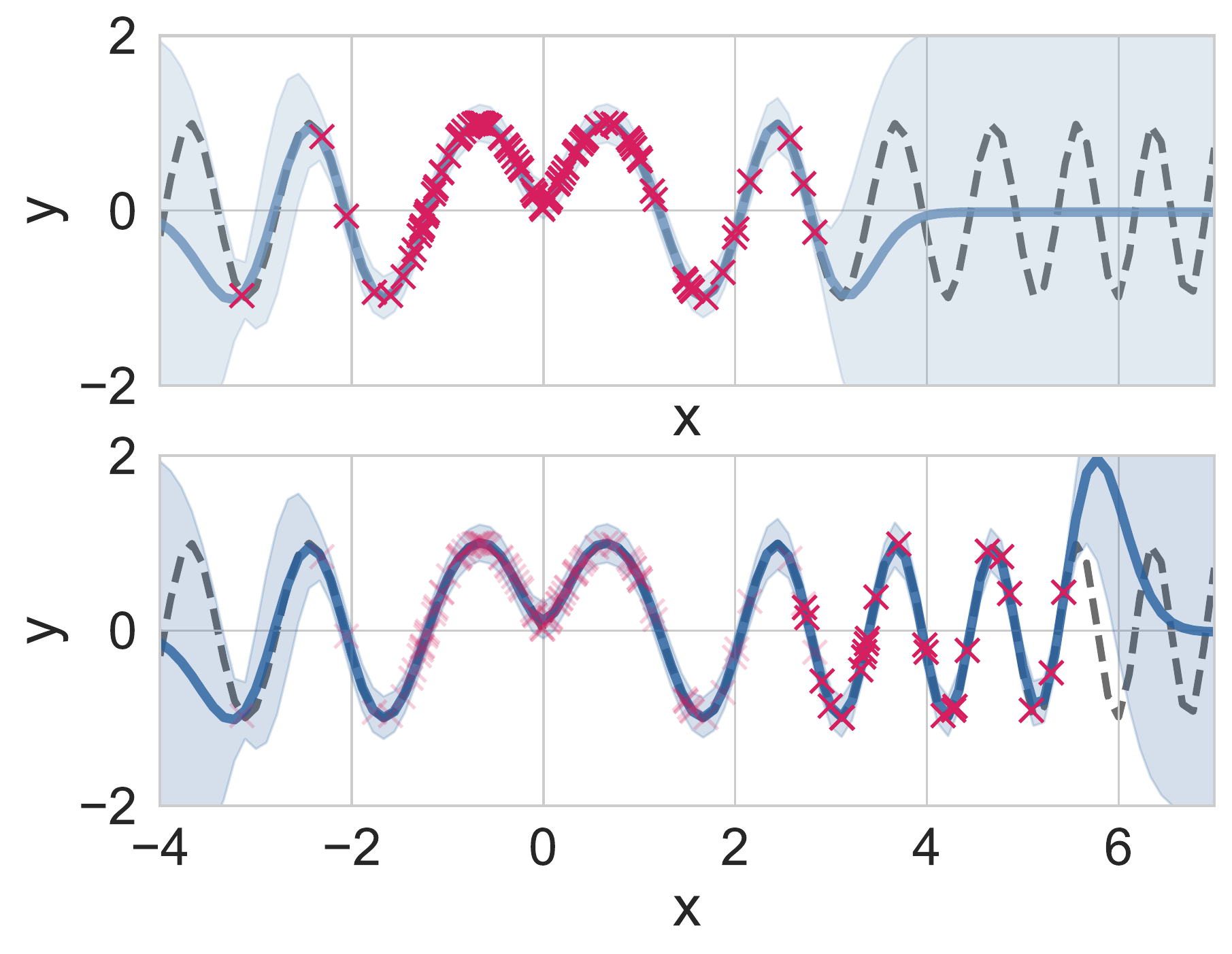}
\caption{Exact GP}
\label{fig:conditioning_exact_gaussian}
\end{subfigure}
\begin{subfigure}{0.33\textwidth}
\centering
\includegraphics[width=\linewidth]{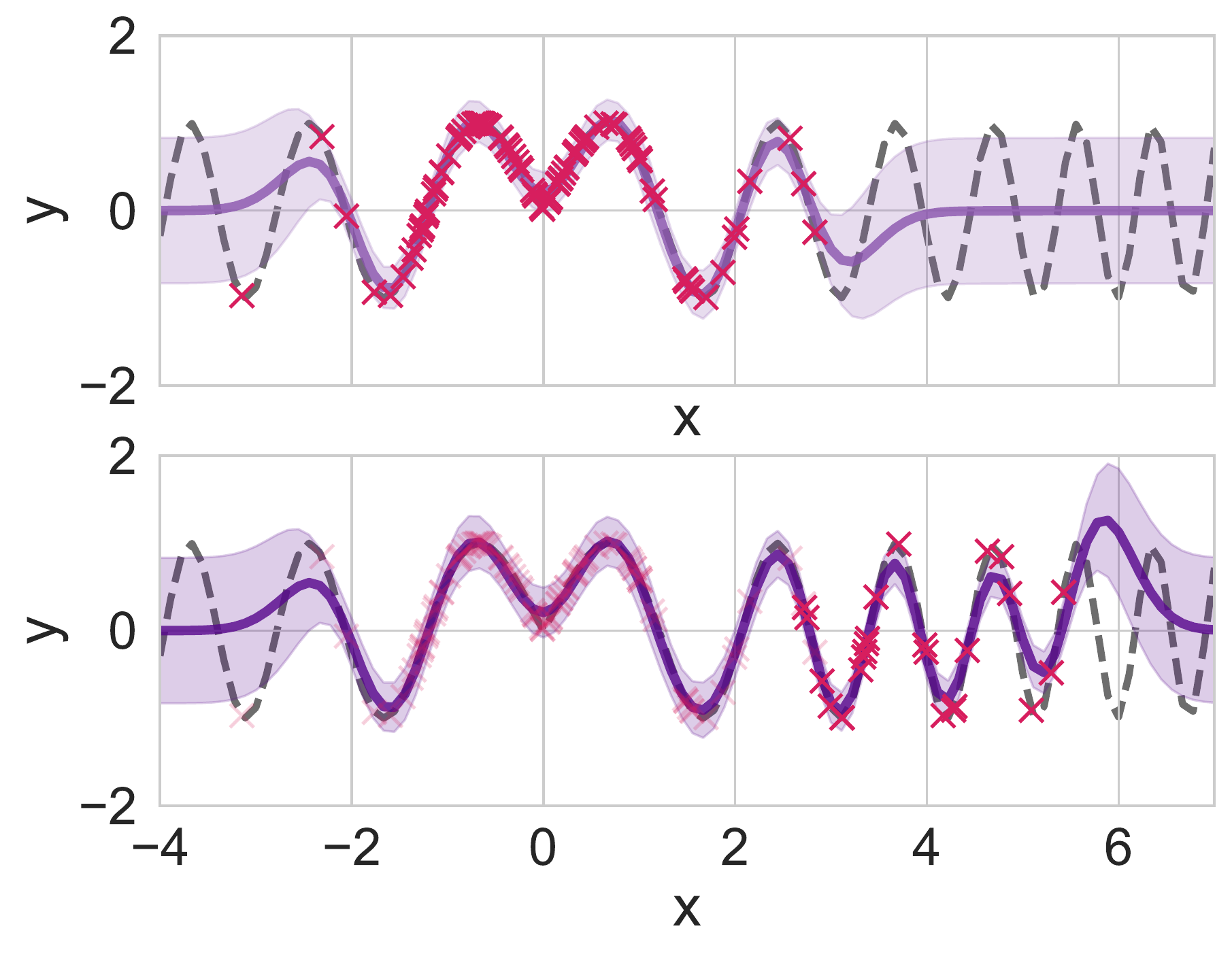}
\caption{SVGP + OVC}
\label{fig:conditioning_svgp_gaussian}
\end{subfigure}
\begin{subfigure}{0.33\textwidth}
\centering
\includegraphics[width=\linewidth]{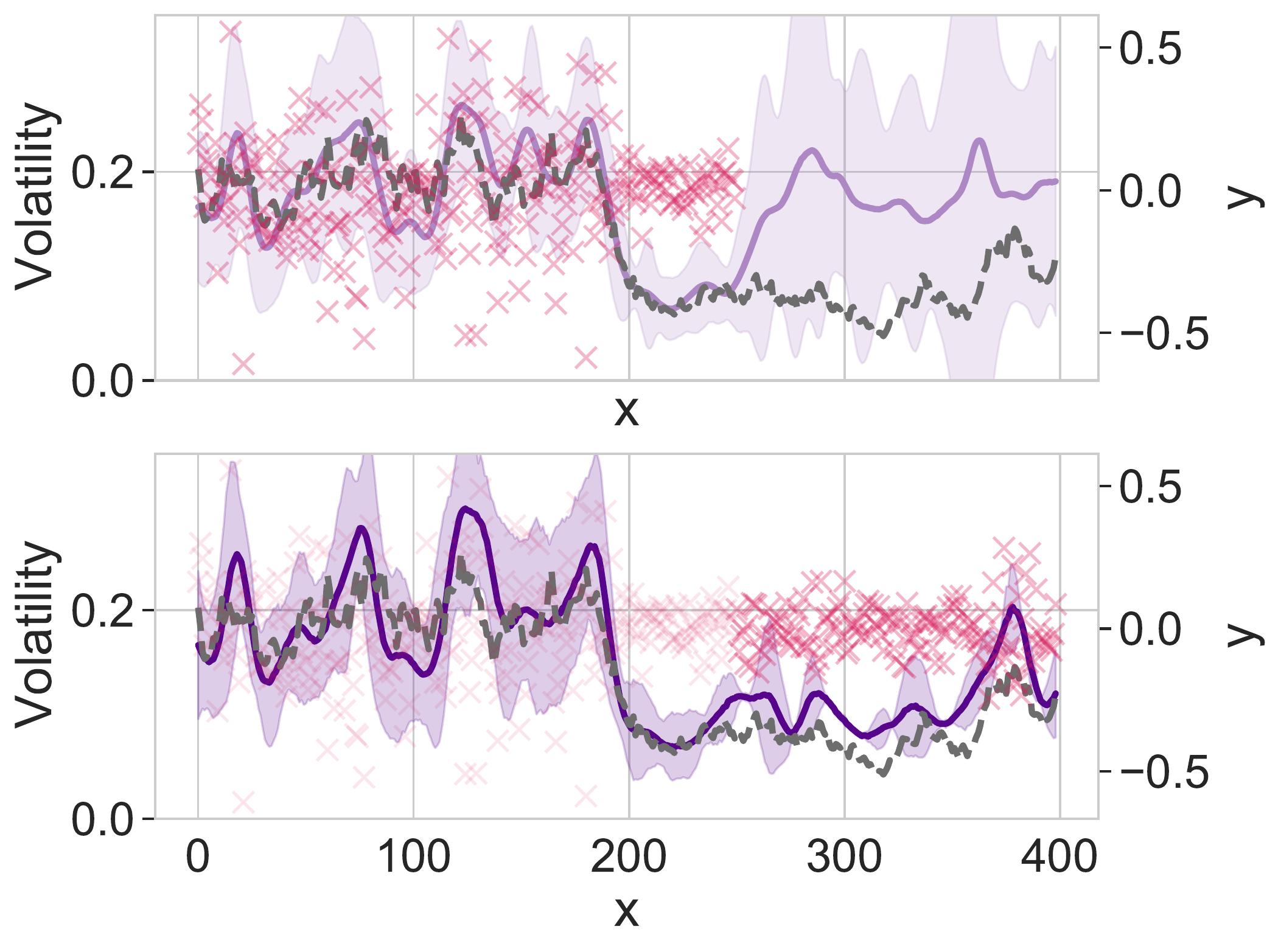}
\caption{SVGP + OVC volatility model}
\label{fig:conditioning_gpcv}
\end{subfigure}
\caption{
An exact GP updates its predictive distribution after conditioning on new data points \textbf{(a,} moving from top row to bottom row). With OVC, we can condition SVGPs on both Gaussian responses \textbf{(b)} and non-Gaussian models \textbf{(c)} such as the Gaussian copula volatility model \citep{wilson_copula_2010}.
}
\label{fig:svgp_conditioning}
\end{figure*}

GP regression has two major limitations that have prevented its large scale deployment for online decision-making.
First, the computational and memory consumption of exact GPs grows at least quadratically with the amount of data \citep{gardner_gpytorch:_2018, rasmussen_gaussian_2008}, generally limiting their usage to BO problems with fewer than $1,000$ function evaluations \citep{frazier2018tutorial, balandat_botorch_2020, wang2018batched}. 
Second, they are limited to applications that have continuous real-valued responses, enabling modelling with solely a Gaussian likelihood.
Stochastic variational Gaussian processes (SVGPs) \citep{hensman_bigdata_gp} have constant computational and memory footprints and are applicable to non-Gaussian likelihoods, but they sacrifice closed form expressions for updated posteriors on receiving new data.
The SVGP posterior is \emph{optimized} through the evidence lower bound (ELBO).
In the online setting, training with the ELBO has
two primary difficulties: the need to specify a fixed number of observations to properly scale the ELBO gradient \citep{broderick2013streaming} and the need to adjust the inducing points without looking at past data \citep{bauer2016understanding,bui_streaming_2017}.
Thus, we are presented with a choice between the simplicity and analytic tractability of exact GPs and the scalability and flexibility of SVGPs. 

In this work, we develop \textbf{O}nline \textbf{V}ariational \textbf{C}onditioning (OVC) to allow SVGPs to be conditioned on-the-fly, as shown in Figure \ref{fig:svgp_conditioning}.
In the top row of each subplot, we fit the data points shown in red, shifting to another batch of data points in the bottom row.
We use an exact GP in Figure \ref{fig:conditioning_exact_gaussian}, with exact conditioning shown in the bottom panel.
The SVGP emulates the exact GP very well before seeing the new data and again after conditioning on the new data by using OVC (Figure \ref{fig:conditioning_svgp_gaussian}). 
In Figure \ref{fig:conditioning_gpcv}, we consider a non-Gaussian data model (a Gaussian copula volatility model \citep{wilson_copula_2010}), where we cannot use exact GPs; 
the SVGP is still able to update its posterior over the latent volatility in response to new data without ``forgetting'' old observations.
OVC is inspired by a new, simple rederivation of streaming sparse GPs (O-SGPR), originally proposed by \citet{bui_streaming_2017}. 
OVC makes SVGPs truly compelling models for online decision-making, augmenting their existing strengths with efficient, closed-form conditioning on new data points.
In short, our contributions are:
\begin{itemize}
    \item The development of OVC, a novel method to condition SVGPs on new data without re-optimizing the variational posterior through an evidence lower bound.
    \item OVC provides both stable inducing point initialization for SVGPs while enabling the inducing points and variational parameters to update in response to the new data.
    \item Enabling the effective application of SVGPs, through OVC, in look-ahead acquisitions in BO for black-box optimization, controlling dynamical systems, and active learning.
\end{itemize}

Please see Appendix \ref{app:limitations} for discussion of the limitations and broader impacts of our work. 
Our code is available at \url{https://github.com/wjmaddox/online_vargp}.

\section{Preliminaries and Related Work}
In this section, we first review exact inference with Gaussian processes before reviewing variational sparse Gaussian processes, introducing sparse Gaussian process regression (SGPR), sparse variational Gaussian processes (SVGP), and streaming sparse GPs (O-SGPR).
Please see Appendix \ref{app/sec:further_background} for further background and Appendix \ref{app:rel_work} specifically for further related work.

\subsection{Gaussian Processes and Exact Inference}\label{sec:gp_intro}

For a full introduction to GPs, see \citet{rasmussen_gaussian_2008}.
We begin by reviewing GP regression, with inputs $\vec x \in \mathcal{X}$ and responses $y \in \mathbb{R}$.
GP regression assumes $y \sim \mathcal{N}(f, \sigma^2)$, where $f \sim \mathcal{GP}(0, k_\theta(\vec x,\vec x'))$ is a GP characterized by the kernel function $k_\theta$ and corresponding hyperparameters $\theta$.
Given $X_{\mathrm{train}} = [\vec x_1, \dots, \vec x_n]^\top$ and $X_{\mathrm{test}} = [\vec x'_1, \dots, \vec x'_k]^\top$, we denote the corresponding function values as $\vec v = [f(\vec x_1), \dots, f(\vec x_n)]^\top$ and $\vec w = [f(\vec x'_1), \dots, f(\vec x'_k)]^\top$, respectively.
Note that $\vec v$ and $\vec w$ are random variables with prior covariance $\mathrm{Cov}[\vec v, \vec w] = K_{\vec v \vec w} := k_\theta(X_{\mathrm{train}}, X_{\mathrm{test}})$. 
A well-known identity for multivariate conditional Gaussians allows us to compute the GP predictive posterior $p(\vec w| X_{\mathrm{test}}, \mathcal{D}, \theta) = \mathcal{N}(\mu_{\vec w | \mathcal{D}}^*, \Sigma^*_{\vec w | \mathcal{D}})$ as follows:
\begin{align}
    \mu_{\vec w | \mathcal{D}}^* = K_{\vec w \vec v} (K_{\vec v \vec v} + \sigma^2 I)^{-1} \mathbf y, \hspace{4mm} \Sigma_{\vec w| \mathcal{D}}^* = K_{\vec w \vec w} - K_{\vec w \vec v} (K_{\vec v \vec v} + \Sigma_{\vec y} )^{-1}K_{\vec v \vec w},
\end{align} 
where $\mathcal{D} := (X_{\mathrm{train}}, \vec y)$ and $\Sigma_{\vec y} = \sigma^2 I$.

Na\"ively, computations with $K_{\vec v \vec v}$ cost $\mathcal{O}(n^2)$ space and $\mathcal{O}(n^3)$ operations. 
When repeatedly computing $p(\vec w| X_{\mathrm{test}}, \mathcal{D}, \theta)$ at different sets of query points $X_{\mathrm{test}}$, it is more efficient to cache the terms which depend only on the training data.\footnote{Cached expressions will be written in \textcolor{orange}{orange}} 
Specifically, we store $\textcolor{orange}{\vec a}:=(K_{\vec v \vec v}+\Sigma_{\vec y})^{-1} \vec y$ (the predictive mean cache) and $\textcolor{orange}{RR^\top}:=(K_{\vec v \vec v}+\Sigma_{\vec y})^{-1}$ (the predictive covariance cache), resulting in simplified forms for the predictive distribution, $\mu_{\vec w | \mathcal{D}}^* = K_{\vec w \vec v}\textcolor{orange}{\vec a}$ and $\Sigma_{\vec w | \mathcal{D}}^* = K_{\vec w \vec w} - K_{\vec w \vec v}\textcolor{orange}{RR^\top}K_{\vec v \vec w}$. 
Adding a new observation is equivalent to adding a single row and column to $K_{\vec v \vec v}$ and an entry to $\vec y$, which enables efficient low-rank updates to the predictive caches \citep{osborne2010bayesian,gardner_gpytorch:_2018,pleiss_constant-time_2018,jiang_efficient_2020}.\footnote{
 What we use as the SGPR covariance cache is slightly different from the implementation in the prediction strategy in GPyTorch, which stores $\textcolor{orange}{R} = K_{\vec v \vec u}K_{\vec u \vec u}^{-1/2}.$
 However, they reduce to the same strategy.
 }

\subsection{Variationally Sparse Gaussian Processes}
\label{subsec:sparse_gp_prelims}

Variational sparse GPs reduce the computational burden of GP inference through sparse approximations of the kernel matrix.
For further reference, see \citet{matthews2017scalable} and \citet{van2019sparse}. 
These methods define a variational distribution $\phi$ over the inducing point \textit{values} $\vec u =[f(\vec z_1), \dots, f(\vec z_p)]^\top$, defined at inducing point \textit{locations} $Z=[\vec z_1, \cdots, \vec z_p]^\top$, where $\vec z_i \in \mathcal{X}$.  
$\phi(\vec u)$ is parameterized as a Gaussian with variational mean and covariance $\vec m_{\vec u}$ and $S_{\vec u}$.
These methods assume the latent function values $f(\vec x), f(\vec x')$ are conditionally independent given $\vec u$ and $\vec x, \vec x' \notin Z,$ so as to cheaply approximate the predictive posterior $p(\vec w | X_{\mathrm{test}}, \mathcal{D}, \theta) \approx q(\vec w) = \mathcal{N}(\mu_{\vec w | \phi}^*, \Sigma_{\vec w | \phi}^*)$. Like exact GP regression, we can compute $q(\vec w)$ in closed form (given $\vec m_{\vec u}, S_{\vec u}$),
\begin{align}
    \mu_{\vec w | \phi}^* := K_{\vec w \vec u}K_{\vec u \vec u}^{-1} \vec m_{\vec u}, \hspace{4mm}
    \Sigma_{\vec w | \phi}^* := K_{\vec w \vec w} - K_{\vec w \vec u}K_{\vec u \vec u}^{-1}(K_{\vec u \vec u} - S_{\vec u})K_{\vec u \vec u}^{-1} K_{\vec u \vec w}. \label{eq:sgpr_predictive}
\end{align}
Similarly the predictive mean and covariance caches are given by $\textcolor{orange}{\vec a} = K_{\vec u \vec u}^{-1}\vec m_{\vec u}$ and $\textcolor{orange}{RR^\top} = K_{\vec u \vec u}^{-1}(K_{\vec u \vec u} - S_{\vec u})K_{\vec u \vec u}^{-1}$, reducing the complexity of inference from $\mathcal{O}(n^3)$ to $\mathcal{O}(np^2),$ which is a significant improvement if $p \ll n$. 

There are two common approaches to finding optimal variational parameters $\vec m_{\vec u}$ and $S_{\vec u}$.
In seminal work, \citet{titsias_variational_2009} proposed sparse GP regression (SGPR), which optimizes $\vec m_{\vec u}$ and $S_{\vec u}$ in closed form, resulting in a "collapsed" evidence lower bound\footnote{A lower bound of the true GP marginal log-likelihood} (ELBO) that only depends on $\theta$ and $Z$. 
The computational cost of each gradient update to the remaining model parameters is still linear in $n$, and like exact GP regression, SGPR requires a Gaussian likelihood.
Stochastic variational GPs (SVGPs) remedies both these limitations by using gradient-based optimization to learn $\vec m_{\vec u}$ and $S_{\vec u}$ alongside $Z$ and $\theta$ \citep{hensman_bigdata_gp,hensman15}. 
The SVGP objective is an ``uncollapsed'' ELBO which decomposes additively across the training examples, allowing gradients to be estimated from minibatches of data, 
reducing the complexity of each gradient update to $\mathcal{O}(b p^2 + p^3)$, where $b$ is the minibatch size.

We emphasize the distinction between constant-time minibatch gradients, and constant-time conditioning.
Given an SVGP already trained on some existing data, conditioning jointly on both the old and new data requires storing all the data and making multiple gradient updates to the variational parameters. 
As the size of the dataset grows, so does the number of gradient steps needed.
In contrast by constant-time conditioning we mean a procedure that takes a posterior conditioned on existing data and produces a new posterior conditioned jointly on the old and new data with a fixed amount of compute and memory, regardless of the number of past observations.

One example of constant-time conditioning is found in \citet{bui_streaming_2017}, who proposed streaming sparse GPs (which we call online SGPR, or O-SGPR, to distinguish from sparse spectrum GPs \citep{lazaro2010sparse}) for incremental learning.
We extend their work, providing an alternative, simpler derivation of their model that highlights the connection with SGPR \citep{titsias_variational_2009}.
Furthermore, our perspective enables us to construct a principled approach to updating inducing point locations as new data arrives, that prevents the ``forgetting'' of old data induced by the resampling heuristic used by \citet{bui_streaming_2017}.

\subsection{Bayesian Optimization and Monte Carlo Acquisitions}
Bayesian optimization (BO) obtains $\vec x^* = \mathrm{argmin}_{\vec x \in \mathcal{X}} f(\vec x)$ by constructing a probabilistic \emph{surrogate model} of $f$, which in turn is used to evaluate an acquisition function. 
GPs are favored for BO due to their sample-efficiency and efficient posterior sampling that enables cheap, gradient based optimization of the acquisition function to propose new query points \citep{frazier2018tutorial}.
Many interesting acquisition functions look ahead into the future to see how the model will change if we query a specific point, a procedure known as ``fantasization'' \citep{hennig2012entropy,wu_parallel_2016,jiang_efficient_2020}.
Fantasization is done by drawing samples from the current surrogate posterior at some set of points and conditioning the surrogate on those samples.
For example, the batch knowledge gradient \citep[qKG, ][]{wu_parallel_2016,balandat_botorch_2020} is given by 
\begin{align}
    a(\vec x, \mathcal{D}) := \expval\limits_{f(\vec x) \sim p(\cdot | \mathcal{D})}\left(\max_{\vec x' \in \mathcal{X}} \expval\limits_{f(\vec x') \sim p(\cdot | \mathcal{D}_{+\vec x})} f(\vec x')\right) - \max_{\vec x' \in \mathcal{X}} \expval\limits_{f(\vec x') \sim p(\cdot | \mathcal{D})} f(\vec x')
    \label{app:batch_kg},
\end{align}
where $\mathcal{D}_{+\vec x} := \mathcal{D} \cup \{(\vec x, f(\vec x)\}$.
The inner expectation in the first term requires conditioning the surrogate model on posterior samples at $\vec x$, before optimizing through predictions of the conditioned surrogate model.
The goal is to simulate the effect on the model if we had observed the batch of data.

\paragraph{Use of sparse GPs in BO:} Sparse GPs have not seen wide adoption in the BO community, with only several preliminary studies that have mostly used basic acquisitions.
\citet{nickson2014automated} and \citet{krityakierne2015global} used expected improvement (EI) with SGPR on several test problems, while \citet{mcintire2016sparse} proposed a sparse GP method using EI to tune free electron lasers \citep{duris2020bayesian}.  \citet{stanton_kernel_2021} proposed WISKI, an online implementation of a scalable kernel approach called SKI \citep{wilson_kernel_2015}, for low-dimensional BO problems using batch upper confidence bound (qUCB) \citep{balandat_botorch_2020}.

\section{Methodology}\label{sec:methodology}

\begin{algorithm}[t!]
	\caption{\label{alg:model_conditioning} Online Variational Conditioning (OVC)}
	\begin{algorithmic}
		\STATE \textbf{Input:} Data batch $(X_{\mathrm{batch}}, \vec y)$, SVGP with inducing points $\textcolor{blue}{Z'}$ and $\phi(\vec u') = \mathcal{N}(\textcolor{blue}{\vec m_{\vec u'}}, \textcolor{blue}{S_{\vec u'}})$.
		\STATE 1. Compute $\textcolor{blue}{\vec c'}$, $\textcolor{blue}{C'}$ (Eq. \ref{eq:sgpr_var_dist}).
		\STATE 2. Compute $\textcolor{blue}{\hat{\vec y}} = \textcolor{blue}{K_{\vec u' \vec u'}' C'^{-1} \vec c'}$ and $\textcolor{blue}{\Sigma_{\hat{\vec y}}} = \textcolor{blue}{K_{\vec u' \vec u'}'^{-1}C' K_{\vec u' \vec u'}'^{-1}}$ (Eq. \ref{eq:pseudo_data}).
		\STATE 3. Construct GP with $\mathcal{D} = ([X_{\mathrm{batch}} \; \textcolor{blue}{Z'}], [\vec y \; \textcolor{blue}{\hat{\vec y}}])$ and $\Sigma = \text{blkdiag}(\Sigma_{\vec y}, \textcolor{blue}{\Sigma_{\hat{\vec y}}})$.
		\STATE 4. Compute predictive mean and covariance caches, $\textcolor{orange}{a}$ and $\textcolor{orange}{RR^\top}$ as in Section \ref{sec:gp_intro}.
		\STATE 5. Use caches to compute conditioned GP posterior on test points, $X_{\mathrm{test}}$.
	\end{algorithmic}
\end{algorithm}

We now briefly describe the key ideas behind OVC with the goal of devising an efficient and stable method for updating the variational parameters with respect to newly observed data.
We begin by highlighting an alternative parameterization of SGPR that will prove useful.
Then we describe the OVC update to the variational distribution from two equivalent points of view, namely the \textit{projection view} and the \textit{pseudo-data view}.
The pseudo-data view is summarized in Algorithm \ref{alg:model_conditioning}.
Next we address a critical detail for good performance, which is how the inducing point locations should be selected.
We then demonstrate how OVC can be applied to compute updated posterior distributions, e.g. $p(f | \mathcal{D}_{+\vec x})$ in Eq. \ref{app:batch_kg}, and quantities of the posterior, during gradient-based acquisition function optimization in BO, with reference to how this can be performed practically in Section \ref{app:building_block}.
Finally, we discuss how to apply OVC to models with non-Gaussian likelihoods.

\subsection{Updating the Variational Posterior}\label{sec:var_updates}

We assume that we have trained a SVGP model (e.g. with the ELBO) on a fixed set of data and have already trained the inducing point locations and variational parameters, $\vec m_{\vec u}, S_{\vec u}$. 
Instead of the traditional $\vec m_{\vec u}, S_{\vec u}$ parameterization used by \citet{titsias_variational_2009,hensman_bigdata_gp,hensman15}, we focus for now on an alternative parameterization which was favored in early work on sparse GP inference \citep{seeger2003fast,opper2009variational}.
The parameterization is also similar to those used in both dual space functional variational inference \citep{khan2017conjugate} and expectation propagation \citep{bui2017unifying}.
More recently, \citet{panos2018fully} used a similar parameterization in the context of large scale multi-label learning with SVGPs.

The SGPR predictive posterior $q(\vec w)$ relies on two terms dependent on the training data,
\begin{align}
    \vec c = K_{\vec u \vec v} \Sigma_{\vec y}^{-1} \vec y, \hspace{1cm}
    &C = K_{\vec u \vec v} \Sigma_{\vec y}^{-1} K_{\vec v \vec u}, \label{eq:sgpr_c_def}
\end{align}
where $\Sigma_{\vec y}$ is the covariance of the likelihood $p(\vec y | \vec f)$.
The optimal $\vec m_{\vec u}, S_{\vec u}$ are then given by
 \begin{align}
     \vec m_{\vec u} = K_{\vec u \vec u}(K_{\vec u \vec u} + C)^{-1}\vec c, \hspace{1cm}
    &S_{\vec u} = K_{\vec u \vec u}(K_{\vec u \vec u} + C)^{-1}K_{\vec u \vec u}, \label{eq:sgpr_var_dist}
\end{align}
which can be substituted into Eq. \eqref{eq:sgpr_predictive} to obtain $q(\vec w)$.
Our first observation is that if $\Sigma_{\vec y}$ is block-diagonal, then $\vec c$ and $C$ are additive across blocks of observations. 
For some intuition, consider i.i.d.  Gaussian noise (i.e. $\Sigma_{\vec y} = \sigma^2 I_n$), which implies 
\begin{align*}
    \vec c_i &= \sum_j  \sigma^{-2} y_j k_\theta(\vec z_i , \vec x_j) = \phi(\vec z_i)^\top \sum_j \sigma^{-2} y_j \phi(\vec x_j), \\ 
    C_{ik} &= \sum_j  \sigma^{-2} k_\theta(\vec z_i, \vec x_j) k(\vec x_j, \vec z_k)
    = \phi(\vec z_i)^\top \sum_j \sigma^{-2} \phi(\vec x_j) \phi(\vec x_j)^\top \phi(\vec z_k),
\end{align*}
where $\phi$ is the (potentially infinite-dimensional) feature map associated with $k_\theta$.
Hence the entries of $\vec c$ and $C$ are both inner products between projected inducing points and weighted sums of features.
For fixed inducing points, $Z,$ and hyper-parameters $\theta$, we can use these updates to produce a streaming version of SGPR by exploiting the additive structure of $c$ and $C$.
Furthermore, this streaming version of SGPR is exactly Gaussian conditioning for SGPR as we show in Appendix \ref{app:special_case}.
We can also allow the inducing points and hyper-parameters to vary, which we address next.\footnote{For full generality, we allow the hyper-parameters to vary; however, in our BO experiments, we only consider varying the inducing points as that's all we need to update when computing acquisition functions.}

\textbf{The projection view:}
We assume we have $\textcolor{blue}{\vec c' = K'_{\vec u' \vec v'}\Sigma_{\vec y'}^{-1}\vec y'}$ and $\textcolor{blue}{C'=K'_{\vec u' \vec v'}\Sigma_{\vec y'}^{-1}K'_{\vec v' \vec u'}}$, computed with inducing point locations $Z'$ from data $(X'_{\mathrm{batch}}, \vec y')$ with kernel hyperparameters $\theta'$ (using shorthand $k_{\theta'} = K'$).\footnote{Cached computations that depend on $(X'_{\mathrm{batch}}, \vec y')$ are highlighted in \textcolor{blue}{blue}.}
After obtaining the next parameters $Z$ and $\theta$ (perhaps from gradient based optimization of the ELBO), we observe new data $(X_{\mathrm{batch}}, \vec y)$ and would like to continue with inference.
One challenge is translating $\textcolor{blue}{\vec c'}, \textcolor{blue}{C'}$ (whose elements are inner products of the old features) to the new feature space associated with $\theta$.
To resolve this challenge, we construct a representative set of responses, $\hat {\vec y} = P^\top \textcolor{blue}{\vec c'}$ and likelihood covariance $\hat \Sigma_{\hat{ \vec y}} = P^\top \textcolor{blue}{C'} P$ to project from the old feature space into the new feature space by passing back through data space.
The choice that minimizes reconstruction error is 
the pseudo-inverse $P = (K'_{\vec v' \vec u'} K'_{\vec u' \vec v'})^{-1}K'_{\vec v' \vec u'}$, but requires storage of the full dataset, $(X'_{\mathrm{batch}}, \vec y')$.
Instead we take $P = \textcolor{blue}{K'^{-1}_{\vec u' \vec u'}}$, resulting in the following modifications to Eq. \eqref{eq:sgpr_c_def}:
\begin{align}
    \vec c &= K_{\vec u\vec{v}} \Sigma_{\vec{y}}^{-1} \vec y + K_{\vec u \vec u'}\textcolor{blue}{K_{\vec u' \vec u'}'^{-1} \vec c'}, \label{eq:streaming_c}\\
    C &= K_{\vec u \vec{v}} \Sigma_{\vec{y}}^{-1} K_{\vec{v} \vec u} + K_{\vec u \vec u'}\textcolor{blue}{(K_{\vec u' \vec u'}'^{-1}C' K_{\vec u' \vec u'}'^{-1})}K_{\vec u \vec u'}, \label{eq:streaming_C}
\end{align}
Note that $\textcolor{blue}{K_{\vec u' \vec u'}'^{-1} \vec c'} = \Sigma_{\vec y'}^{-1}\vec y'$ and $\textcolor{blue}{K_{\vec u' \vec u'}'^{-1}C' K_{\vec u' \vec u'}'^{-1}} = \Sigma_{\vec y'}^{-1}$ in the special case where $X'_{\mathrm{batch}} = Z'$.
We also want to emphasize that although we have only considered two batches of data for the sake of clarity, the approach applies to any number of incoming batches.

\textbf{The pseudo-data view:} The above update is equivalent to having an SGPR model with a Gaussian likelihood with covariance $\Sigma = \texttt{blkdiag}(\textcolor{blue}{\Sigma_{\hat {\vec y}}}, \Sigma_{\vec y})$ on the data $\{\texttt{cat}(Z', X_{\mathrm{batch}}), \texttt{cat}(\textcolor{blue}{\hat{\vec y}}, \vec y)\}$, where 
\begin{align}
\textcolor{blue}{\hat{\vec y} = K_{\vec u' \vec u'}' C'^{-1} \vec c'}, \hspace{4mm}
\textcolor{blue}{\Sigma_{\hat {\vec y}}^{-1} = K_{\vec u' \vec u'}'^{-1}C' K_{\vec u' \vec u'}'^{-1}}.
\label{eq:pseudo_data}
\end{align}
This interpretation is reminiscent of prior online variational approaches of \citet{csato2002sparse} and \citet{opper1998bayesian}.
That is, in the context of conditioning a SVGP, we can assume that we began with data $\{Z', \hat {\vec y}\}$ and are now observing the new data $\{X_{\text{batch}}, \mathbf{y}\}.$
See Appendix \ref{app:interp} for a more details.

\textbf{Extending to SVGPs:} 
SGPR computes $\vec m_{\vec u}$ and $S_{\vec u}$ as a function of $c$ and $C$ in Eq. \eqref{eq:sgpr_var_dist}.
However the equations can be reversed to solve for $c$ and $C$ given $\vec m_{\vec u}$ and $S_{\vec u}$, allowing us to condition \textit{any} variational sparse GP into an SGPR model, without touching any previous observations due to the conditional independence assumptions of variationally sparse GPs.\footnote{Alternatively, we could construct SVGPs via direct optimization of $c$ and $C.$} 
Note that if the variational parameters are not at the optimal solution when the variational distribution is projected back to the pseudo-data, the projection will be to the targets and likelihood \textit{for which the current variational parameters would be optimal}, which may not correspond well to the data that originally created the model. 
This potential pitfall is mitigated if the variational parameters are well optimized and is offset by the practical advantages of SVGPs.

\paragraph{Connection to O-SGPR \citep{bui_streaming_2017}: }

Formally, the updates described in Eqs. \eqref{eq:streaming_c} and \eqref{eq:streaming_C} are equivalent to the O-SGPR approach of \citet{bui_streaming_2017}, as we show in Appendix \ref{app:interp}.
The original derivation of O-SGPR is very technical, and does not highlight the similarities between the batch and online SGPR variants. 
Both the projection and pseudo-data views we have just described provide a much more intuitive way to reason about the behavior of O-SGPR models. 
Our formulation also eliminates a matrix subtraction operation, which is beneficial for numerical stability.

\subsection{Inducing Point Selection}\label{sec:ind_pt_selection}

Here, we describe inducing point selection during the conditioning procedure to enable better variance reduction on new inputs.
While heuristics including re-sampling \citep{bui_streaming_2017} and data sufficient statistics \citep{hoang2015unifying} have been proposed, they either require the number of inducing points to grow or gradually forget old observations. 
We show in Appendix \ref{app/subsec:o_sgpr_degrades} that relying exclusively on gradient-based optimization of inducing locations works very poorly in the online setting.

\begin{wrapfigure}{l}{0.35\textwidth}
\includegraphics[width=\linewidth]{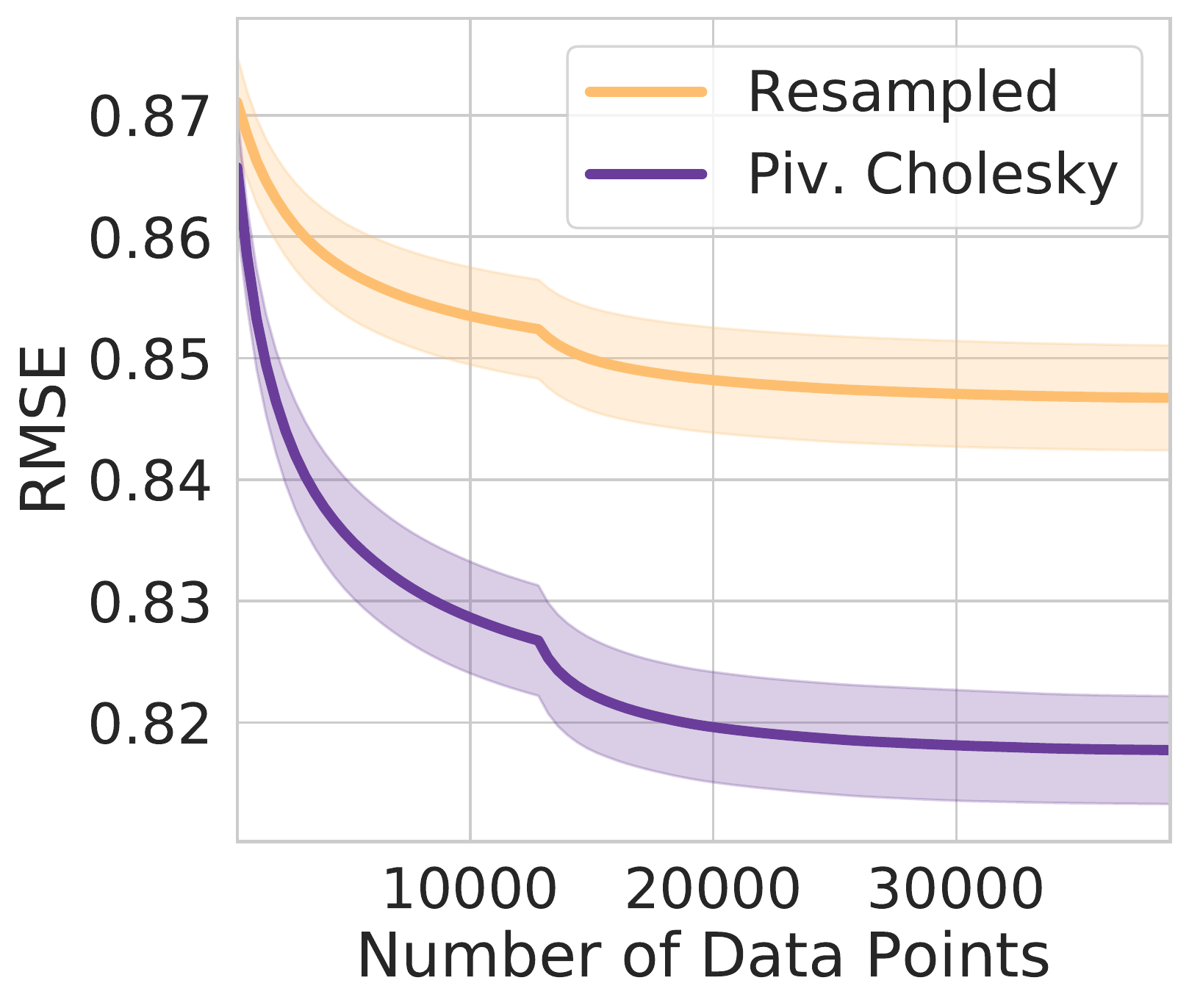}
\caption{Incremental learning RMSE on the UCI \textit{protein} dataset. Pivoted cholesky initialization outperforms resampling.}
\label{fig:inc_protein}
\vspace{-0.3in}
\end{wrapfigure}

To update the inducing point locations during conditioning, we extend \citet{burt_convergence_19}'s batch inducing point initialization approach to heteroskedastic Gaussian likelihoods.
They consider theoretical bounds on the marginal likelihood, finding that for homoscedastic Gaussian likelihoods a good strategy is to minimize the trace of the error of a rank-$p$ Nystr\"om approximation, e.g. $\varepsilon = tr\left(\Sigma_{\vec y}^{-1/2}(K_{\vec v \vec v} - K_{\vec v \vec u}K_{\vec u \vec u}^{-1}K_{\vec u \vec v}) \Sigma_{\vec y}^{-1/2}\right) = \sigma^{-1/2}tr(K_{\vec v \vec v} - Q_{\vec v \vec v})$ for $Q_{\vec v \vec v} = K_{\vec v \vec u}K_{\vec u \vec u}^{-1}K_{\vec u \vec v}.$ 
They follow a classical approach of \citet{fine2001efficient} by a greedy minimiziation strategy: choosing as inducing points the pivots of a rank $p$ pivoted Cholesky factorization of $K_{\vec v \vec v}.$

We denote the function values over the batch+pseudo dataset as $\hat{\vec v} = [f(\vec x_1), \dots, f(\vec x_b), f(\vec z'_1), \dots, f(\vec z'_p)]^\top$.
In our case the covariance of the pseudo-likelihood is no longer homoscedastic, so the slack term becomes $\varepsilon = \mathrm{tr}(\Sigma^{-1/2}(K_{\vec{\hat v} \vec{\hat v}} - Q_{\vec{\hat v} \vec{\hat v}}) \Sigma^{-1/2})$ and hence the pivoted Cholesky decomposition is instead performed over $\Sigma^{-1/2} K_{\vec{\hat v} \vec{\hat v}} \Sigma^{-1/2}$ to select the top $p$ pivots of the $p + n_{\text{new}}$ matrix. 
When compared to re-sampling the inducing points \citep{bui_streaming_2017}, pivoted cholesky updates perform significantly better, as shown in Figure \ref{fig:inc_protein} on the UCI protein dataset \citep{Dua:2019}.
Experimental details are given in Appendix \ref{app:data}.

\textbf{Application to Bayesian Optimization: }
In the context of BO, we condition on \emph{hypothetical} observations, and the conditioned surrogates are discarded after each acquisition function evaluation.
Since the SVGP will not be conditioned on more than a few batches of observations, we can sidestep the issue of updating inducing locations entirely by instead conditioning into an \textit{exact GP} trained the combined pseudo-data through the pseudo-likelihood.
That is, we model the data as $(\vec y, \textcolor{blue}{\vec{\hat y}}) \sim \mathcal{N}(f, \Sigma)$ (Gaussian with block-diagonal covariance) assuming $f \sim \mathcal{GP}(\mu_\theta, k_\theta(\cdot, \cdot)).$
We reach the same model by choosing $\hat X$ as the inducing points in our conditioned SGPR (Section \ref{sec:var_updates}).
For small $n_t,$ an exact GP is not much slower, taking only $(n_t + p)^3$ computations instead of $p^3$ computations, 
further reduced by using low rank updates.

\subsection{Local Laplace Approximations for Non-Gaussian Observations}
Thus far, we have solely considered Gaussian observations. The introduction of a non-Gaussian likelihood presents a new challenge, since it implies that the current observation batch and the pseudo-data are no longer jointly Gaussian. To adapt the conditioning procedure to the non-Gaussian setting, we can simply perform a Laplace approximation of the likelihood at the new points \citep[Ch. 3]{rasmussen_gaussian_2008}. Specifically, this gives us an approximate likelihood, $\hat p(y | f) = \mathcal{N}(\tilde y ; f, \mathcal{H}_*^{-1})$,
where $\mathcal{H}_* = \nabla^2_f \log p(y | f) |_{f^*(y)}$  and  $f^*(y)$ is the maximizer of $\log p(y | f) + f^\top K^{-1} f$, computed via Newton iteration.\footnote{One could consider using the posterior covariance instead of $K$. Our experiments with the posterior covariance produced more extreme values of $f$ and thus less regularization.} When conditioning on new observations $\vec y$, we substitute $f^*(\vec y)$ instead. That is, our new model has responses $(f^*(\vec y), \textcolor{blue}{\hat{\vec y}})$ instead of $(\vec y, \textcolor{blue}{\hat{\vec y}})$, and the pseudo-likelihood remains Gaussian with covariance $\Sigma = \mathrm{blkdiag}(\mathcal{H}_*^{-1}, \textcolor{blue}{\Sigma_{\hat{\vec y}}})$.
We primarily consider natural parameterizations of one-dimensional exponential families, so that $\mathcal{H}_*$ is positive, diagonal and depends solely on $f$.
Computing $\nabla \log p(y | f)$ and $\mathcal{H}_*$ is possible by hand but one can also use automatic differentiation \citep[AD, ][]{pearlmutter1994fast}.\footnote{Specifically, we use PyTorch's functional API, \url{https://pytorch.org/docs/stable/autograd.html\#functional-higher-level-api}.}
In Appendix \ref{app:exp_data}, Figure \ref{fig:bananas_rollout} we show the effect of repeated Laplace approximations across several batches for online classification.

\begin{figure*}[t!]
\centering
\begin{subfigure}{0.24\textwidth}
\centering
\includegraphics[width=\linewidth]{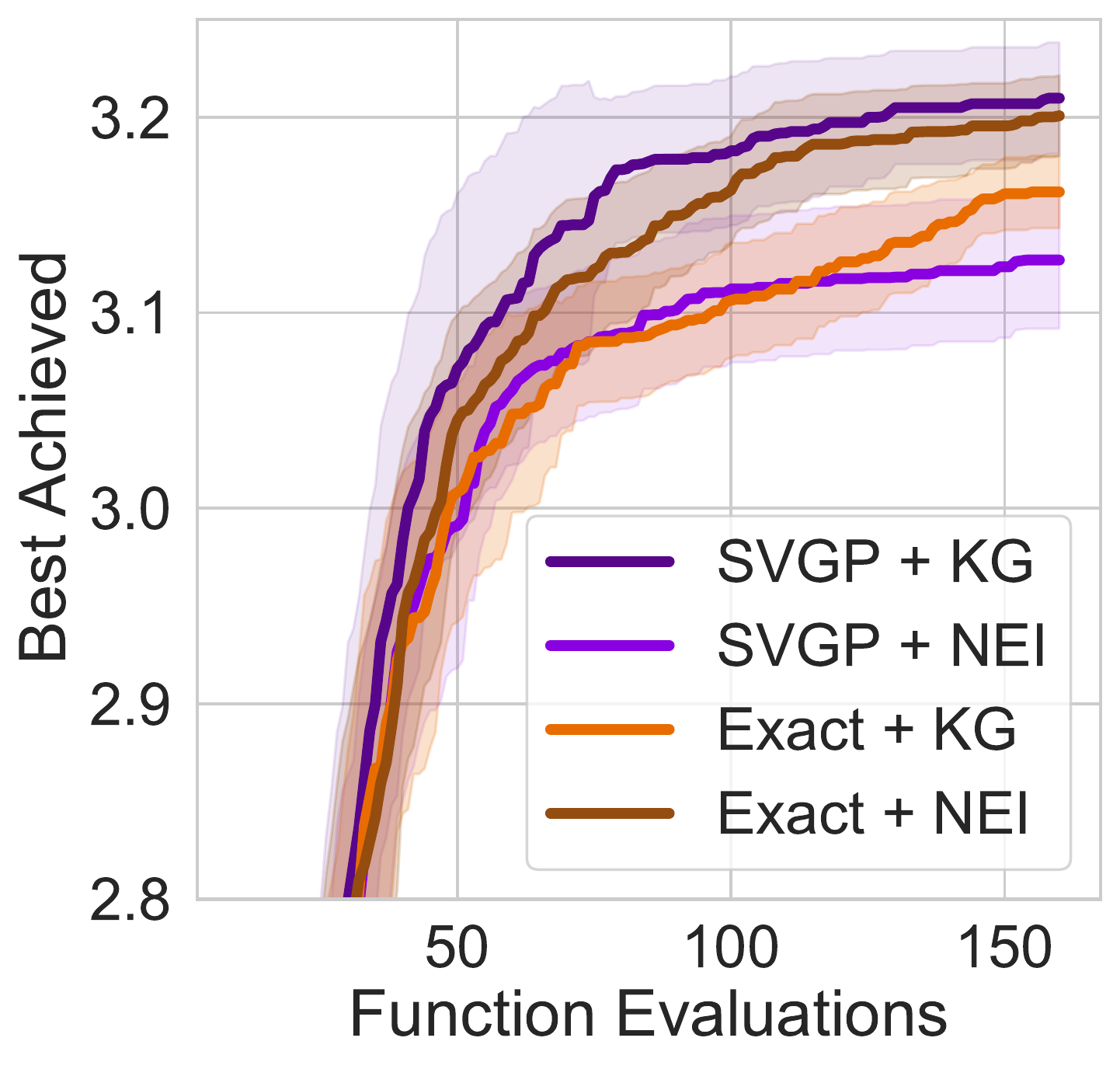}
\caption{Hartmann6}
\label{fig:kg_hartmann}
\end{subfigure}
\begin{subfigure}{0.24\textwidth}
\centering
\includegraphics[width=\linewidth]{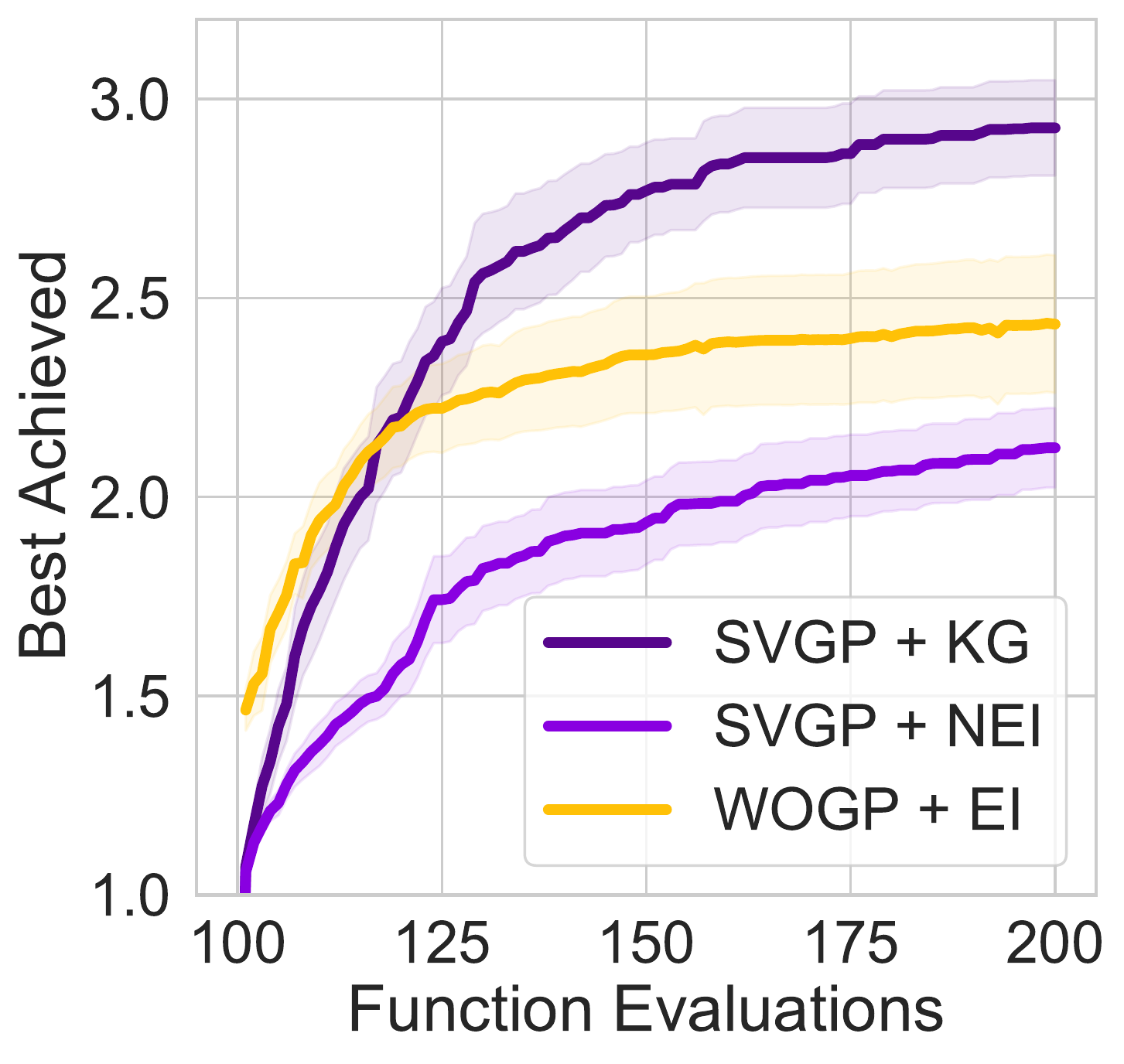}
\caption{Laser}
\label{fig:kg_laser}
\end{subfigure}
\begin{subfigure}{0.24\textwidth}
\centering
\includegraphics[width=\linewidth]{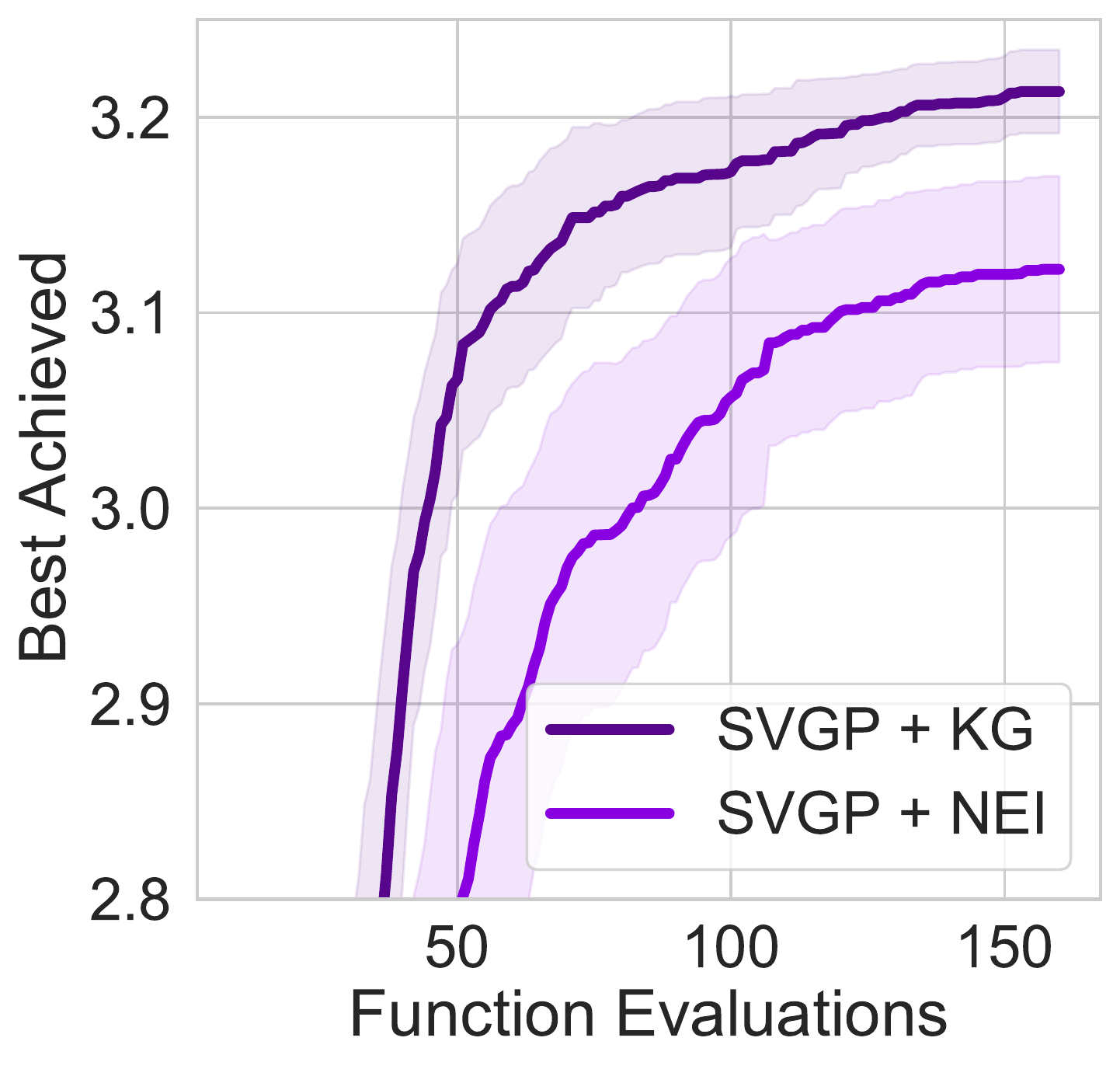}
\caption{Poisson-Hartmann6}
\label{fig:kg_poisson}
\end{subfigure}
\begin{subfigure}{0.24\textwidth}
\centering
\includegraphics[width=\linewidth]{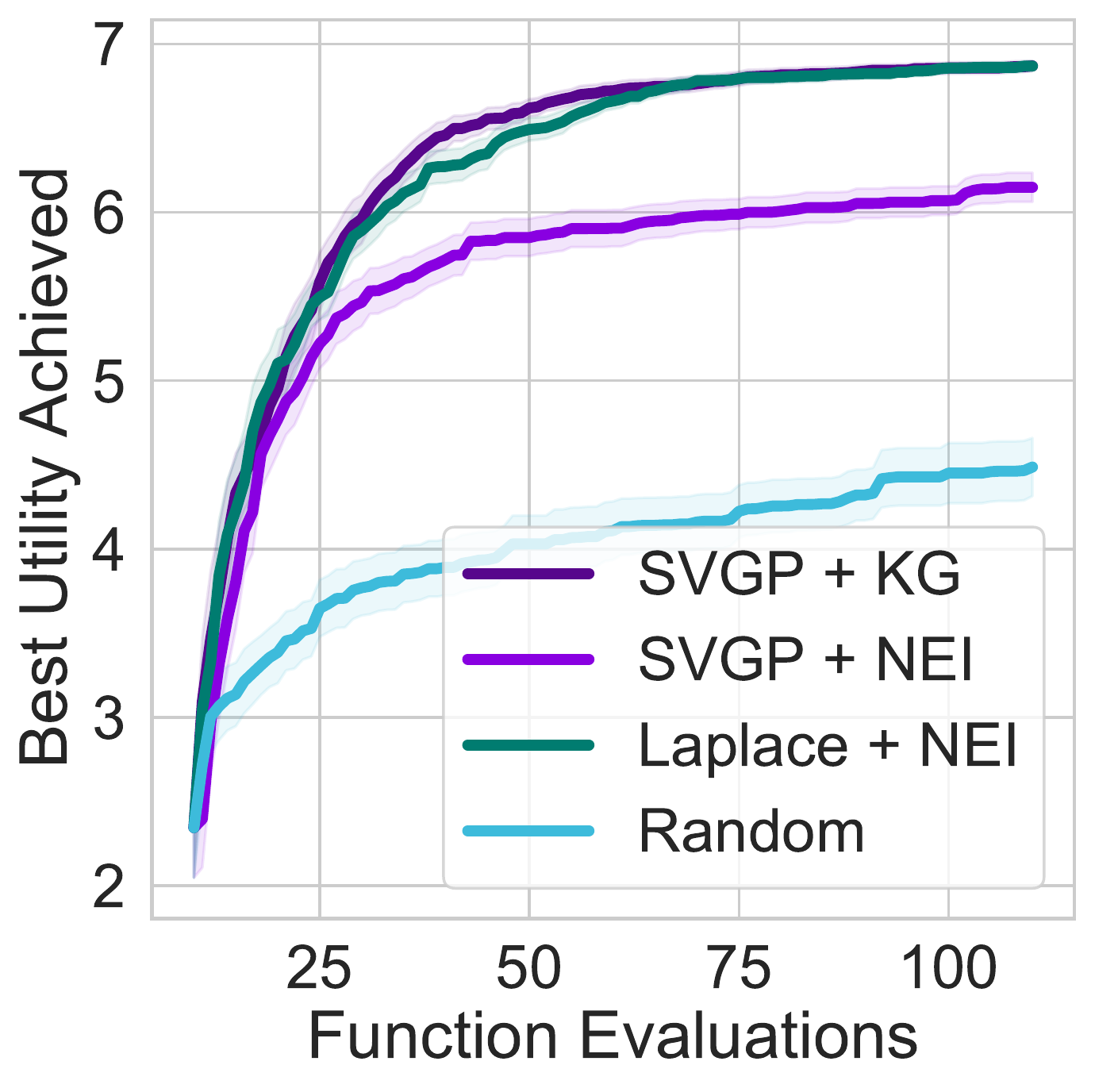}
\caption{Preference Learning}
\label{fig:kg_preferences}
\end{subfigure}
\caption{\textbf{(a)} Hartmann6 test problem with one constraint. Here, SVGPs with noisy expected improvement (qNEI) and KG match the performance of exact GPs with qNEI and qKG. \textbf{(b)} Free electron laser problem from \citet{mcintire2016sparse}; SVGPs with knowledge gradient outperforming weighted sparse GPs. \textbf{(c)} Constrained Hartmann6 test problem with count responses (Poisson likelihood). Only SVGPs can be used here, and qKG outperforms qNEI. \textbf{(d)} Preference learning; SVGPs with qKG are similar to Laplace approximations with NEI, and outperform qNEI with SVGPs.}
\label{fig:svgp_kg_bo}
\end{figure*}

\section{Experiments}
\label{sec:experiments}

Our experimental evaluation demonstrates that SVGPs using OVC can be successfully used as surrogate models 
with advanced acquisition functions in Bayesian optimization, even in the large batch and non-Gaussian settings. In keeping with the BO literature, we will refer to the query batch size as $q$ (not to be confused with the variational posterior $q(f)$ in previous sections).
\emph{All SVGP models that use conditioning (or fantasization) require OVC to even be practical to implement.}

\subsection*{Using OVC as a Building Block inside of BO}\label{app:building_block}

In all of our experiments, we use OVC as a \emph{building block} to enable fantasization (Algorithm \ref{alg:model_conditioning}) within a standard BO acquisition function that requires fantasiziation.
These acquisitions are generally ``look-ahead" as a result; specifically, qKG \citep{balandat_botorch_2020,jiang_efficient_2020}, LTSs, our version of qGIBBON which uses a fantasy batch \citep{moss2021gibbon}, and qMultiStepLookAhead \citep{jiang_efficient_2020} all use the fantasization model.
After adding in OVC as the \texttt{condition\_on\_observations} function within a BoTorch model class \citep{balandat_botorch_2020}, we can simply optimize qKG or qGIBBON with an SVGP exactly as an exact GP surrogate, by using gradient based optimizers such as L-BFGS-B.
In general, we need to differentiate through the fantasy model with respect to the inputs and then use gradient based methods to find the optimum.
Please see \citet{balandat_botorch_2020} and \citet{frazier2018tutorial} for description of how a BO loop is constructed and how acquisition functions are optimized.

\paragraph{Experimental Setup}
In general, a Bayesian optimization loop consists of the steps of training the model and then using the trained model to optimize an acquisition function to acquire new data points, which are then added into the training data for the next model.
All experiments use PyTorch \citep{paske2019pytorch}, GPyTorch \citep{gardner_gpytorch:_2018}, and BoTorch \citep{balandat_botorch_2020}. 
Unless otherwise specified, we run each experiment $50$ times and report the mean and two standard deviations of the mean.

In the first step, we first train the inducing points, variational distribution, and kernel hyper-parameters using the evidence lower bound given in Eq. \ref{eq:svgp_elbo}. As all components are differentiable, we use the Adam optimizer with a learning rate of $0.1$ and optimize for $1000$ steps or until the loss converges, whichever is shorter. To initialize the inducing points, we compute a pivoted cholesky factorization on the initial kernel on the training data (described in Section \ref{sec:ind_pt_selection} following \citet{burt_convergence_19}). The kernel hyper-parameters are initialized to GPyTorch defaults (which sets all lengthscales to one), while the variational distribution is initialized to $\vec m_{\vec u} = 0$, $S_{\vec u} = I$ (again, GPyTorch defaults). 
Further experimental details and dataset descriptions are in the Appendix.

\subsection{Knowledge Gradient with SVGPs}
These experiments use the one-shot formulation of the batch knowledge gradient (qKG) (Eq. \ref{app:batch_kg}) from \citet{balandat_botorch_2020}, who demonstrated that qKG outperforms other acquisitions due to being able to plan two steps into the future.
\emph{Using and optimizing qKG has only been available for exact GPs previously.}
By using OVC, we have enabled SVGPs to also efficiently and tractably optimize qKG, even for non-Gaussian observations.
We compare to batch noisy expected improvement \citep[qNEI,][]{letham2019constrained} which is myopic and does not use fantasization (e.g. conditioning).
Here, for the SVGPs we used $\min(N, 25)$ inducing points.

\textbf{Gaussian observations: }
We use the \textbf{Hartmann6} test function, with one black box constraint, maximizing $f(x) = -\sum_{i=1}^4\alpha_i\exp\{-\sum_{j=1}^6 A_{ij}(x_j - P_{ij})^2\}$ subject to the constraint that $c(x) = ||x||_1 \leq 3$ for fixed $A,P, \alpha$. 
We use $10$ initial points and a batch size of $3$ optimizing for $50$ iterations, comparing to SVGPs and exact GPs using qNEI. 
We show the results in Figure \ref{fig:kg_hartmann} where SVGPs with qKG match exact GPs with both qNEI and qKG, and outperform SVGPs using qNEI. 

Second, we mimic the \textbf{laser tuning} experiment of \citep{mcintire2016sparse,duris2020bayesian}, demonstrating that SVGPs outperform even weighted online GPs (WOGP), which were designed for this task.
Here, we use $100$ initial points, with $d=14,$ and and wish to tune a laser's output energy as a function of the magnet settings that produce the beam.
Like \citet{mcintire2016sparse} we treat a pretrained GP fit on experimental data as a simulator.
We use a batch size of $1,$ finding that SVGPs + KG outperform WOGP (Figure \ref{fig:kg_laser}).
However, exact GPs outperform the variational approaches (Appendix Fig. \ref{fig:app:kg_laser}).

\textbf{Non-Gaussian likelihoods: }
Next, we extend the knowledge gradient to problems with non-Gaussian likelihoods.
First, we take the constrained Hartmann6 test function from the previous section, and use \textbf{Poisson} responses, $y \sim \text{Poisson}(\exp\{f(x)\}),$ repeating the same settings as for the Gaussian case.
Now, the data is non-Gaussian and cannot be well-modelled by a Gaussian likelihood, so we compare to only SVGPs with qNEI.
qNEI is outperformed by qKG, as shown in Figure \ref{fig:kg_poisson}.

Second, in Figure \ref{fig:kg_preferences}, we consider a \textbf{preference learning} problem inspired by \citet{lin2020mdmaking}.
Here, the latent data is described by $f(x) = -10^{-1/2} \sum_{i=1}^{10} \sqrt{i} \cos(2\pi x_i)$ for $x \in [0,1]^{10},$ comparing to Laplace approximations \citep{chu2005preference}. 
Again, we see that SVGPs with qKG outperform qNEI with both SVGPs and Laplace approximations.

\subsection{Active Learning of Disease Incidence}
We next present results for two active learning tasks governing the collection of disease incidence data. 
In both tasks the acquisition functions again require efficient conditioning on hypothetical data, and
the second task has Binomial responses, so exact GPs cannot be applied.
In both settings, applying OVC to SVGPs gives strong results competitive with either exact GPs or random forests.

\begin{figure*}[t!]
\centering
\begin{subfigure}{0.32\textwidth}
\centering
\includegraphics[width=\linewidth]{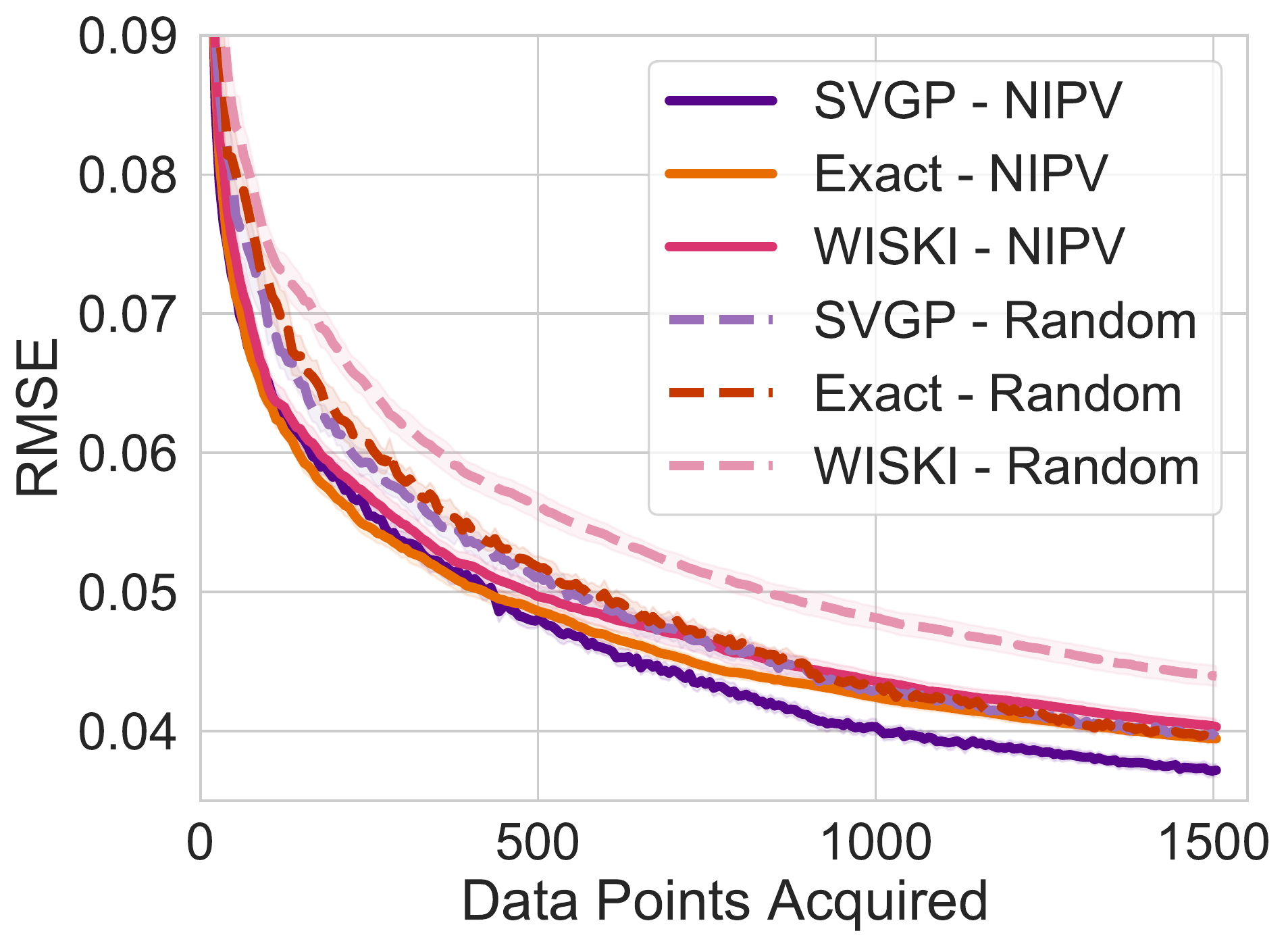}
\caption{Malaria incidence in Nigeria}
\label{fig:nipv_malaria}
\end{subfigure}
\begin{subfigure}{0.32\textwidth}
\centering
\includegraphics[width=\linewidth]{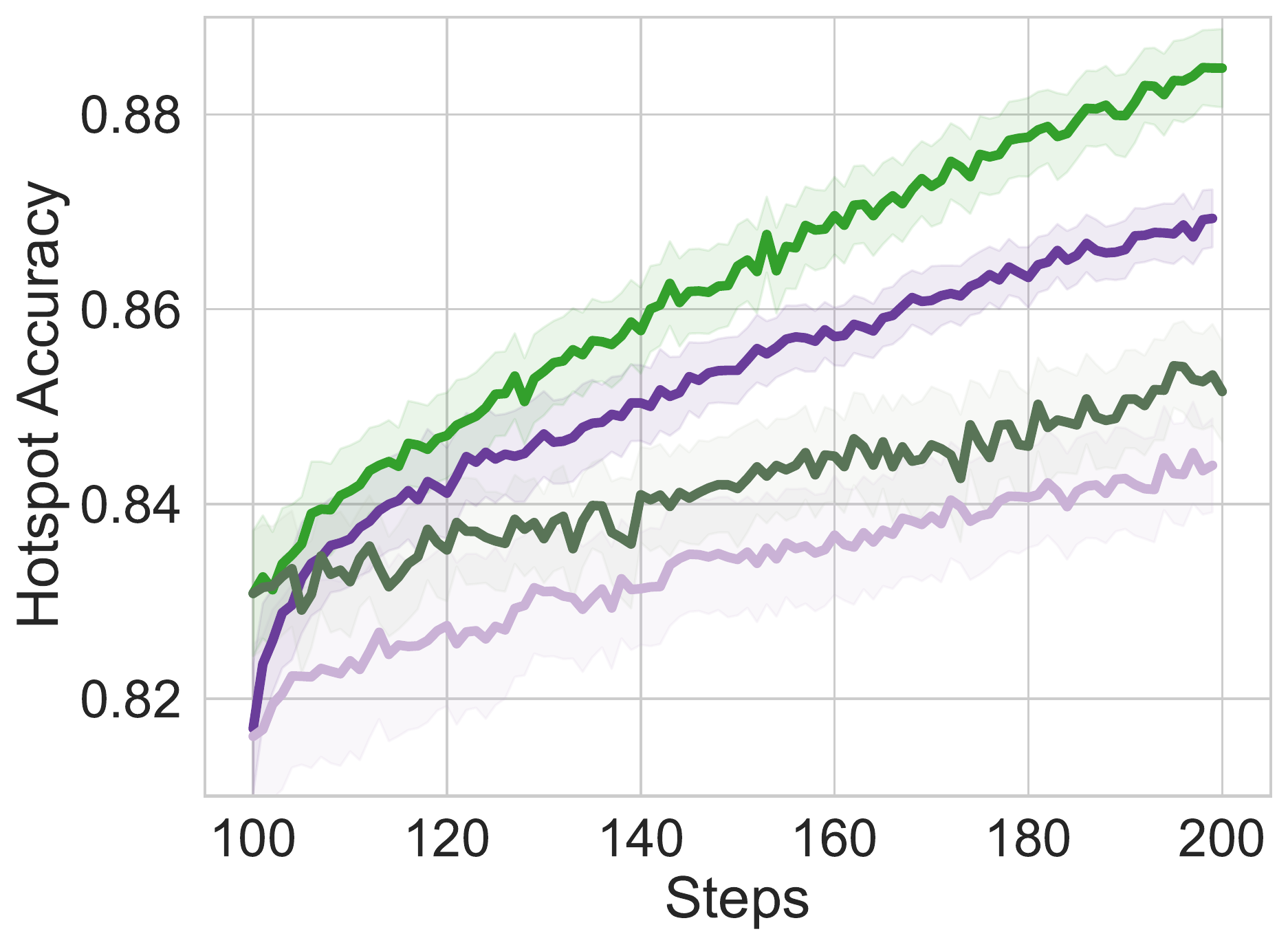}
\caption{Hotspot prediction accuracy}
\label{fig:nipv_binomial_acc}
\end{subfigure}
\begin{subfigure}{0.32\textwidth}
\centering
\includegraphics[width=\linewidth]{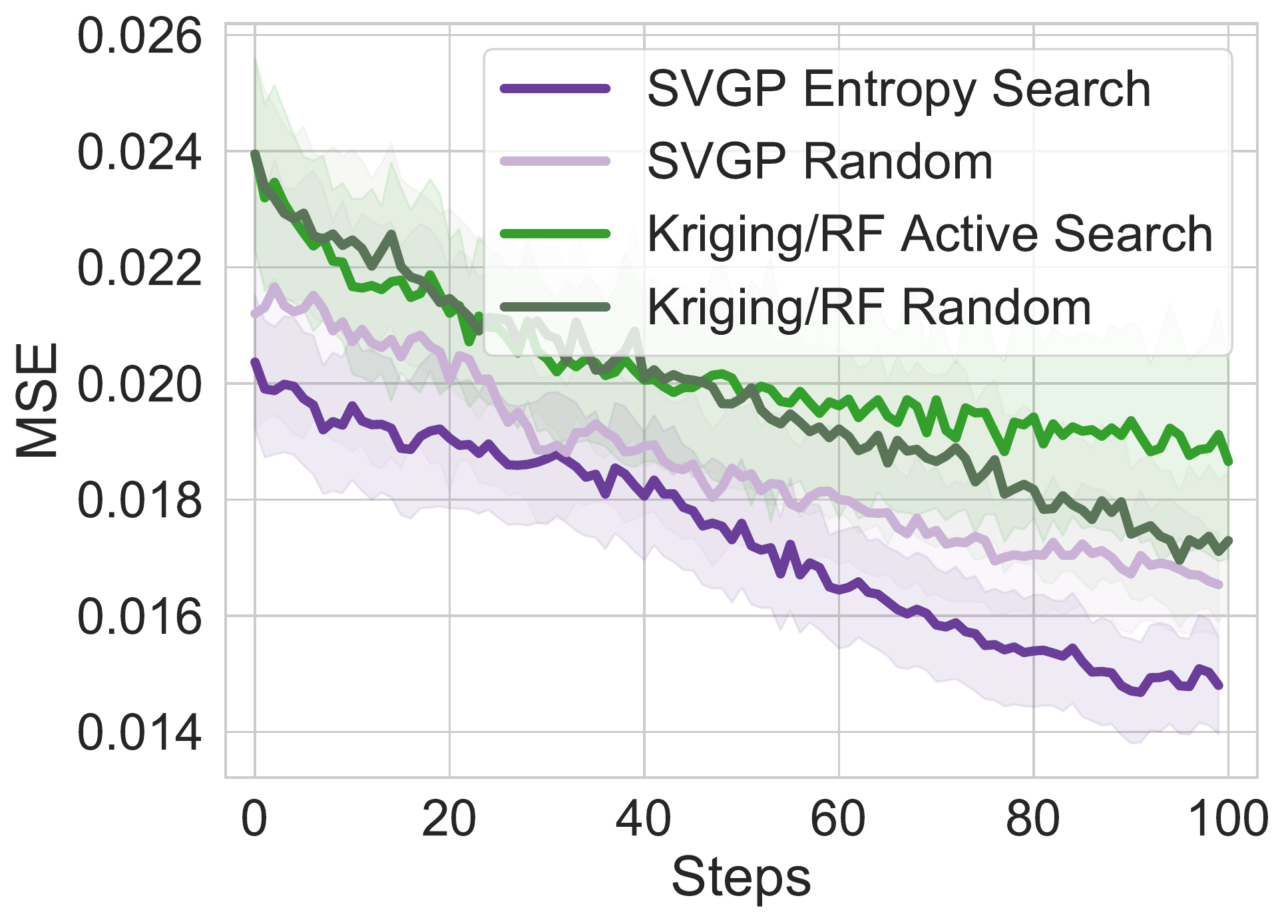}
\caption{Prevalence modelling}
\label{fig:nipv_binomial_global}
\end{subfigure}
\caption{\textbf{(a)} Active learning of malaria incidence from satellite data.
Using qNIPV outperforms randomly selecting points, while the SVGPs slightly outperform both exact GPs and WISKI.
\textbf{(b,c)} Active learning of schistomiasis incidence in Cote d'Ivoire from \citet{andrade2020finding}. Comparison is to the random forest based approach using active sampling. While the GP based models are somewhat less accurate at predicting hotspots \textbf{(b)}, they are better as a global model of prevalence \textbf{(c)}.}
\label{fig:svgp_nipv_al}
\end{figure*}

\textbf{Modelling of Malaria Incidence:}
We consider data from the Malaria Global Atlas \citep{weiss2019mapping} describing the infection rate of a parasite known to cause malaria in 2017.
We wish to choose spatial locations to query malaria incidence in order to make the best possible predictions on a withheld test set, the entire country of Nigeria.
Following \citet{stanton_kernel_2021}, we minimize the negative integrated posterior variance \citep[NIPV,][]{seo2000gaussian}, defined as 
$a(\vec x; \mathcal{D}) := -\int_{\mathcal{X}} \mathbb{E}(\mathbb{V}(f(\vec x) | \mathcal{D}_{+\vec x} ) | \mathcal{D}) d\vec x,$
again with $\mathcal{D}_{+\vec x} = \mathcal{D} \cup \{(\vec x, y)\}.$
Intuitively, the minimizer of this acquisition will be the batch of data points that when added into the model will most reduce the total posterior uncertainty across the domain, requiring efficient conditioning to do so in a tractable manner.
The results are shown for a batch size of $q = 6$ across $15$ trials in Figure \ref{fig:nipv_malaria} where we see that each method outperforms random baselines.
Perhaps due to the optimization freedom, the SVGP outperforms both the exact GP and WISKI \citep{stanton_kernel_2021}.

\textbf{Hotspot Modelling: }
We follow \citet{andrade2020finding} and model the prevalence of schistomiatosis in C\^{o}te d'Ivoire using simulated responses from $1500$ villages in that country, and taking into account six other demographic variables.
We model the responses $y$ (incidence) at locations $\vec x$ with population $n(\vec x)$ with a Binomial likelihood $p(y | f, \vec x) \sim \mathrm{Binomial}(n(\vec x), r(f))$, where $r(f) = (1 + \exp\{-f\})^{-1}$. 
Letting $\tau \in (0, 1)$ be a threshold on the prevalence for a location to be considered a ``hotspot" \citep{andrade2020finding}, we compute the entropy:
\begin{align*}
    h_\tau(\vec x, \mathcal{D}) := \mathbb{E}_{p(f|\mathcal{D})}(\mathbb{H}(\text{Bernoulli}(f > \text{logit}(\tau)))) \approx \frac{1}{K}\sum_{i=1}^K 1_{f > \text{logit}(\tau)} \mathbb{H}(\text{Bernoulli}(f)),
\end{align*}
taking the acquisition value to be the reduction in the entropy of the posterior predictive distribution over the incidence under the hotspot-focused likelihood. 
\begin{align}
   a_\tau(\vec x, \mathcal{D}) := \int_{\vec x' \in \mathcal{X}} \big(h_\tau(\vec x', \mathcal{D}_{+\vec x}) - h_\tau(\vec x', \mathcal{D})\big)d\vec x'. \label{eq:hotspot_acq_fn}
\end{align}
A location is given a high acquisition value if observing the incidence at that location reduces the uncertainty of the model on the predicted set of hotspots. In Figure \ref{fig:nipv_binomial_acc}, we compare to \citet{andrade2020finding} who use spatial kriging on the residuals of a random forest model.
Both their random baseline and their exploration based procedure (a variant of UCB) start off with higher prediction accuracy; however our SVGP models ultimately outperform the kriging approach with random selection.
The SVGP is a better predictor of true prevalence, as shown in Figure \ref{fig:nipv_binomial_global}.
In both cases, our acquisition function significantly outperforms random selection with a SVGP surrogate.

\begin{figure*}[t!]
\centering
	\begin{subfigure}{\textwidth}
		\centering
		\includegraphics[width=0.9\linewidth,clip,clip,trim=0cm 0cm 0cm 14.5cm]{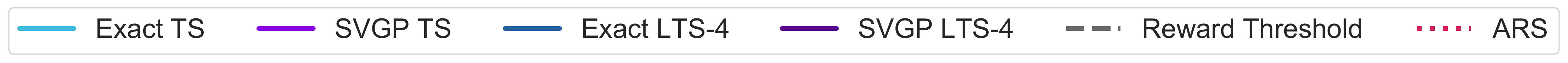}
	\end{subfigure}
\captionsetup[subfigure]{justification=centering}
\begin{subfigure}{0.24\textwidth}
\centering
\includegraphics[width=\linewidth]{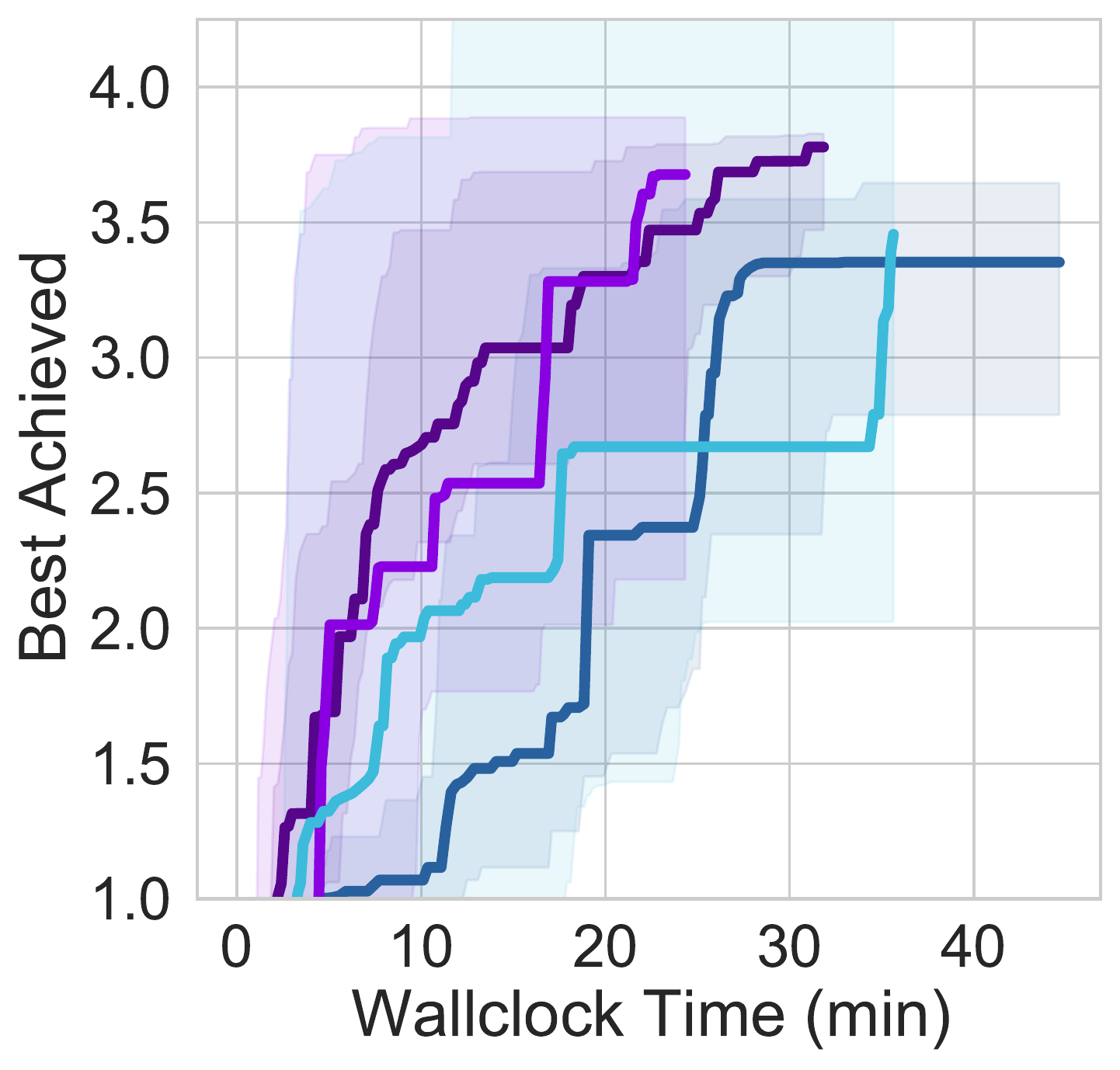}
\caption{Wall-Clock time, rover, $d=60$}
\label{fig:trbo_wallclock}
\end{subfigure}
\begin{subfigure}{0.24\textwidth}
\centering
\includegraphics[width=\linewidth]{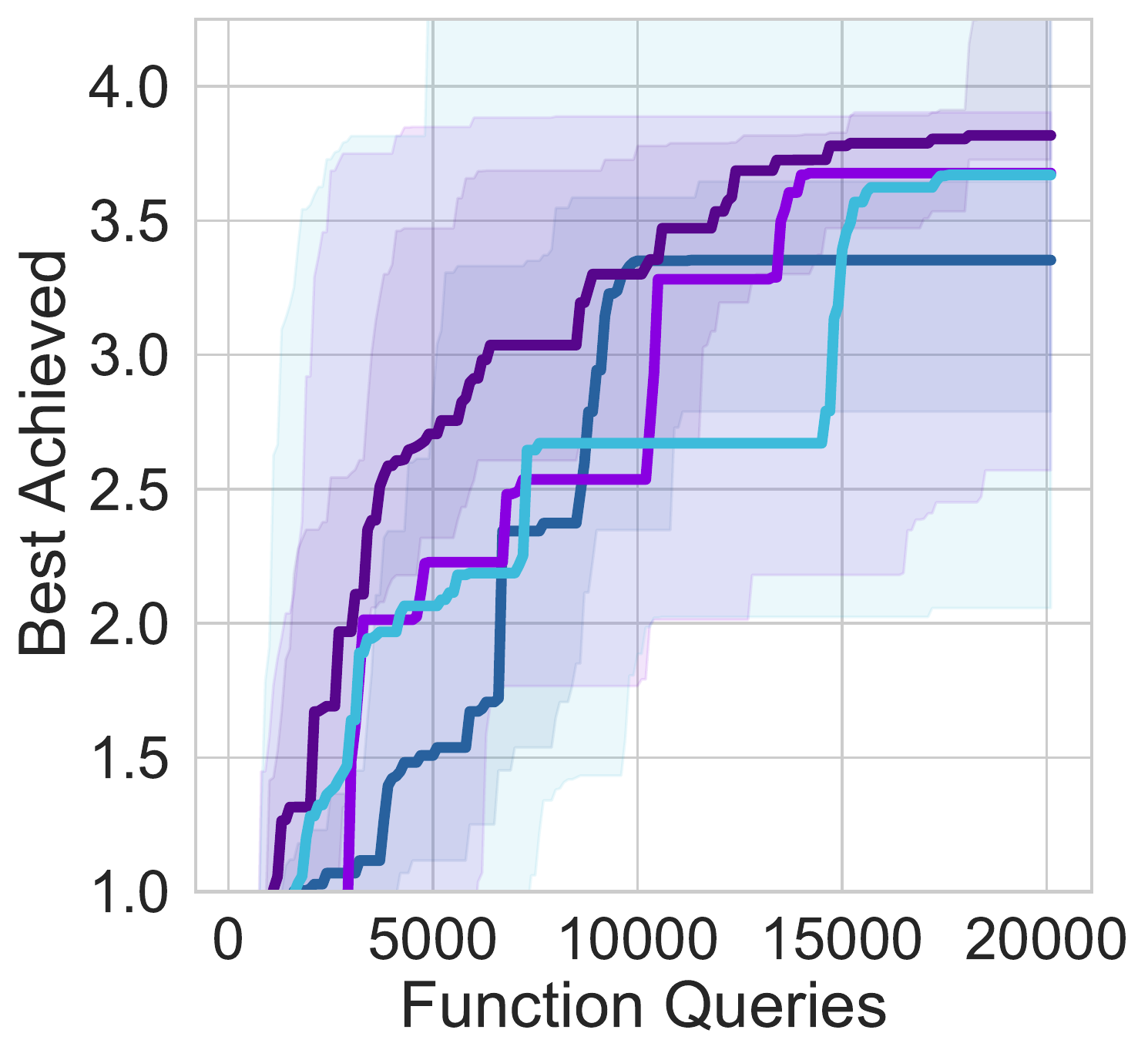}
\caption{rover, \newline $d=60$}
\label{fig:trbo_60_main}
\end{subfigure}
\begin{subfigure}{0.24\textwidth}
\centering
\includegraphics[width=\linewidth]{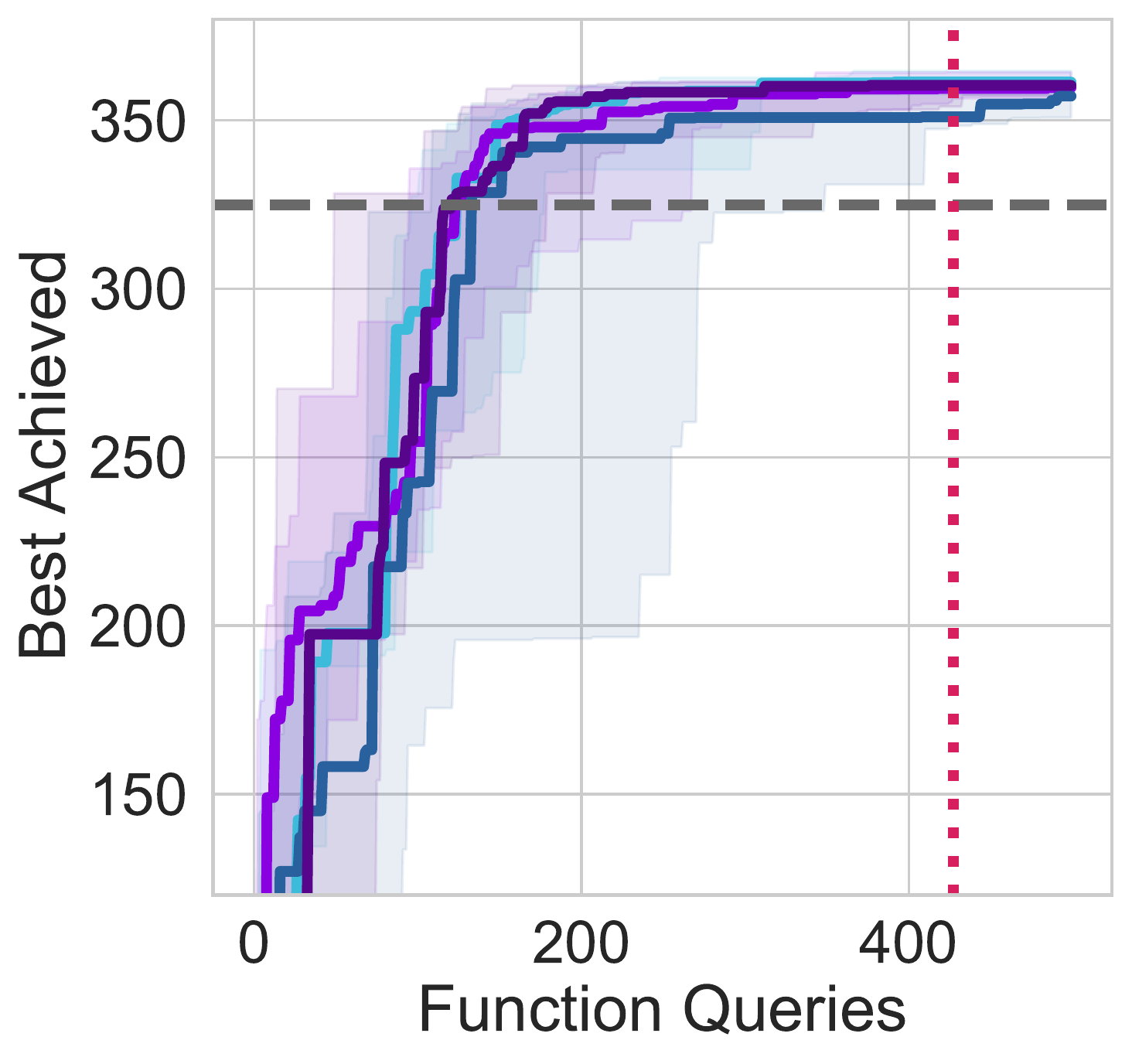}
\caption{\texttt{swimmer},\newline $d=16$}
\label{fig:trbo_swimmer}
\end{subfigure}
\begin{subfigure}{0.24\textwidth}
\centering
\includegraphics[width=\linewidth]{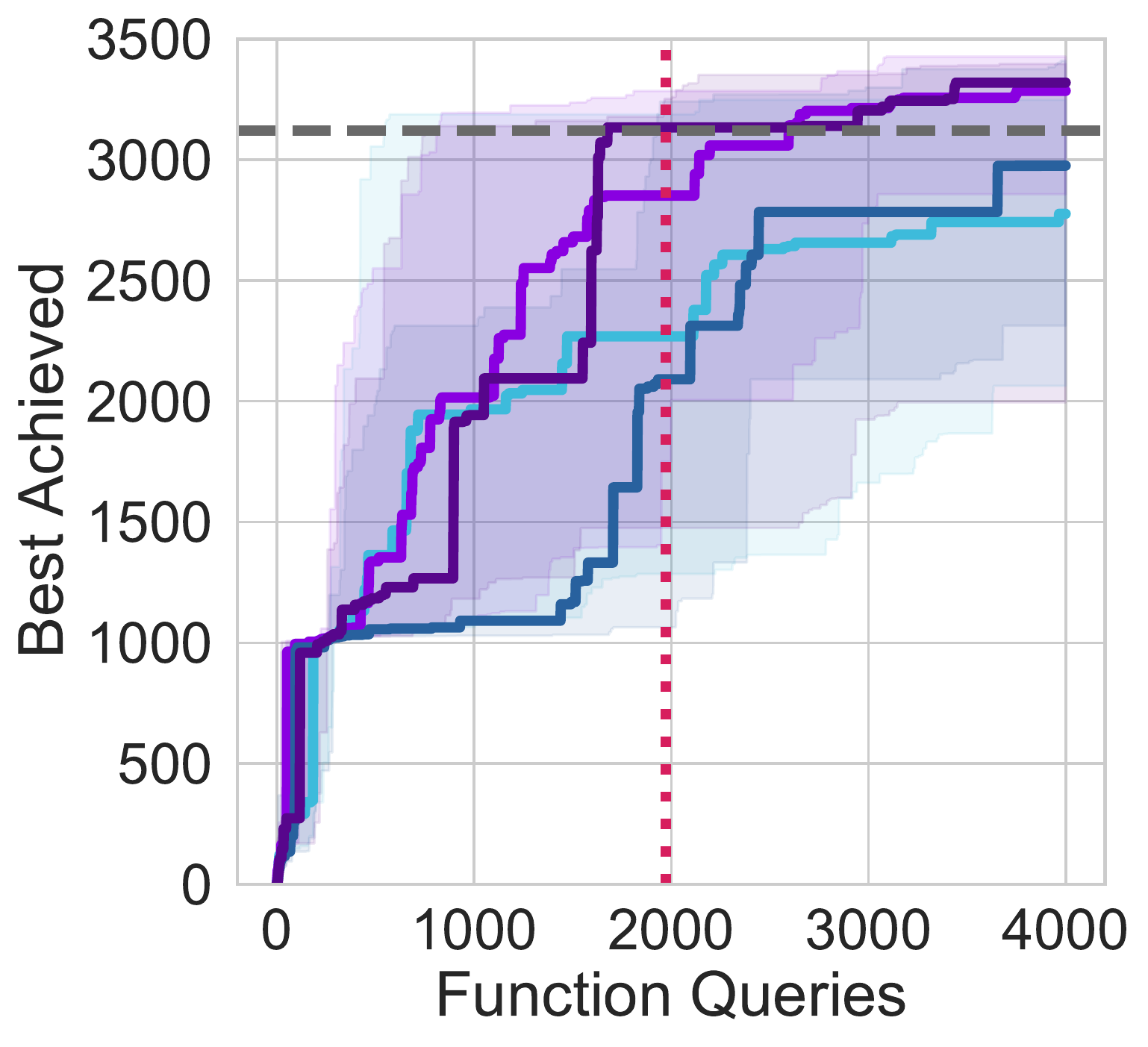}
\caption{\texttt{hopper},\newline $d=33$}
\label{fig:trbo_hopper}
\end{subfigure}
\caption{
Multi-step rollouts with OVC and SVGPs provides sample-complexity and wall-clock time improvements on high dimensional BO problems when using TurBO and LAMCTS \citep{eriksson2019scalable,wang2020learning}. 
SVGP rollouts are as time efficient (to $150$ iterations) as standard TS \textbf{(a)} on lunar rover, $d=60.$
\textbf{(c-d)} MuJoCo environments using LAMCTS + TurBO.
Also shown is the reward threshold (dashed grey lines) and augmented random search (ARS)'s performance (dotted red lines)  \citep{mania2018simple}. 
The median and its $95\%$ confidence interval are shown over $24$ trials for rover and $10$ trials for swimmer and hopper.
}
\label{fig:trbo}
\end{figure*}

\subsection{Rollouts within Thompson Sampling for High Dimensional BO}

For our final set of experiments, we solve control problems using trust region Bayesian optimization \citep[TurBO, ][]{eriksson2019scalable}.
Inspired by multi-step look-aheads \citep{jiang_efficient_2020,bertsekas2020rollout}, we propose $h$-step look-ahead Thompson sampling (LTS-$h$). 
In BO, Thompson sampling (TS) is often implemented by drawing samples from the posterior at points all over the domain, then selecting the $q$ best to form a query batch \citep{thompson1933likelihood, eriksson2019scalable}. 
LTS-$h$ extends the idea by conditioning the surrogate independently on each posterior sample in the original TS query batch, Thompson sampling again from the updated posterior with a new set of points, and appending the best sample to its predecessor to form a \emph{path}. The process is repeated $h$ times. 
Finally, we condition the original surrogate jointly on each path, then perform TS again to choose the new query batch.
Informally, each path corresponds to a distinct, coherent draw from $p(f | \mathcal{D})$, allowing the inner-loop to refine its guess of the global optimum for different $f$, and the final round of TS chooses the query batch based on those guesses. See Appendix \ref{subsec:look_ahead_ts_defn} for a formal description. Like other look-ahead acquisitions, LTS-$h$ is only practical if posterior conditioning and samples are very efficient and numerically stable. LTS-$h$ is conceptually similar to path sampling for look-ahead in \citet{jiang_efficient_2020} and kriging believer \citep{ginsbourger2010kriging}.

For validation, we consider tuning the $60$ dimensional path that a \textbf{lunar rover} takes across a field stacked with obstacles  \citep{wang2018batched,eriksson2019scalable}.
We use batches of $q = 100$ with $200$ initial points, and use trust region Bayesian optimization (TurBO) to split up the space effectively \citep{eriksson2019scalable}, comparing to TurBO with Thompson sampling (TS) as the acquisition \citep{thompson1933likelihood}.
We show the wall clock times per iteration in Figure \ref{fig:trbo_wallclock},
where we see that the TurBO + LTS approaches compare well in wallclock time to TurBO+TS, while being more function efficient (Figure \ref{fig:trbo_60_main}). 

Finally, we consider \textbf{MuJoCo} problems using the OpenAI gym \citep{todorov2012mujoco,brockman2016openai} with LTSs inside of TurBO with trust regions generated by Monte Carlo Tree Search following the procedure of \citet{wang2020learning}.
Following \citet{wang2020learning}, we learn a linear policy and consider the \texttt{swimmer-v2, hopper-v2} environments over $10$ trials displaying the median and its $95\%$ confidence band due to high variance.
On both problems, SVGP with LTSs tend to be the most sample efficient, with SVGP + TS performing at least as well on \texttt{swimmer-v2}.
We also show the reward threshold and the performance achieved by augmented random search (ARS), which is a strong baseline reinforcement learning method that uses random search to tune linear controllers \citep{mania2018simple}.
Our results suggest that LTSs are promising overall; however, more work needs to be done for high dimensional kernels on these problems.

\section{Discussion}
In conclusion, we have demonstrated how to efficiently condition on new data points with stochastic variational Gaussian processes via closed form updates to the variational distribution.
Our conditioning approach generalizes exact GP conditioning via Laplace approximations for non-Gaussian likelihoods.
As a result, we have decoupled look-ahead BO acquisition functions from their dependence on exact GP inference through a Gaussian likelihood, increasing the range, scale, and efficiency of BO applications. In the future, we hope to extend OVC to multi-task and deep Gaussian processes for use in Bayesian optimization \citep{hebbal2021bayesian,cutajar2019deep}.

\section*{Acknowledgements}
WJM, SS, AGW are supported by an Amazon Research Award, NSF I-DISRE 193471, NIH R01 DA048764-01A1, NSF IIS-1910266, and NSF 1922658 NRT-HDR: FUTURE Foundations, Translation, and Responsibility for Data Science.
WJM was additionally supported by an NSF Graduate Research Fellowship under Grant No. DGE-1839302.
SS is additionally supported by the United States Department of Defense through the National Defense Science \& Engineering Graduate (NDSEG) Fellowship Program.
We'd like to thank Greg Benton for setting up the volatility experiment and for helpful discussions and Nate Gruver and Eytan Bakshy for helpful comments.

\bibliographystyle{apalike}
\bibliography{refs}

\clearpage
\appendix

\renewcommand\theequation{A.\arabic{equation}}   
\setcounter{equation}{0}

  \vbox{
    \hsize\textwidth
    \linewidth\hsize
    \vskip 0.1in
      \hrule height 4pt
  \vskip 0.25in
  \vskip -\parskip
    \centering
    {\centering \LARGE\bf Supplementary Materials for Conditioning Sparse Variational Gaussian Processes for Online Decision-making\par}
      \vskip 0.29in
  \vskip -\parskip
  \hrule height 1pt
  \vskip 0.1in
  }

\section*{Table of Contents}
The Appendix is structured as follows:
\begin{itemize}
    \item Appendix \ref{app:limitations} discusses broader impacts and limitations.
    \item Appendix \ref{app/sec:further_background} discusses the background in more detail.
    \item Appendix \ref{app/sec:further_methods} discusses Newton's method and training objectives for the variational models, before giving more detail on the methodological work, including implementation details.
    \item Appendix \ref{app:exp_data} shows two more understanding experiments along with ablation studies, with Appendix \ref{app:data} giving detailed experimental and data descriptions.
\end{itemize}
\section{Limitations and Societal Impacts}\label{app:limitations}

\subsection{Limitations}
From a practical perspective, we see several interrelated limitations:
\begin{itemize}
    \item In our software implementation, we currently support single-output functions. This is partly because GPyTorch currently has limited support for fantasization of multi-task Gaussian processes; see \url{https://github.com/cornellius-gp/gpytorch/pull/805}. 
    \item For non-Gaussian likelihoods, the approximation may break down for long rollout time steps due to accumulating error from the Laplace approximation.
    \item Incorrect modelling and thus incorrect rollouts will tend to be even more influential in high dimensional settings, as the kernels we use do not tend to work particularly well in high dimensions \citep{eriksson2019scalable}.
    \item If the underlying data is non-stationary, then long range predictions may suffer. 
    For example, local models  are superior on the rover problem in Figure \ref{fig:trbo}, even  compared to global models with advanced acquisition functions (e.g. Figure \ref{fig:rover_global}). This limitation is remedied somewhat by the local modelling approaches we use \citep{eriksson2019scalable,wang2020learning}.
    \item Using many inducing points can worsen numerical conditioning, leading to less stable multi-step fantasization, despite the theoretical advantages of more inducing points \citep[e.g.,][]{bauer2016understanding}.
\end{itemize}

\subsection{Societal Impacts}
We do not anticipate that our work will have negative societal impacts.
To the contrary, as we demonstrate in this paper, Bayesian optimization is naturally suited to applications in the public interest, such as public health surveillance \citep{andrade2020finding}.
These types of applications should help benefit global populations by allowing more targeted interventions in the public health setting.
However, reliance solely on machine learning models, for example, in quantitative finance settings (as our volatility model in Figure \ref{fig:conditioning_gpcv} is designed for), could potentially lead to over-confidence and financial shocks as has previously been the case \citep{mackenzie2014formula}.

\section{Further Background}
\label{app/sec:further_background}

In this section, we describe further related work on both variational inference in the streaming setting as well as the use of sparse GPs in both BO and control before describing Newton's iteration for Laplace approximations and training methods for exact and variational GPs.

\subsection{Extended Related Work}\label{app:rel_work}
Because our work explores three distinct applications, namely black-box optimization, active learning, and control, there was more noteworthy related work than the space constraints of the main text would permit. Here we present a more complete literature review. 

We do not focus on other approaches for streaming variational GP inference ranging from decoupled inducing points \citep{cheng_incremental_2016,kapoor2020variational,shi2020sparse} to alternative variational constructions \citep{moreno2019continual,moreno2020recyclable} to Kalman filtering based approaches \citep{huber2014recursive,huber2013recursive} to deep linear model approaches \citep{pan2020continual,titsias2019functional}.
These approaches could potentially be folded into OVC; however, we leave exploration of these for related work.

We also focus solely on the variationally sparse GPs introducted by \citet{titsias_variational_2009} and \citet{hensman_bigdata_gp}, rather than other approaches, which are potentially amenable to being used within OVC.
Of particular note, the classical variational approximations for non-Gaussian likelihoods such as the approaches proposed originally by \citet{csato2002sparse} and later \citet{opper2009variational} seem particularly promising, as is the development of kernel methods within the exponential family more generally \citep{CANU2006714}.

From a software perspective, both GPFlowOpt \citep{GPflowOpt2017} and its successor Trieste (\url{https://github.com/secondmind-labs/trieste}) seem to support variational GPs for BO but we do not know of a comprehensive benchmarking of their implementation.
Implementing variational GPs for most acquisitions in BoTorch \citep{balandat_botorch_2020} is entirely possible using GPyTorch, as we did, although it requires some engineering work and is not natively supported at this time.

Recent work on molecular design has combined variational auto-encoders (VAEs) trained to map high dimensional structured molecular representations to low-dimensional latent representations with SVGPs trained to predict a molecular property of interest from the latent encoding \citep{gomez2018automatic,tripp2020sample}, but they primarily use simple acquisition functions.

In the control literature, sparse GPs are more popular, stemming from the seminal works of \citet{csato2002sparse} and \citet{girard2002gaussian}.
\citet{chowdhary2014bayesian} directly apply the approach of \citet{csato2002sparse} to optimal control problems, while \citet{ling2016gaussian} sparsify GPs for active learning and planning.
\citet{groot2011multiple}, \citet{boedecker2014approximate}, and \citet{bijl2016online} perform multiple time step look ahead via moment matching using sparse GPs, while \citet{pan2017prediction} use a similar approach with random fourier features.
\citet{deisenroth2011pilco} use sparse GPs to speed up dynamics models for robotics, while \citet{saemundsson2018meta} use sparse GPs for meta reinforcement learning.
Finally, \citet{xu2014gp} use sparse GPs for robot localization tasks in control.

\subsection{Laplace Approximations and Newton's Iteration}
Newton iteration iterates 
\begin{align*}
    f_{t+1} &= (K^{-1} + W)^{-1}(W f_t + \nabla \log p(y | f_t)) \\
    &= (K - KW^{-1/2}B^{-1}W^{-1/2}K)(W f_t + \nabla \log p(y | f_t)), \hspace{1cm} B := (I + W^{1/2} K W^{1/2}),
\end{align*}
with $W = - \nabla^2_f \log p(y | f),$ e.g. the negative Hessian of the log likelihood, until $f_{t+1} - f_t < \epsilon$ (e.g. until a stationary point is reached).
We implemented a batch version of the convergence rule, stopping when the total difference is under a threshold for all batches of $f.$
The expensive computational cost is that of the solve of $B,$ as $W$ is diagonal.
In our case, we only consider Newton iteration on test points, so that the time complexity is just $\mathcal{O}(n_{\text{test}}^3).$
Please see \citet[Chapter 3 of][]{rasmussen_gaussian_2008} for further detail.

Implementation wise, for all but the GPCV experiment in Figure \ref{fig:conditioning_gpcv}, we manually implemented the gradient and Hessian terms as they are well known due to being natural parameterizations of exponential families.
For the preference learning experiment, we followed the gradient derivation in \citet{chu2005preference}.

\subsection{Training Mechanisms for Exact Gaussian Processes and Variational Gaussian Processes}\label{app:training_losses}

Please see the more detailed summaries of \citet{rasmussen_gaussian_2008} for training methods of exact Gaussian processes as well as the theses of \citet{van2019sparse} and \citet{matthews2017scalable} for training methods of sparse Gaussian processes.
In what follows, we assume solely a zero mean function and suppress dependence on $\theta$ for the kernel matrices.

\paragraph{Log Marginal Likelihood for Exact GPs}
Training the GP's hyper-parameters, $\theta,$ proceeds via maximizing the marginal log-likelihood (MLL).
The MLL is given by
\begin{align}
	\log p(\mathbf{y} | X, \theta) = &-\frac{1}{2}\mathbf{y}^\top (K_{\vec v \vec v} + \sigma^2 I )^{-1} \mathbf{y} - \frac{1}{2}\log|K_{\vec v \vec v} + \sigma^2 I | - \frac{n}{2} \log 2\pi,
	\label{app:eq:marginal_likelihood}
\end{align}
over the training set, $\mathcal{D} = (X, \mathbf y).$

\paragraph{Collapsed Evidence Lower Bound for SGPR}
Sparse Gaussian process regression (SGPR) begins by approximating the kernel with a lower rank version.
The training data covariance, $K_{\vec v \vec v},$ is replaced by the Nystr\"om approximation $K_{\vec v \vec v} \approx Q_{\vec v \vec v} := K_{\vec v \vec u} K_{\vec u \vec u}^{-1} K_{\vec u \vec v},$ 
where $K_{\vec u \vec u}$ is the kernel evaluated on the inducing point locations, $Z.$
SGPR learns the locations of the inducing points and the kernel hyperparmeters through a 'collapsed' form of the evidence lower bound (ELBO), yielding a variational adaptation of older Nystr\"om or projected process approximations \citep[Chapter 7]{rasmussen_gaussian_2008}.

The ELBO is called ``collapsed" because the Gaussian likelihood allows the parameters of the variational distribution to be analytically optimized and integrated out (collapsed), yielding a bound that depends only on the inducing point locations and the kernel hyperparameters \citep{titsias_variational_2009}. The SGPR bound is as follows (see \citet{titsias_variational_nodate} for the full derivation):
\begin{align}
    \mathcal{F}(\theta, Z) := \log p(y | 0, \sigma^2 I + Q_{\vec v \vec v}) - \frac{1}{2\sigma^2}\mathrm{trace}(K_{\vec v \vec v} - Q_{\vec v \vec v}) \label{eq:sgpr_elbo}.
\end{align}
We can apply standard gradient based training to both the kernel hypers $\theta$ as well as the inducing locations $Z.$
\citet{jankowiak2020parametric} derive a variational version of this bound which enables sub-sampling across data points; we leave training with that bound for future work.

\paragraph{Evidence Lower Bound for SVGP}
The advance of \citet{hensman_bigdata_gp} is that they do not compute the optimal variational parameters at each time step proposing the sparse variational GP (SVGP).
Their derivation, see \citet{hensman_bigdata_gp,hensman15} for a full derivation, yields an 'uncollapsed' ELBO (so named due to its explicit dependence on the variational parameters),
\begin{align}
    \mathcal{F}(\theta, Z, \vec m_{\vec u}, S_{\vec u}) := \mathbb{E}_{q(f)} \log p(y | f) - \text{KL}(\phi(\vec u) || p(\vec u)) \label{eq:svgp_elbo},
\end{align}
where $q(f) = \int p(f | \vec u) \phi(\vec u) d\vec u.$ 
The latent GP $f$ is replaced with the variational distribution, $\phi(\vec u) = \mathcal{N}(\vec m_{\vec u}, S_{\vec u}).$
$q(f)$ can be determined via Gaussian marginalization from $\phi(\vec u).$
and is 
$
q(f) = \mathcal{N}(K_{\vec v \vec u}K_{\vec u \vec u}^{-1}\vec m_{\vec u}, K_{\vec v \vec v} - K_{\vec v \vec u}K_{\vec u \vec u}^{-1}(K_{\vec u \vec u} - S_{\vec u})K_{\vec u \vec u}^{-1}K_{\vec u \vec v}).
$
Mini-batching is possible because the first term in Eq. \ref{eq:svgp_elbo} splits over each of the $n$ data points as each $y_i$ is conditionally independent given $f$, allowing sub-sampling.
Sub-sampling and mini-batching enable the use of stochastic optimization techniques, reducing the per iteration cost to be constant in $n,$ the number of data points.

\citet{bui_streaming_2017} introduce a streaming version of the ELBO that involves two further terms; the method is called O-SVGP.
Their primary model, O-SGPR (called the ``collapsed bound'' of a streaming sparse GP in that paper) is trained through a variant of their ELBO bound that integrates out the variational parameters.
We describe this collapsed bound in more detail in Appendix \ref{app/subsec:o_sgpr_degrades}.

\paragraph{Predictive Log Likelihood for SVGP}
We close this section by quickly describing the predictive log likelihood (PLL) method of \citet{jankowiak2020parametric}, which is motivated by attempting to improve the calibration of SVGP models trained via the ELBO.
The key advance in \citet{jankowiak2020parametric} is that they consider the expectation over both the data and the response, rather than simply the expectation over the response in the variational distribution, producing an objective that becomes:
\begin{align*}
    \mathcal{F}(\theta, x_m, \vec m_{\vec u}, S_{\vec u})&:= \mathbb{E}_{p(y,x)}\log p(y |f, x) - \text{KL}(\phi(\vec u) || p(\vec u)) \\
    &\approx \sum_{i=1}^N \log \left(\mathbb{E}_{\phi(\vec u)p(f_i | \vec u, x_i)}p(y_i | f_i)\right) - \text{KL}(\phi(\vec u) || p(\vec u)).
\end{align*}
We consider this optimization objective for several of the larger-scale problems here.
\section{Further Methodological Details}\label{app/sec:further_methods}

In this Appendix, we begin by presenting our approach, OVC, as a generalization of exact Gaussian conditioning for SGPR (\ref{app:special_case}) before describing an alternative interpretation of \citet{bui_streaming_2017} that is equivalent to our approach in \ref{app:interp}. 
Then, in Appendix \ref{app/subsec:whitened_details}, we describe the practical implementation of OVC. 
In \ref{app/subsec:o_sgpr_degrades}, we describe a flaw of the O-SGPR bound for small batch sizes. 
Finally, in \ref{subsec:look_ahead_ts_defn}, we give an extended description of look-ahead Thompson sampling (LTS).

\subsection{OVC Generalizes Efficient Batch SGPR Conditioning}\label{app:special_case}

In this section, we show that OVC can also be viewed as a generalization of Gaussian conditioning for SGPR. Gaussian conditioning for SGPR is recovered as a special case of OVC when $\theta ' = \theta$ and $Z' = Z$. 
Under this assumption, Eqns. \eqref{eq:streaming_c} and \eqref{eq:streaming_C} simplify as follows:
\begin{align}
    \vec c = K_{\vec u \vec v} \Sigma_{\vec y}^{-1} \vec y + \textcolor{blue}{\vec c'}, \hspace{4mm}
    C = K_{\vec u \vec v} \Sigma_{\vec y}^{-1} K_{\vec v \vec u} + \textcolor{blue}{C'}.
    \label{app:eq_sgpr_c_updates}
\end{align}

Considering the predictive distribution given by Eq. \ref{eq:sgpr_predictive} in the main text, as we add new data points, $(X_{\mathrm{batch}}, \vec y),$ into the model, we need to update $A = (K_{\vec u \vec u} + C)^{-1}$, $\textcolor{orange}{\vec a} = A \vec c$, and $\textcolor{orange}{R} = (K_{\vec u \vec u}^{-1} - A )^{1/2}$. 
For homoscedastic likelihoods the $A$ update is a fast low-rank update via the Woodbury matrix identity,
\begin{align}
    A = &\textcolor{blue}{A'} - \textcolor{blue}{A'} K_{\vec u \vec v}(\sigma^2 I + K_{\vec v \vec u}\textcolor{blue}{A'} K_{\vec u \vec v})^{-1}K_{\vec v \vec u}\textcolor{blue}{A'}.
\end{align}
This produces efficient Sherman-Morrison updates to generate $\textcolor{orange}{\vec a}$ (via addition and matrix vector multiplication) and Woodbury based updates to update $\textcolor{orange}{R},$ the predictive covariance cache via low-rank updates (e.g. Proposition 2 of \citet{jiang_efficient_2020}).

Furthermore, these low rank updates can be used to produce updates to the exact caches.
These updates are simply exact Gaussian conditioning with an approximate kernel.
That is, the SGPR caches are merely transformed exact caches for any given set of data points.
To demonstrate, we use a Nystr\"om approximation for the kernel throughout, e.g. $K_{\vec v \vec w} \approx K_{\vec v \vec u}K_{\vec u \vec u}^{-1} K_{\vec u \vec w},$
then after applying Woodbury we can write the exact caches using $A$:
\begin{align*}
    \textcolor{orange}{R R^\top} &= (K_{\vec v \vec v} + \Sigma_{\vec y})^{-1} = \Sigma_{\vec y}^{-1} - \Sigma_{\vec y}^{-1}K_{\vec v \vec u}(K_{\vec u \vec u} + K_{\vec u \vec v}\Sigma_{\vec y}^{-1} K_{\vec v \vec u})^{-1} K_{\vec u \vec v} \Sigma_{\vec y}^{-1} \\
    &= \Sigma_{\vec y}^{-1} - \Sigma_{\vec y}^{-1}K_{\vec v \vec u}A K_{\vec u \vec v}\Sigma_{\vec y}^{-1}
\end{align*}
with a similar expression for the predictive mean cache as 
\begin{align*}
    \textcolor{orange}{\vec a} = (K_{\vec v \vec v} + \Sigma_{\vec y})^{-1} \vec y = \Sigma_{\vec y}^{-1} \vec y - \Sigma_{\vec y}^{-1}K_{\vec v \vec u}A \vec c.
\end{align*}
Next we take the caches computed on every training point and project them into the space of inducing points by multiplying them by $K_{\vec u \vec u}^{-1} K_{\vec u \vec v}$. 
It takes a bit of algebra, but we can derive updated expressions for $\textcolor{orange}{R R^\top}$ and $\textcolor{orange}{\vec a}$ in terms of solely the new covariance matrix, $K_{\vec{v} \vec u}$ and the new responses, $\Sigma_{\vec{y}}^{-1} \vec{y}.$
That is, $\textcolor{orange}{\vec a_{\text{SGPR}}} = K_{\vec u \vec u}^{-1} K_{\vec u \vec v} \textcolor{orange}{\vec a}$ and $\textcolor{orange}{RR^\top_{\text{SGPR}}} = K_{\vec u \vec u}^{-1} K_{\vec u \vec v}\textcolor{orange}{R R^\top} K_{\vec v \vec u} K_{\vec u \vec u}^{-1}$.
\begin{align*}
       \textcolor{orange}{a_{\text{SGPR}}} &= K_{\vec u \vec u}^{-1} K_{\vec u \vec v} ( \Sigma_{\vec y}^{-1} \vec y - \Sigma_{\vec y}^{-1}K_{\vec v \vec u}A K_{\vec u \vec v} \Sigma_{\vec y}^{-1} \vec y) = K_{\vec u \vec u}^{-1} \vec c - K_{\vec u \vec u}^{-1} C (K_{\vec u \vec u} + C)^{-1} \vec c \\
       &=K_{\vec u \vec u}^{-1}((K_{\vec u \vec u}+C)(K_{\vec u \vec u}+C)^{-1} - C(K_{\vec u \vec u}+C)^{-1}) \vec c = (K_{\vec u \vec u} + C)^{-1} \vec c
\end{align*}
and similarly 
\begin{align*}
    \textcolor{orange}{RR^\top_{\text{SGPR}}} & = K_{\vec u \vec u}^{-1} K_{\vec u \vec v} \textcolor{orange}{R R^\top} K_{\vec v \vec u}K_{\vec u \vec u}^{-1} = K_{\vec u \vec u}^{-1} K_{\vec u \vec v}(\Sigma_{\vec y}^{-1} - \Sigma_{\vec y}^{-1}K_{\vec v \vec u}A K_{\vec u \vec v}\Sigma_{\vec y}^{-1}) K_{\vec v \vec u}K_{\vec u \vec u}^{-1} \\
    &=K_{\vec u \vec u}^{-1}K_{\vec u \vec u} - K_{\vec u \vec u}^{-1} C A C K_{\vec u \vec u}^{-1} = K_{\vec u \vec u}^{-1}(K_{\vec u \vec u}(K_{\vec u \vec u}+C)^{-1}CK_{\vec u \vec u}^{-1}) \\
    &= (K_{\vec u \vec u}+C)^{-1}CK_{\vec u \vec u}^{-1} = (K_{\vec u \vec u}+K_{\vec u \vec u}C^{-1}K_{\vec u \vec u})^{-1} \\
    &= K_{\vec u \vec u}^{-1} - (K_{\vec u \vec u} + C)^{-1}.
\end{align*}
Similarly one could follow this logic in reverse to go from SGPR caching to caching for exact GP inference.
We can also use the updates in Eq. \ref{app:eq_sgpr_c_updates} to update the exact GPs caches via first updating the SGPR caches.

To summarize, exact GP regression is just Gaussian conditioning, which can be viewed as a special case of SGPR if one inducing point is placed at every data point. 
SGPR in turn is again Gaussian conditioning through an approximate kernel on projected features, which can be viewed as a special case of O-SGPR if the inducing points and kernel hyperparameters are held fixed. Finally O-SGPR can be viewed as a special case of O-SVGP if the variational parameters are constrained to be optimal.

\subsection{Interpreting \citet{bui_streaming_2017} as O-SGPR}\label{app:interp}
The approach outlined in Section \ref{sec:var_updates} can be verified to be equivalent to streaming sparse GPs (e.g. the un-collapsed bound of \citet{bui_streaming_2017}) mechanically by verifying that the expressions for the ELBO are equivalent (up to constants) and that the predictive mean and variance are exactly equivalent.
Although verifying the equivalence is simply a matter of manipulating algebraic expressions, we have not yet justified the choice of $\textcolor{blue}{\hat{\vec y}}$ and $\textcolor{blue}{\Sigma_{\vec{\hat y}}}$. 
\citet{bui_streaming_2017} obtained their expressions by means of variational calculus, and arrived at the correct result, but did not provide much in the way of intuition for the nature of the optimal solution. 

We now show how the choice of pseudo-targets $\textcolor{blue}{\hat{\vec y}}$ and pseudo-likelihood covariance $\textcolor{blue}{\Sigma_{\vec{\hat y}}}$ has a clear interpretation that obviates any need to appeal to variational calculus except as a formal guarantee of optimality.
Again, suppose we are given a sparse variational GP with inducing points $\textcolor{blue}{Z'}$, kernel hyperparameters $\textcolor{blue}{\theta'}$ and a pre-computed optimal variational distribution $\phi(\vec u') = \mathcal{N}(\textcolor{blue}{\vec m_{\vec u'}}, \textcolor{blue}{S_{\vec u'}})$, and then asked to find the likelihood and dataset of size $p$ that produced the model. 
Although the problem as stated is under-determined, if we choose $X = \textcolor{blue}{Z'}$ and assume the likelihood is some Gaussian centered at $f$, then we can reverse Eqn. \eqref{eq:sgpr_var_dist} (in the main text) to solve for $\vec y$ and $\Sigma_{\vec y}$ as follows:
\begin{align}
    \textcolor{blue}{\vec m_{\vec u'}} = \textcolor{blue}{K_{\vec u' \vec u'}'}(\textcolor{blue}{K_{\vec u' \vec u'}'} + \textcolor{blue}{K_{\vec u' \vec u'}'} \Sigma_{\vec{y}} &\textcolor{blue}{K_{\vec u' \vec u'}'})^{-1} \textcolor{blue}{K_{\vec u' \vec u'}'} \Sigma_{\vec y}^{-1} \vec y, \nonumber \\
    &\Rightarrow \vec y = (\Sigma_{\vec y} \textcolor{blue}{K_{\vec u' \vec u'}'^{-1}} + I) \textcolor{blue}{\vec m_{\vec u'}} = \textcolor{blue}{\hat{\vec y}} \; \\
    \textcolor{blue}{S_{\vec u'}} = \textcolor{blue}{K_{\vec u' \vec u'}'}(\textcolor{blue}{K_{\vec u' \vec u'}'} + \textcolor{blue}{K_{\vec u' \vec u'}'} \Sigma_{\vec{y}} &\textcolor{blue}{K_{\vec u' \vec u'}'})^{-1} \textcolor{blue}{K_{\vec u' \vec u'}'}, \nonumber \\
    &\Rightarrow \Sigma_{\vec y} = (\textcolor{blue}{S_{\vec u'}^{-1}} - \textcolor{blue}{K_{\vec u' \vec u'}'^{-1}})^{-1} = \textcolor{blue}{\Sigma_{\hat{\vec y}}}. \label{eq:pseudo_covariance}
\end{align}

As a result, we can now provide new, intuitive interpretations of $\textcolor{blue}{\hat{\vec y}}$, $\textcolor{blue}{\Sigma_{\hat{\vec y}}}$ and $\phi(\vec u)$. 
In simple terms, the streaming sparse GP (i.e. O-SGPR) of \citet{bui_streaming_2017} is equivalent to a sequence of SGPR models, where instead of training on all previously observed data through the original likelihood at each timestep, each model trains \textit{only} on the combination of the current batch of data $(X_{\mathrm{batch}}, \vec y)$, and the pseudo-data $(\textcolor{blue}{Z'}, \textcolor{blue}{\hat{\vec y}})$ through a pseudo-likelihood with covariance $\Sigma = \mathrm{blkdiag}(\textcolor{blue}{\Sigma_{\hat{\vec y}}}, \Sigma_{\vec y})$. The pseudo-data and pseudo-likelihood together represent all the past data and models. 
Furthermore, $(\textcolor{blue}{Z'}, \textcolor{blue}{\hat{\vec y}})$ and $\textcolor{blue}{\Sigma_{\hat{\vec y}}}$ are the \textit{unique} size-$m$ dataset with $X = \textcolor{blue}{Z'}$ and $f$-centered Gaussian likelihood that could have produced $\phi(\vec u')$, given $\textcolor{blue}{Z'}$ and $\textcolor{blue}{\theta'}$. 
In other words we can think of the tuple $(\textcolor{blue}{\theta'}, \textcolor{blue}{Z'}, \textcolor{blue}{\hat{\vec y}}, \textcolor{blue}{\Sigma_{\hat{\vec y}}})$ as a compressed representation of the sparse GP it defines. 

\subsection{Practical Implementation}
\label{app/subsec:whitened_details}
Implementation wise and to reduce our engineering overhead, we focused on computing $\textcolor{blue}{\hat{\vec y}}$ and $\textcolor{blue}{\Sigma_{\hat{\vec y}}}$ in a numerically stable manner.
We start with the pseudo-covariance term, which can be simplified as
\begin{align*}
    \textcolor{blue}{\Sigma_{\hat{\vec y}}} = I + \textcolor{blue}{S_{\vec u'}(K'_{\vec u' \vec u'} - S_{\vec u'})^{-1}S_{\vec u'}}.
\end{align*}
After some more algebra, we can rewrite the pseudo-observations that depend on the inducing points, 
\begin{align}
    \textcolor{blue}{\hat{\vec y}} &= \textcolor{blue}{\Sigma_{\hat{\vec y}} S_{\vec u'}^{-1} \vec m_{\vec u'}} = \left(I + \textcolor{blue}{S_{\vec u'}(K'_{\vec u' \vec u'} - S_{\vec u'})^{-1}S_{\vec u'}}\right)\textcolor{blue}{S_{\vec u'}^{-1} \vec m_{\vec u'}} \nonumber \\
    &= \textcolor{blue}{S_{\vec u'}^{-1} \vec m_{\vec u'}} + \textcolor{blue}{S_{\vec u'}(K'_{\vec u' \vec u'} - S_{\vec u'})^{-1}\vec m_{\vec u'}}
    \label{eq:pseudo_obs}
\end{align}
For numerical stability, we replace inverses of matrix subtractions as $\left((K - S)^{-1}(K - S)^{-\top}\right)(K - S)^\top,$
dropping subscripts.
While there is still a matrix subtraction here, squaring the system improves the numerical stability of the systems, as we are forcing all of the eigenvalues of the matrices that we are solving linear systems with to be non-negative.

In practice however, we use ``whitening" of the variational distribution as introduced by \citep{matthews2017scalable}.
We instead optimize $\vec{\bar m_{\vec u}} = K_{\vec u \vec u}^{-1/2} \vec m_{\vec u}$ and $\bar S_{\vec u} = K_{\vec u \vec u}^{-1/2} S_{\vec u} K_{\vec u \vec u}^{-1/2}$.
We can rewrite $\textcolor{blue}{\Sigma_{\hat{\vec y}}}$ using the whitened variational covariance matrix $\textcolor{blue}{\bar S_{\vec u'}}$ producing, 
$\textcolor{blue}{\Sigma_{\hat{\vec y}}} = \textcolor{blue}{K_{\vec u' \vec u'}'^{1/2}(\bar S_{\vec u'} + \bar S_{\vec u'} (I - {\bar S_{\vec u'}})^{-1} {\bar S_{\vec u'})K_{\vec u' \vec u'}'^{1/2}}}.$
Again, we square the second term to enhance stability, although it already has a symmetric form; that is, we compute
\begin{align*}
    \textcolor{blue}{\Sigma_{\hat{\vec y}}} = \textcolor{blue}{K_{\vec u' \vec u'}'^{-1/2}(\bar S_{\vec u'} + \bar S_{\vec u'} (I - \bar S_{\vec u'})^{-1}(I - \bar S_{\vec u'})^{-\top} (I - \bar S_{\vec u'})^\top \bar S_{\vec u'})K_{\vec u' \vec u'}'^{-1/2}}.
\end{align*}
Similarly, Eq. \ref{eq:pseudo_covariance} simplifies to become 
\begin{align*}
    \textcolor{blue}{\hat{\vec y}} =\textcolor{blue}{ K_{\vec u' \vec u'}'^{1/2} (I - \tilde S_{\vec u'})^{-1} \vec{\bar m}_{\vec u'}} = \textcolor{blue}{ K_{\vec u' \vec u'}'^{1/2} (I - \tilde S_{\vec u'})^{-1} (I - \bar S_{\vec u'})^{-\top} (I - \bar S_{\vec u'})^\top \vec{\bar m}_{\vec u'}}.
\end{align*}

From an engineering point of view, as we condition into an exact GP for most of our applications, we are able to use the pseudo-likelihood covariance alongside the cache of the pseudo-data covariance, caching a root decomposition of the matrix $(K_{\mathrm{joint}} + \texttt{blkdiag}(\Sigma_{\vec{\hat y}}, \Sigma_{\vec y}))^{-1}$ via low-rank updates to a pre-existing root decomposition of $K_{\vec u \vec u} + \Sigma_{\hat{\vec y}}$ (and its inverse), where $K_{\mathrm{joint}} = k_\theta( \texttt{cat}(Z, X_{\mathrm{batch}}), \texttt{cat}(Z, X_{\mathrm{batch}}) )$ \citep{pleiss_constant-time_2018,jiang_efficient_2020}.
On performing several steps of conditioning (e.g. rollouts), we then use exact GP conditioning via low rank updates as implemented in GPyTorch (which uses the strategy of \citet{jiang_efficient_2020} internally) after the first step.

\subsection{Incremental O-SGPR Tends to Underfit the Data}
\label{app/subsec:o_sgpr_degrades}

In this section we discuss a pathology of the ELBO derived in \citet{bui_streaming_2017} that occurs when O-SGPR is updated on very small batches on new data (e.g. 1 new observation). Tellingly, \citet{bui_streaming_2017} only considered tasks with large batch sizes (around 100 new observations per batch in each task).

\citet{bui_streaming_2017} propose a ``collapsed" evidence lower bound to train their O-SGPR model for each new batch of data.
It is written as:
\begin{align}
    \mathcal{F}_{\mathrm{OSGPR}}(\theta, \vec u) &= \log \mathcal{N}(\texttt{cat}(\hat{\vec y}, \vec y) | \vec 0, Q_{\mathrm{joint}} + \texttt{blkdiag}(\Sigma_{\vec{\hat y}}, \Sigma_{\vec y})) \nonumber \\ 
    & \hspace{8mm} - \frac{1}{2}(\mathrm{trace}_1 + \mathrm{trace}_2) + \mathrm{constants}, \\
    \mathrm{trace}_1 &= \sigma^{-2}\mathrm{Tr}(K_{\vec v \vec v} - K_{\vec v \vec u}K_{\vec u \vec u}^{-1}K_{\vec u \vec v}), \nonumber \\
    \mathrm{trace}_2 &= \mathrm{Tr}( \textcolor{blue}{(S_{\vec u'}^{-1} - K_{\vec u' \vec u'}'^{-1})}(K_{\vec u' \vec u'} - K_{\vec u' \vec u}K_{\vec u \vec u}^{-1}K_{\vec u \vec u'})), \nonumber
\end{align}
where $Q_{\mathrm{joint}} = [K_{\vec u \vec u}, K_{\vec u \vec v}]^\top K_{\vec u \vec u}^{-1}[K_{\vec u \vec u}, K_{\vec u \vec v}]$, and $\mathrm{constants}$ is composed of terms that do not depend on $\theta$ or $Z$. 
In a close parallel to the observations of \citet{titsias_variational_2009} (e.g. Eq. \ref{eq:sgpr_elbo}), we see that the O-SGPR objective is composed of a likelihood term and a trace term (written as $\mathrm{trace}_1 + \mathrm{trace}_2$), the latter acting as a regularizer \citep{titsias_variational_2009,bauer2016understanding}. 
The first part of the trace term, $\mathrm{trace}_1,$ has the same interpretation as the trace term in the batch setting -- it is minimized when the N\"ystrom approximation of the kernel matrix at $X_{\mathrm{batch}}$ is exact (i.e. $K_{\vec v \vec v} = K_{\vec v \vec u}K_{\vec u \vec u}^{-1}K_{\vec u \vec v}$). 
The second trace term, $\mathrm{trace}_2,$ is minimized when $\textcolor{blue}{Z'} = Z$, so it regularizes the new inducing point locations to be close to the old locations. 
If the batch size, $b,$ is much less than $p$ then the trace term is dominated by $\mathrm{trace}_2$, which is after all a sum over $p$ terms, compared to $\mathrm{trace}_1$ which is a sum over $b$ terms.
Since the loss encourages the model to keep $\textcolor{blue}{Z'}$ close to $Z$ to minimize $\mathrm{trace}_2$, the model can simply increase $\sigma^2$ to also decrease $\mathrm{trace}_1$ to explain new observations by under-fitting.
An analogous problem for small batch sizes was described in the Appendix of \citet{stanton_kernel_2021} for the un-collapsed bound (e.g. the training procedure of an O-SVGP model) of \citet{bui_streaming_2017}, necessitating \citet{stanton_kernel_2021} to propose a variant that down-weights the prior terms in the objective.
The cost of down-weighting is an increased tendency towards over-fitting as well as an additional hyper-parameter, both of which were observed by \citet{stanton_kernel_2021}.

\begin{figure*}
    \centering
    \begin{subfigure}{0.19\textwidth}
    \includegraphics[width=\textwidth]{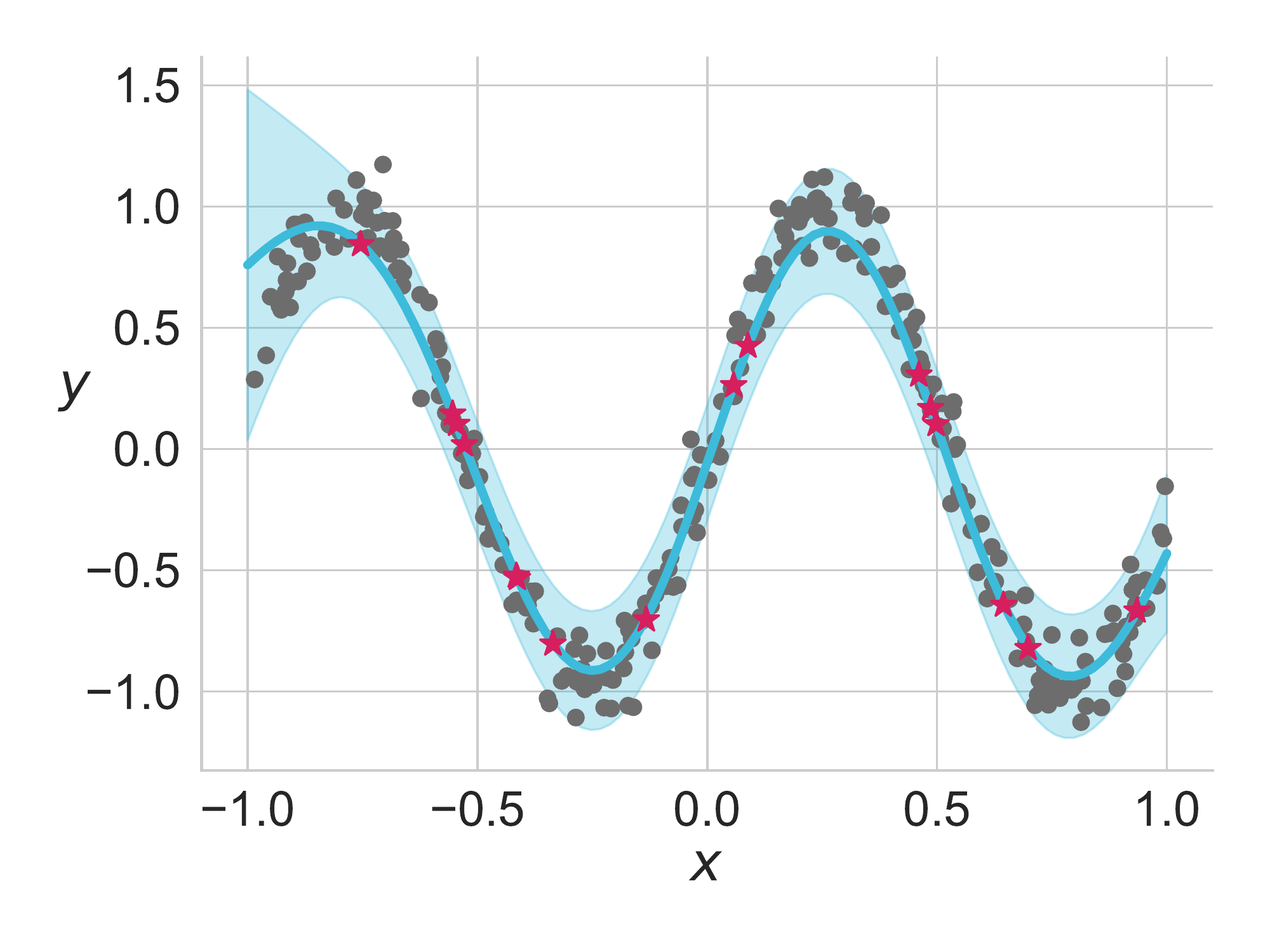}
    \end{subfigure}    
    \begin{subfigure}{0.19\textwidth}
    \includegraphics[width=\textwidth]{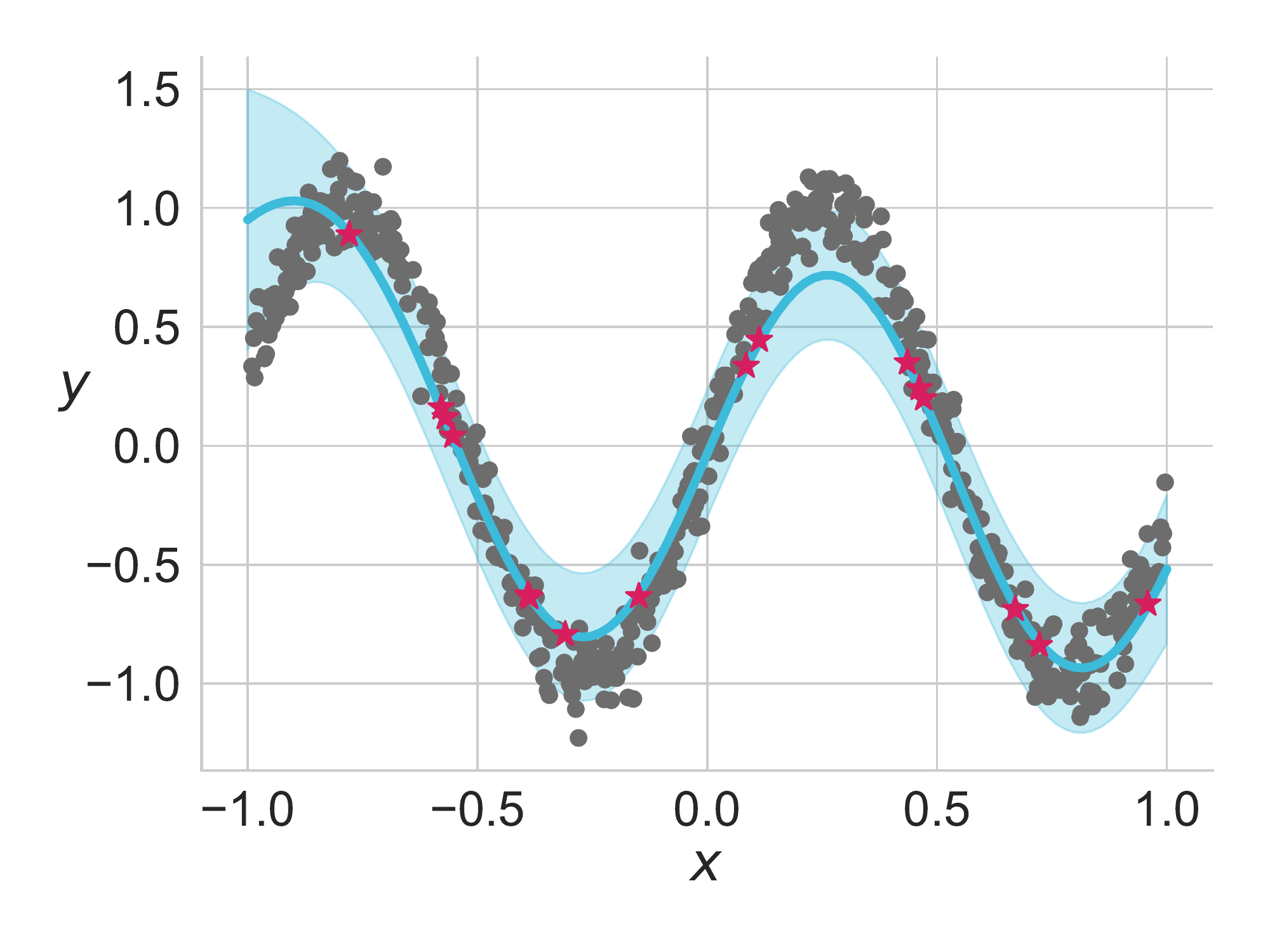}
    \end{subfigure}    
    \begin{subfigure}{0.19\textwidth}
    \includegraphics[width=\textwidth]{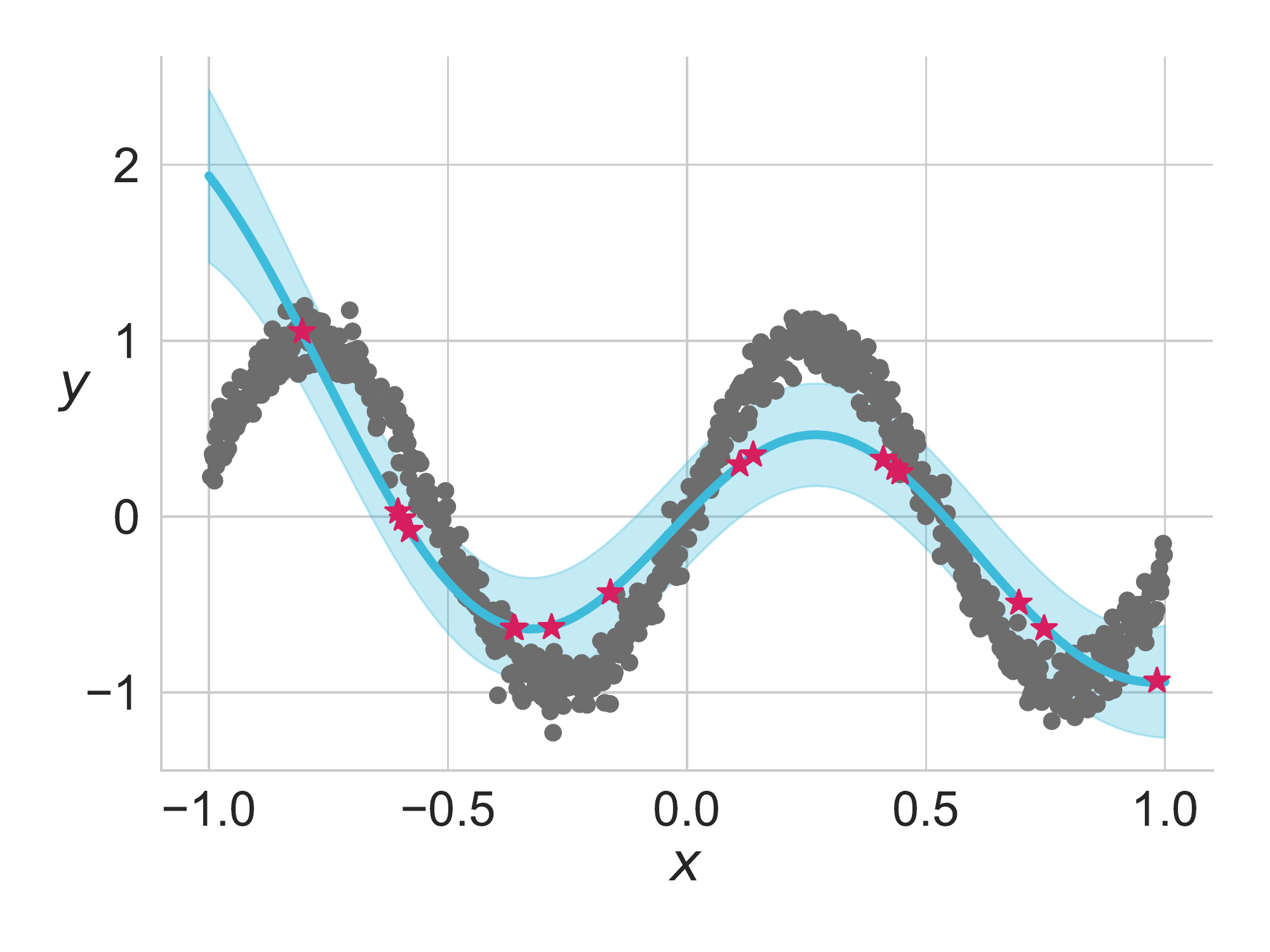}
    \end{subfigure}    
    \begin{subfigure}{0.19\textwidth}
    \includegraphics[width=\textwidth]{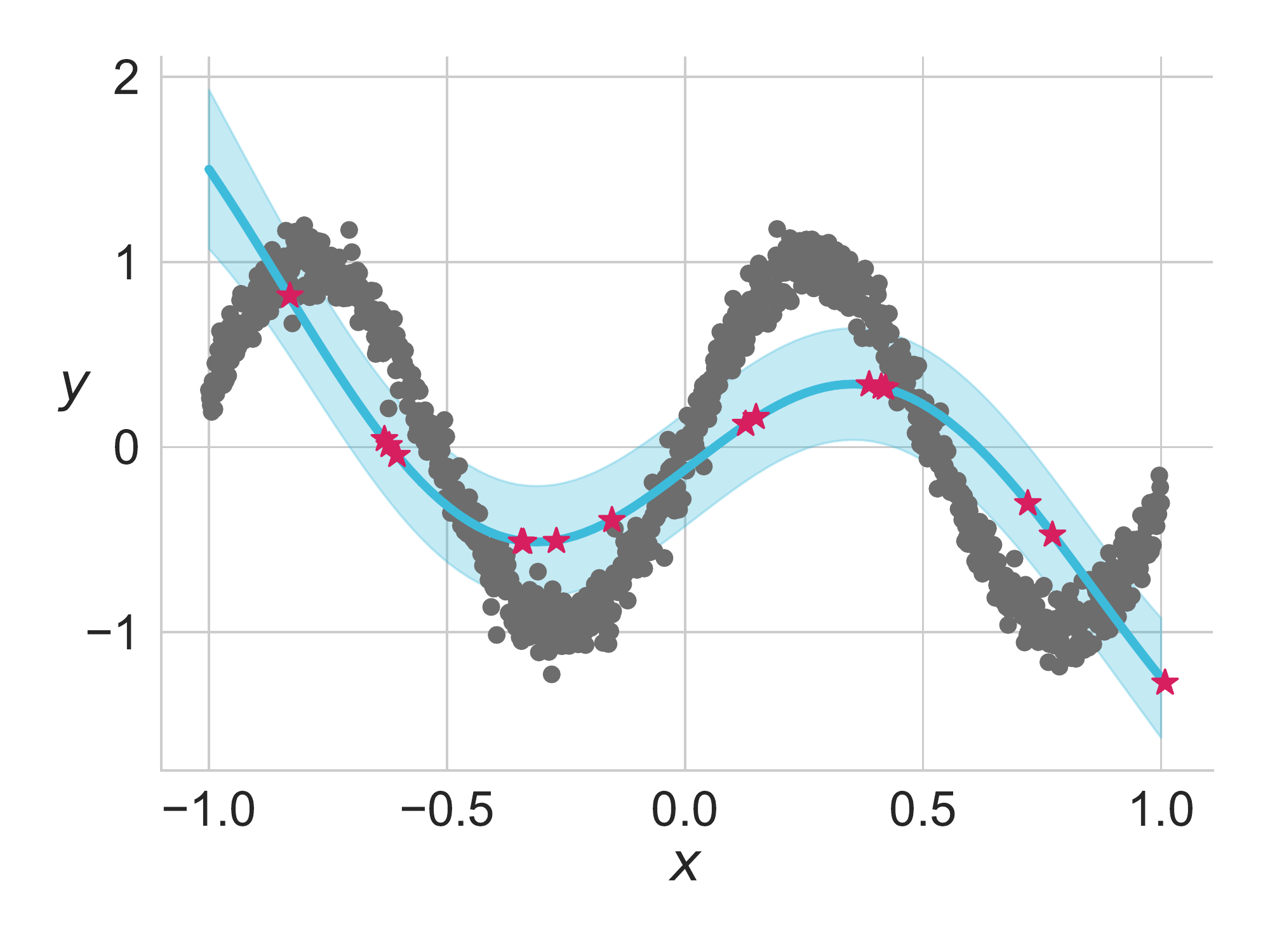}
    \end{subfigure}   
    \begin{subfigure}{0.19\textwidth}
    \includegraphics[width=\textwidth]{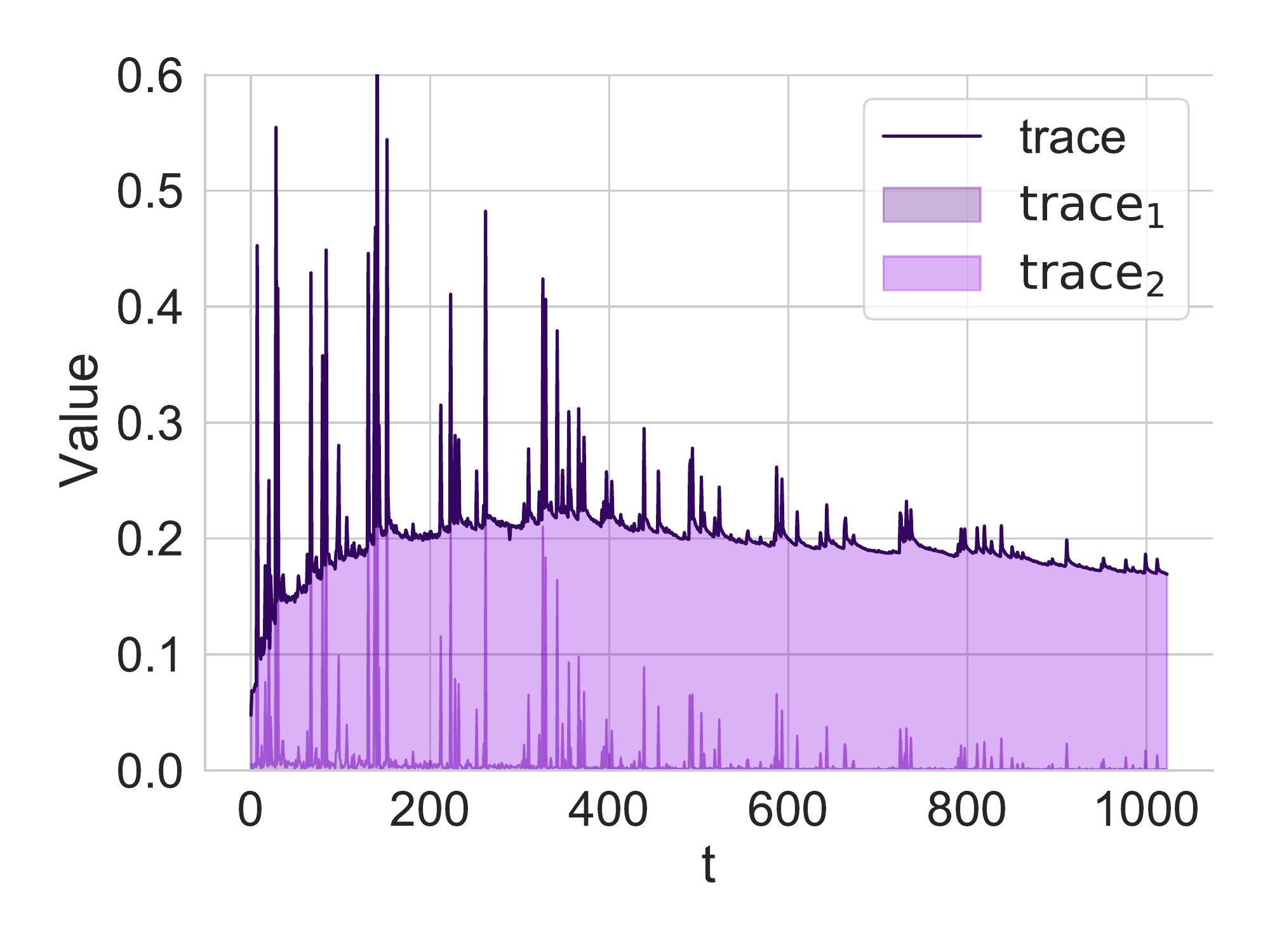}
    \end{subfigure}   
    \begin{subfigure}{0.19\textwidth}
    \includegraphics[width=\textwidth]{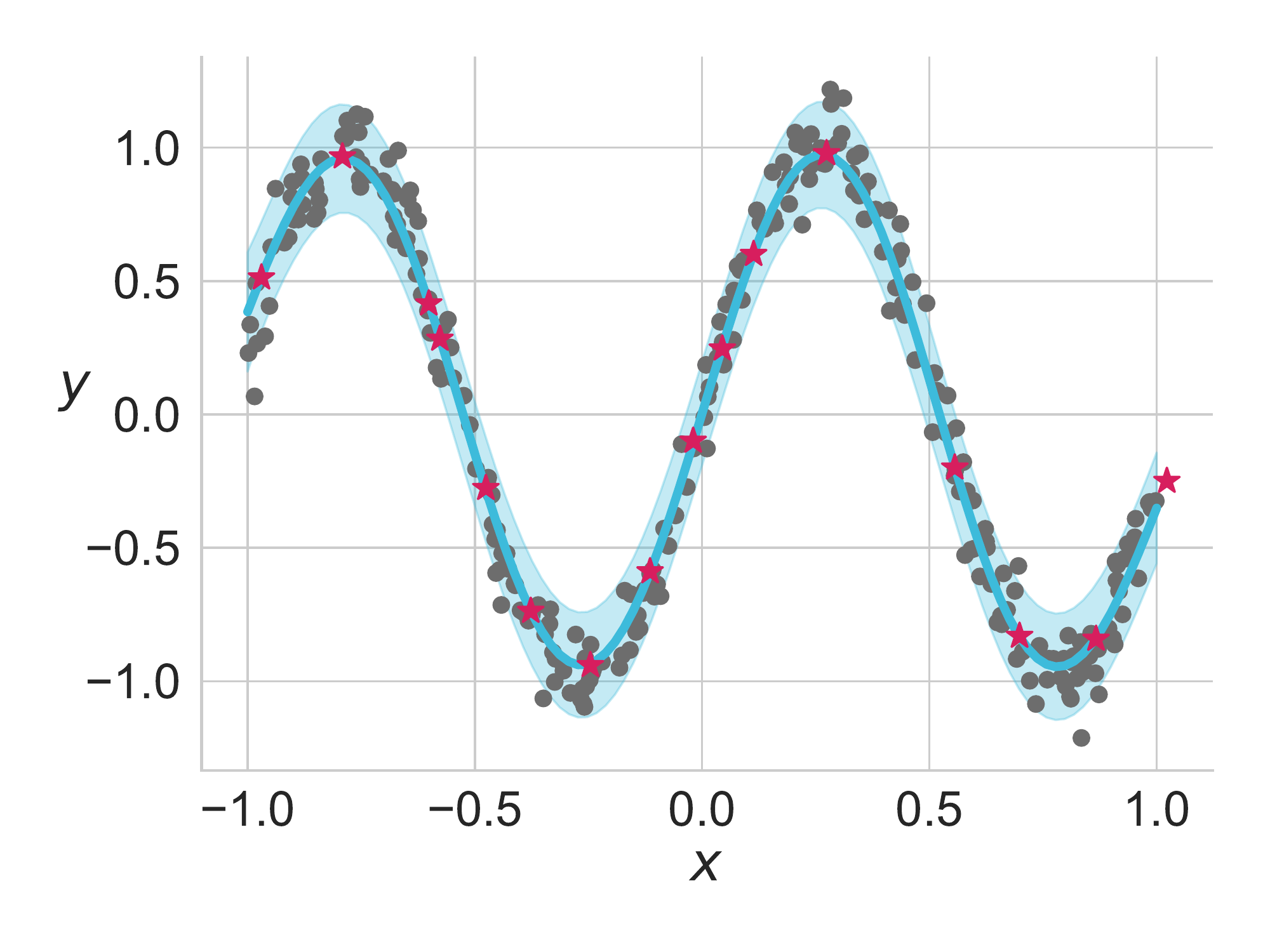}
    \caption{$T=256$}
    \end{subfigure}    
    \begin{subfigure}{0.19\textwidth}
    \includegraphics[width=\textwidth]{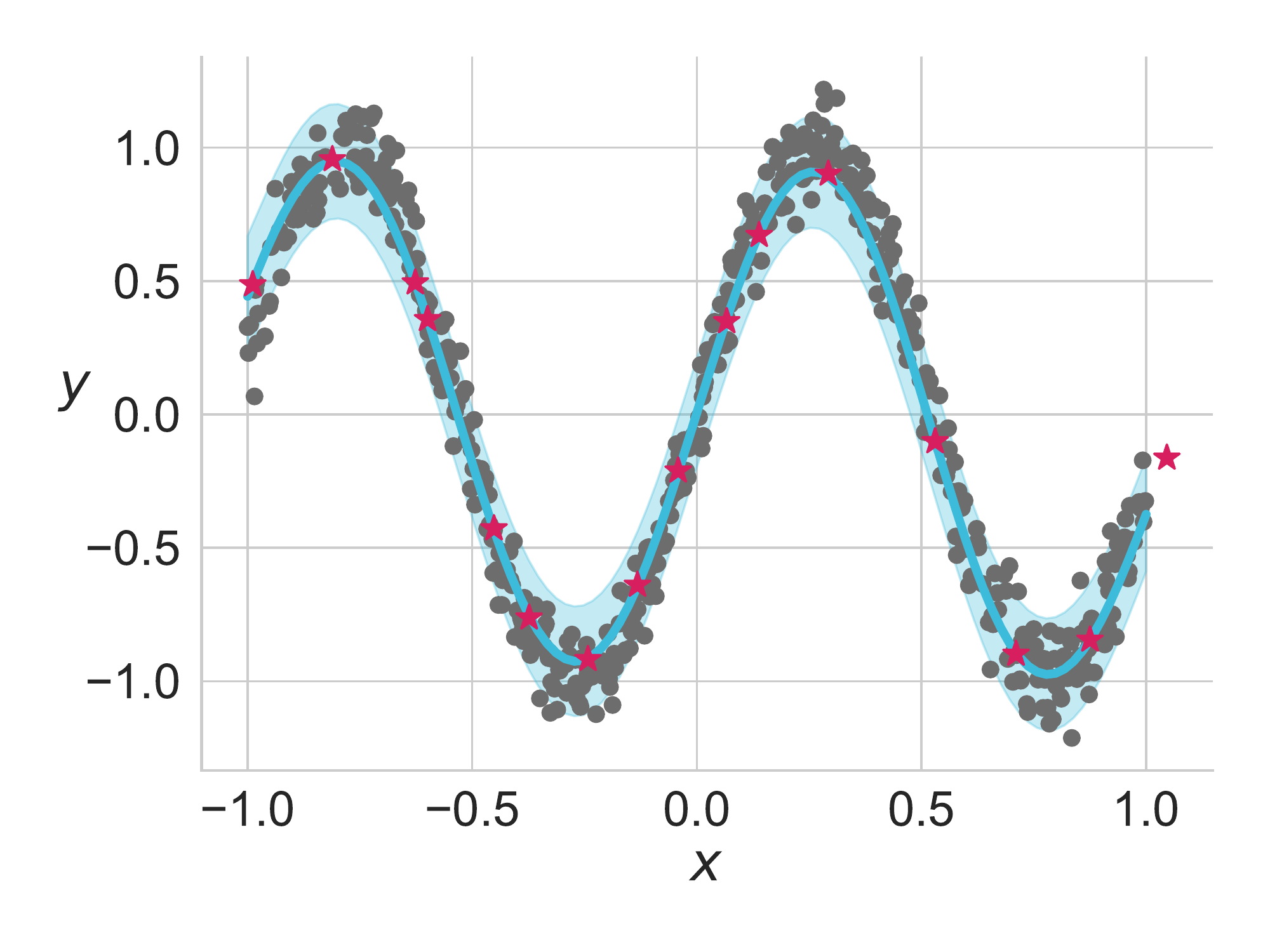}
    \caption{$T=512$}
    \end{subfigure}    
    \begin{subfigure}{0.19\textwidth}
    \includegraphics[width=\textwidth]{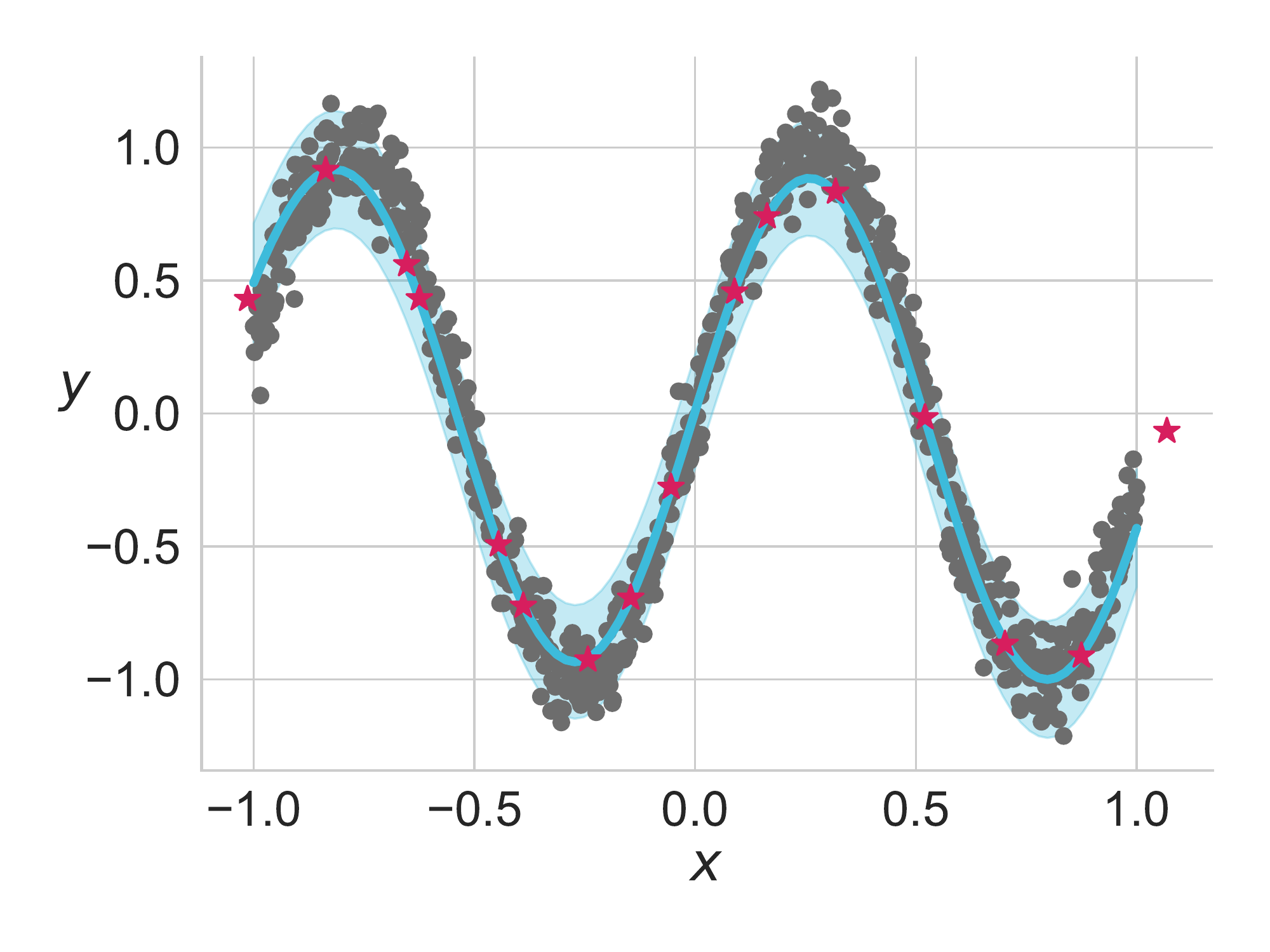}
    \caption{$T=768$}
    \end{subfigure}    
    \begin{subfigure}{0.19\textwidth}
    \includegraphics[width=\textwidth]{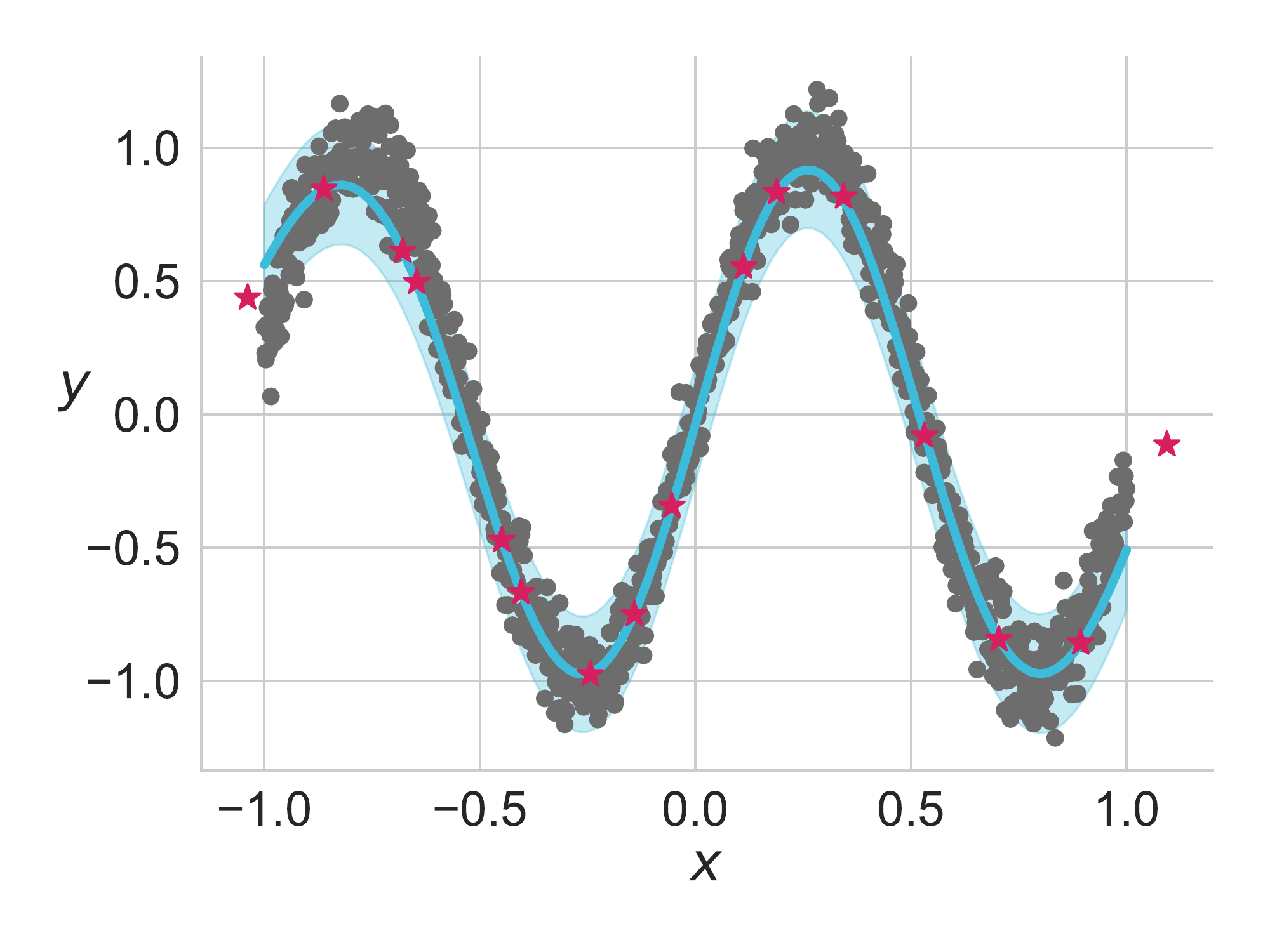}
    \caption{$T=1024$}
    \end{subfigure}
    \begin{subfigure}{0.19\textwidth}
    \includegraphics[width=\textwidth]{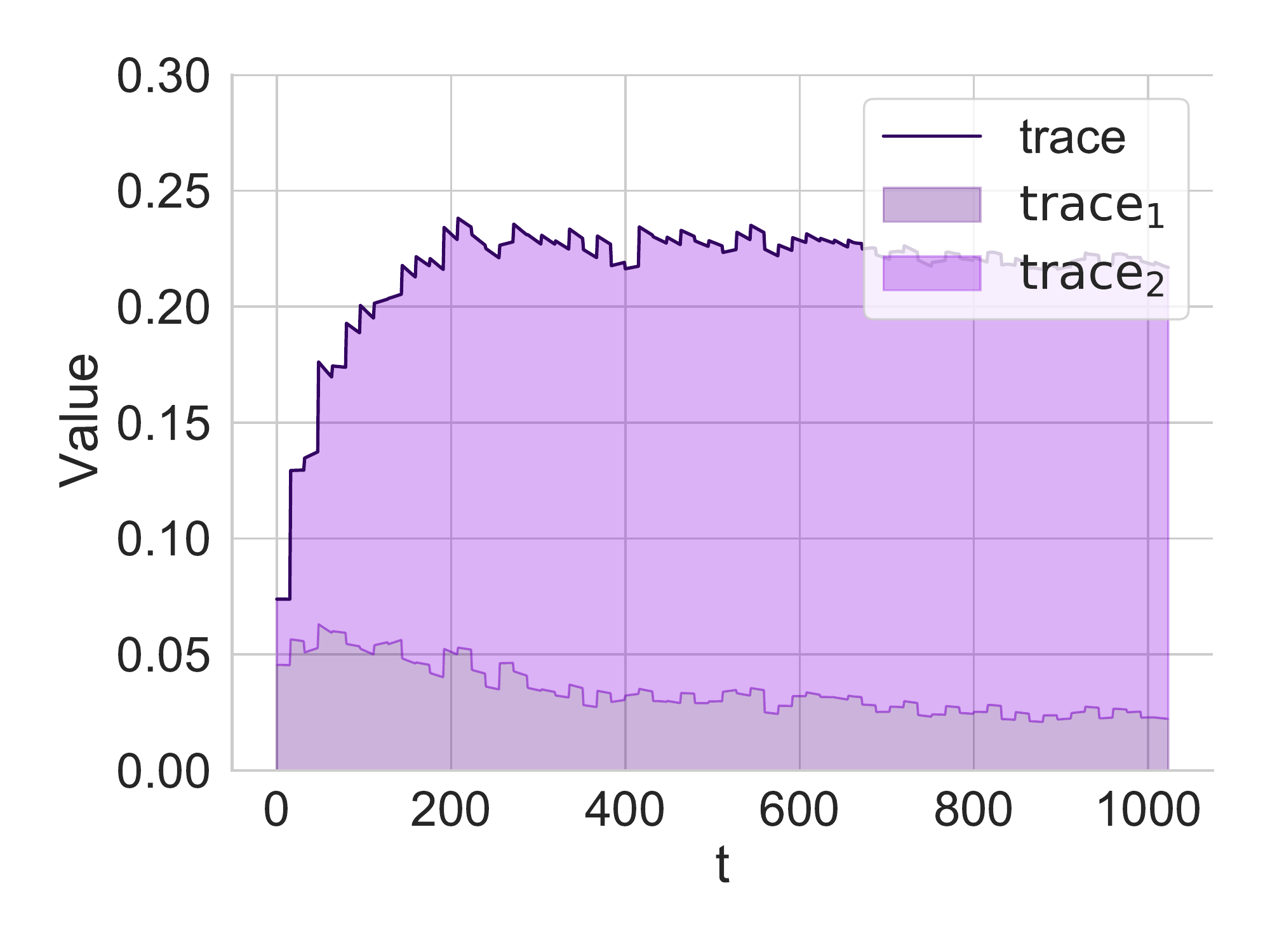}
    \caption{Trace term}
    \end{subfigure}
    \caption{O-SGPR ($p = 16$) learning on i.i.d observations without re-sampling the inducing points from a noisy sine function. \textbf{Top row: } As the number of data points increases for a batch size of $1$, the model progressively underfits due to excess regularization. The trace term is entirely dominated by $\text{trace}_2.$ \textbf{Bottom row: } For a larger batch size ($b = 16$), there is no under-fitting and $\text{trace}_1$ takes up more of the overall trace term. }
    \label{supp/fig:osgpr_sine}
\end{figure*}

In Figure \ref{supp/fig:osgpr_sine}, we empirically demonstrate the pathology of the O-SGPR bound with small batch sizes in the i.i.d. setting. In the top row, we add $b = 1$ data point at a time while continuing to re-train. 
Although the O-SGPR model originally fits the data well at $T = 256,$ it progressively begins underfitting, which becomes more and more noticeable, especially by $T = 1024.$
By comparison, a larger batch size, $b = p = 16,$ prevents any under-fitting from occurring. 
In the far right panel, we see the two terms in the trace component; in the small batch setting, the inducing trace term dominates the total trace.
The effect is mediated by a larger batch size, as shown in the bottom right panel.

Informally, if the number of old inducing points is much greater than the number of new observations, then O-SGPR will focus on replicating the old variational distribution.
Note that either aggregating multiple batches for each update or conditioning O-SGPR into an exact GP remedies the issue.

\subsection{Look-Ahead Thompson Sampling}
\label{subsec:look_ahead_ts_defn}

\begin{algorithm}[t!]
	\caption{\label{alg:ts_rollout_svgp} LTS with SVGP}
	\begin{algorithmic}
		\STATE \textbf{Input:} Observed data $\mathcal{D} = (X, \vec y);$ SVGP model, $\mathcal{M};$ candidate set generation utility, $C_{\text{gen}}(),$ rollout steps, $T,$ parallel path parameter $l,$ top $k$ parameter, $q$ batch size.
		\STATE \textbf{Output: } Candidates for evaluation $\mathbf{\bar X}_{\text{end}}.$
		\vspace{0.2cm}
		\hrule
		\vspace{0.2cm}
		\STATE 1. Generate initial candidate set, $X_1 = C_{\text{gen}}().$
		\STATE 2. Compute posterior over candidate set, drawing a posterior sample: $y_1 \sim p(y | X_1, \mathcal{M}).$
		\STATE 3. Sort $y_1$ and keep top $l$ samples $\tilde y_0$ and corresponding candidates, $\tilde X_1.$
		\STATE 4. Generate $\mathcal{M}_1 \leftarrow \text{OVC}(M, (\tilde X_i, \tilde y_i)\text{.unsqueeze(-1)})$ via Algorithm 1. $\mathcal{M}_1$ is a batch of $l$ models each conditioned on a single data point.
		\vspace{0.1cm}
		\FOR{t in 2:T}
        		\STATE 5. Generate initial candidate set, $X_t = C_{\text{gen}}().$
		    \STATE 6. Compute posterior over candidate set, drawing a posterior sample: $y_t \sim p(y | X_t, \mathcal{M}_{t-1}).$
		    \STATE 7. Sort $y_t$ and keep top $l$ samples $\tilde y_t$ and corresponding candidates, $\tilde X_t.$
		    \STATE 8. Generate $\mathcal{M}_{t} \leftarrow \text{OVC}(\mathcal{M}_{t-1}, (\tilde X_t, \tilde y_t))$ using Algorithm \ref{alg:model_conditioning}.
		\ENDFOR
		\STATE 9. Generate $\mathcal{M}_{\text{end}}\leftarrow \text{OVC}(\mathcal{M}, (\tilde X_i, \tilde y_i)_{i=1}^T)$ using Algorithm \ref{alg:model_conditioning}.
		\STATE Generate final candidate set, $X_{\text{end}} = C_{\text{gen}}().$
		\STATE 10. Compute posterior over candidate set, drawing a posterior sample: $y_{\text{end}} \sim p(y | X_{\text{end}}, \mathcal{M}).$
		\STATE 11. Sort $y_{\text{end}}$ and return top $q$ corresponding candidates, $\mathbf{\bar X}_{\text{end}}.$
		
	\end{algorithmic}
\end{algorithm}

\textbf{Thompson sampling in continuous domains}: in the context of black-box optimization, the action space is simply the input space, $\mathcal{X}$, since we are deciding which input $\vec x \in \mathcal{X}$ we will query next. 
Thompson sampling draws the next query point from the Bayes-optimal distribution over the possible choices, $\vec x^* \sim p_{\mathrm{TS}}(\vec x)$, where \begin{align}
    p_{\mathrm{TS}}(\vec x) \propto \int \mathbbm{1}\{f(\vec x) = \sup\limits_{\vec x' \in \mathcal{X}} f(\vec x') \} p(f | \mathcal{D}) df. \label{eq:thompson_sampling_dist}
\end{align}
When $\mathcal{X}$ is a continuous domain, as is usually the case, we replace the $\sup_{\vec x' \in \mathcal{X}}$ with a $\max_{\vec x' \in X_{\mathrm{cand}}}$, where $X_{\mathrm{cand}} \subset \mathcal{X}$. 
As the name suggests, rather than attempting to evaluate the integral in Eq. \eqref{eq:thompson_sampling_dist}, Thompson sampling instead draws samples $f_i \sim p(f | \mathcal{D})$, $i \in \{1, \dots, q\}$ and take $\vec x_i^* = \argmax_{\vec x' \in X_{\mathrm{cand}}} f_i(\vec x')$.

\textbf{The look-ahead case}:
If we were able to evaluate each $f_i$ on every point $\vec x \in \mathcal{X}$ and compute $\sup_{\vec x' \in \mathcal{X}} f_i(\vec x')$ exactly, there would be no benefit to multiple rounds of Thompson sampling. 
However, as we noted above, typically we rely on a $\max$ over a discrete set $X_{\mathrm{cand}}$, typically obtained from a (quasi-)Monte Carlo method (e.g. Sobol sequences) to cover $\mathcal{X}$, and the number of candidate points is restricted by compute and memory. 
Therefore we can do multiple rounds of Thompson sampling to try to refine the estimate of $\vec x_i^*$ by evaluating $\argmax_{\vec x' \in X_j} f_i(\vec x')$ for a sequence of candidate sets $X_j$, $j \in \{0, \dots, h\}$.
The key challenge with GPs is to ensure that $f_i$ is consistent across the sequence of candidate sets, which we accomplish by drawing $f_i(\vec x') \sim p(f | \mathcal{D} \cup \{(\vec x^*_{j-1}, f_i(\vec x^*_{j-1}))\}$ for $\vec x' \in X_j$ and $j > 0$.
The result is again $\vec x^*_i = \max_{\vec x' \in X_{\mathrm{cand}}} f_i(\vec x')$, but now $X_{\mathrm{cand}} = \bigcup_j X_j$.

In Algorithm \ref{alg:ts_rollout_svgp}, we describe how OVC is used within LTSs as an example of its usecase. 
Here, of course, we are only performing Thompson sampling over a discrete set of values and so do not end up needing to use gradient based acquisitions.
Specifically, we continue using OVC (or really after $T = 1$, exact GP conditioning with low-rank updates \citep{jiang_efficient_2020} as the GP is now exact), to condition our model on each step's fantasy responses $\tilde y_t$ and the observations $\tilde X_{\mathrm{batch}}.$

\section{Further Experimental Details and Results}
\label{app:exp_data}
\subsection{Updating O-SGPR Inducing Points}

\begin{figure*}[h!]
	\begin{subfigure}{\textwidth}
		\centering
		\includegraphics[width=0.8\textwidth,clip,clip,trim=0cm 0cm 0cm 10cm]{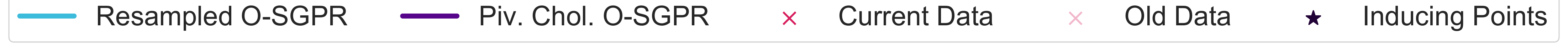}
	\end{subfigure}
\centering
\begin{subfigure}{0.31\textwidth}
\centering
\includegraphics[width=\linewidth]{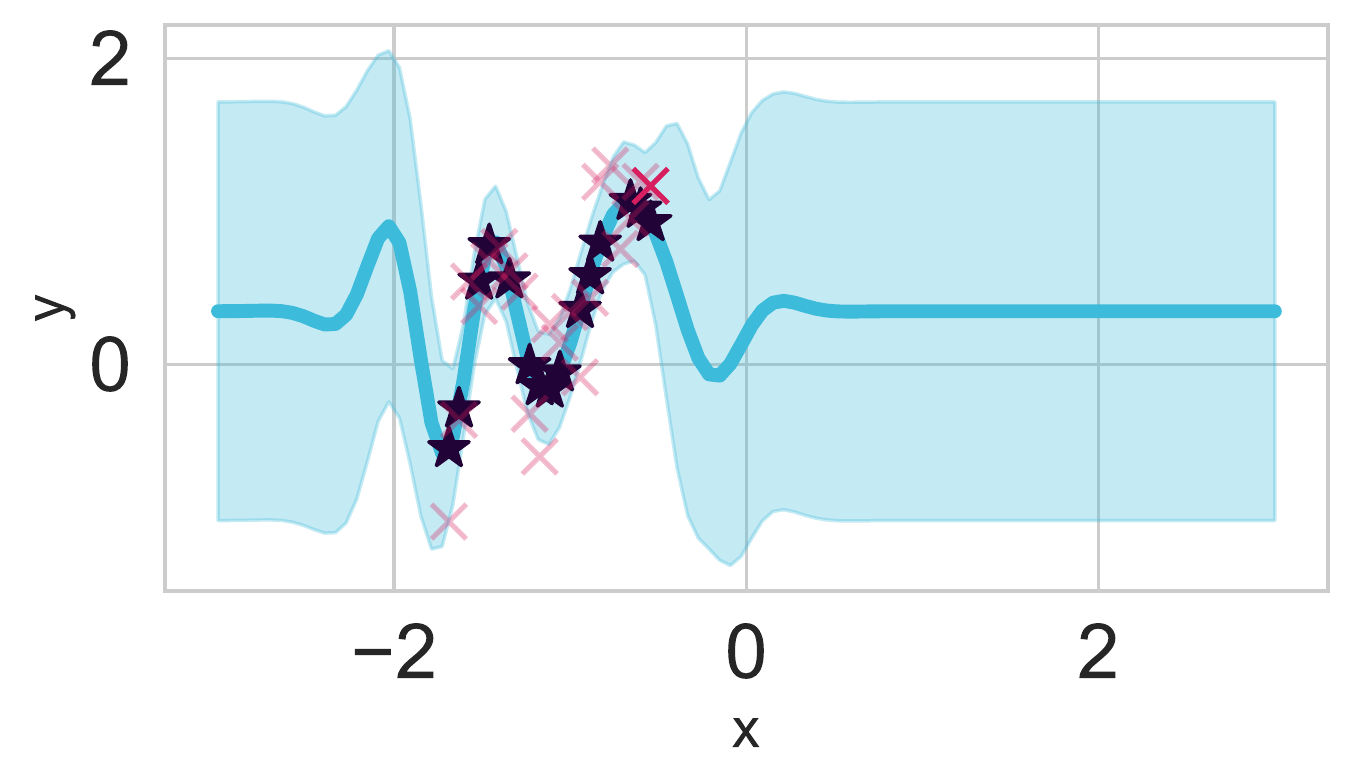}
\end{subfigure}
\begin{subfigure}{0.31\textwidth}
\centering
\includegraphics[width=\linewidth]{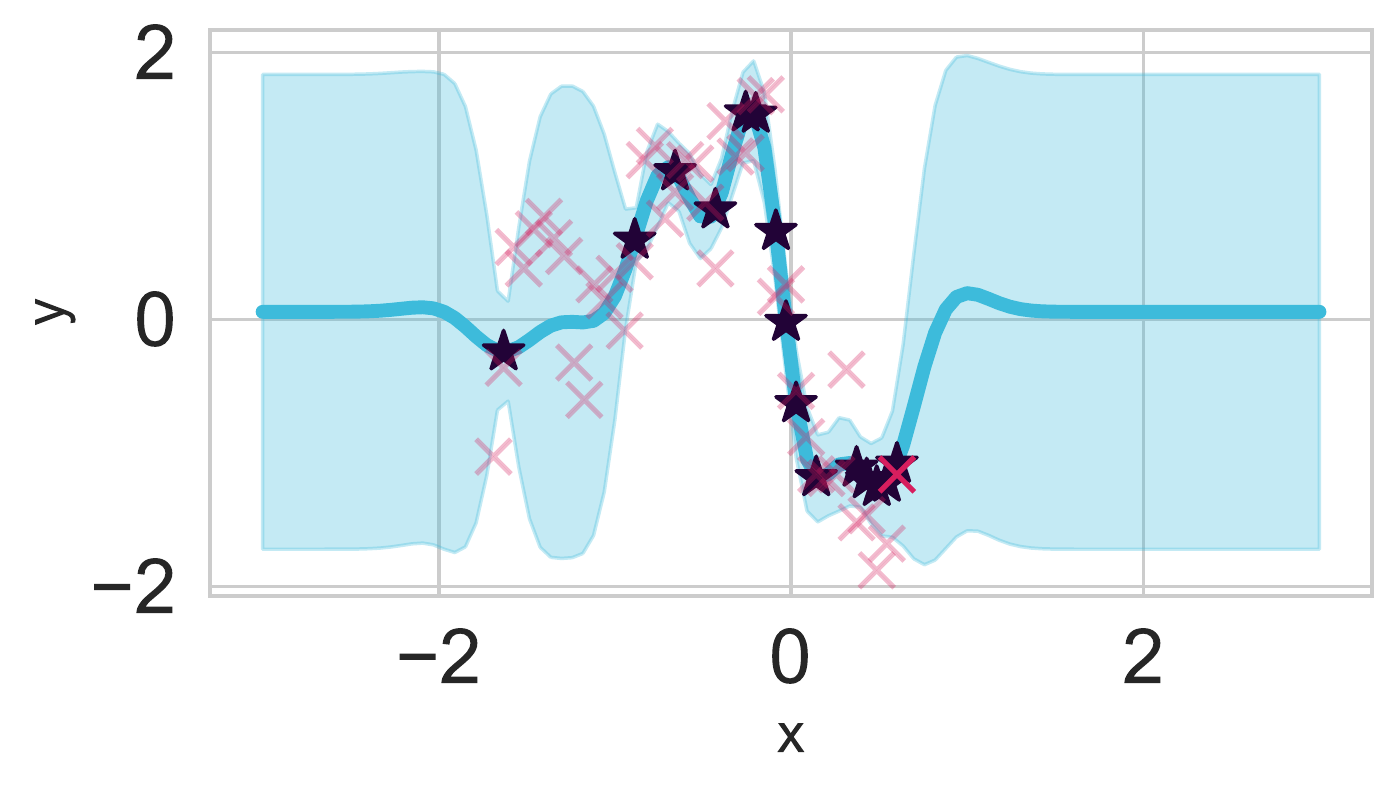}
\end{subfigure}
\begin{subfigure}{0.31\textwidth}
\centering
\includegraphics[width=\linewidth]{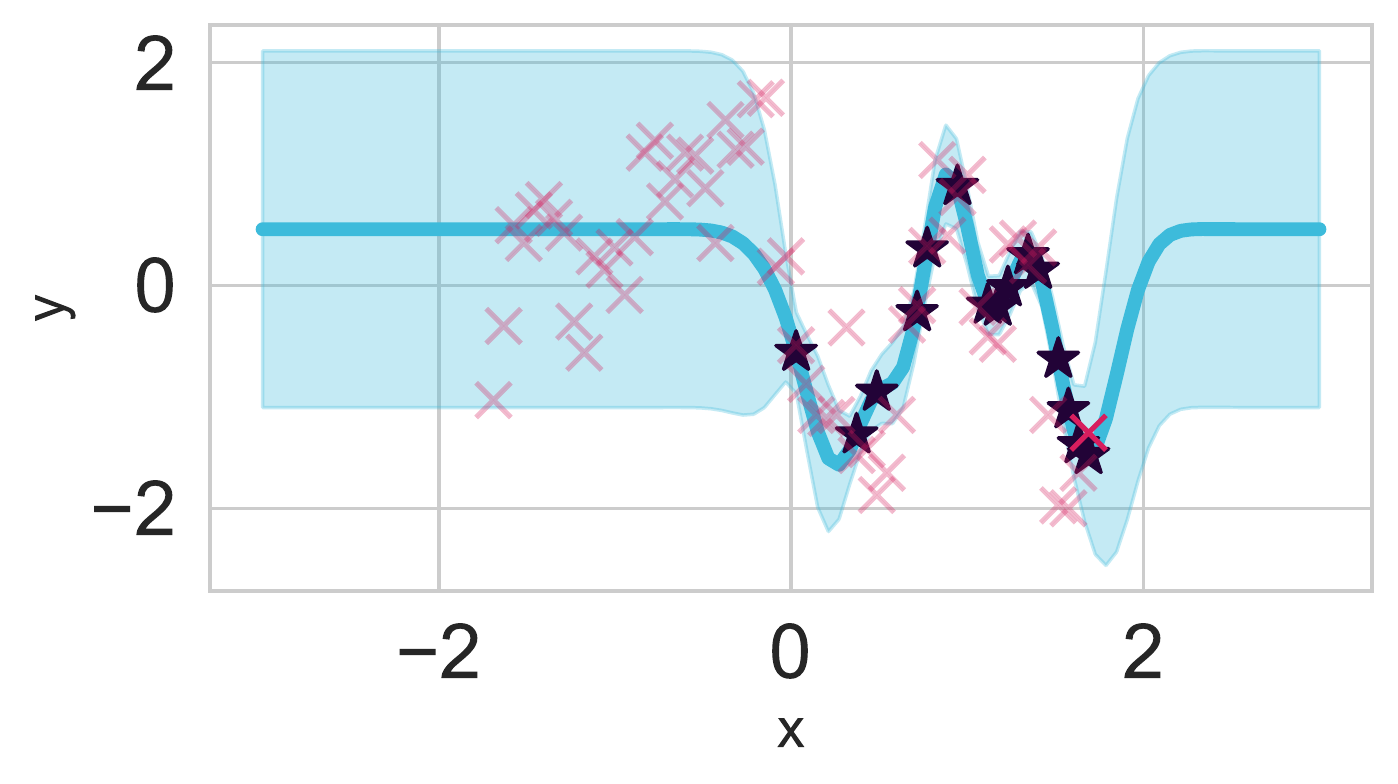}
\end{subfigure}
\begin{subfigure}{0.31\textwidth}
\centering
\includegraphics[width=\linewidth]{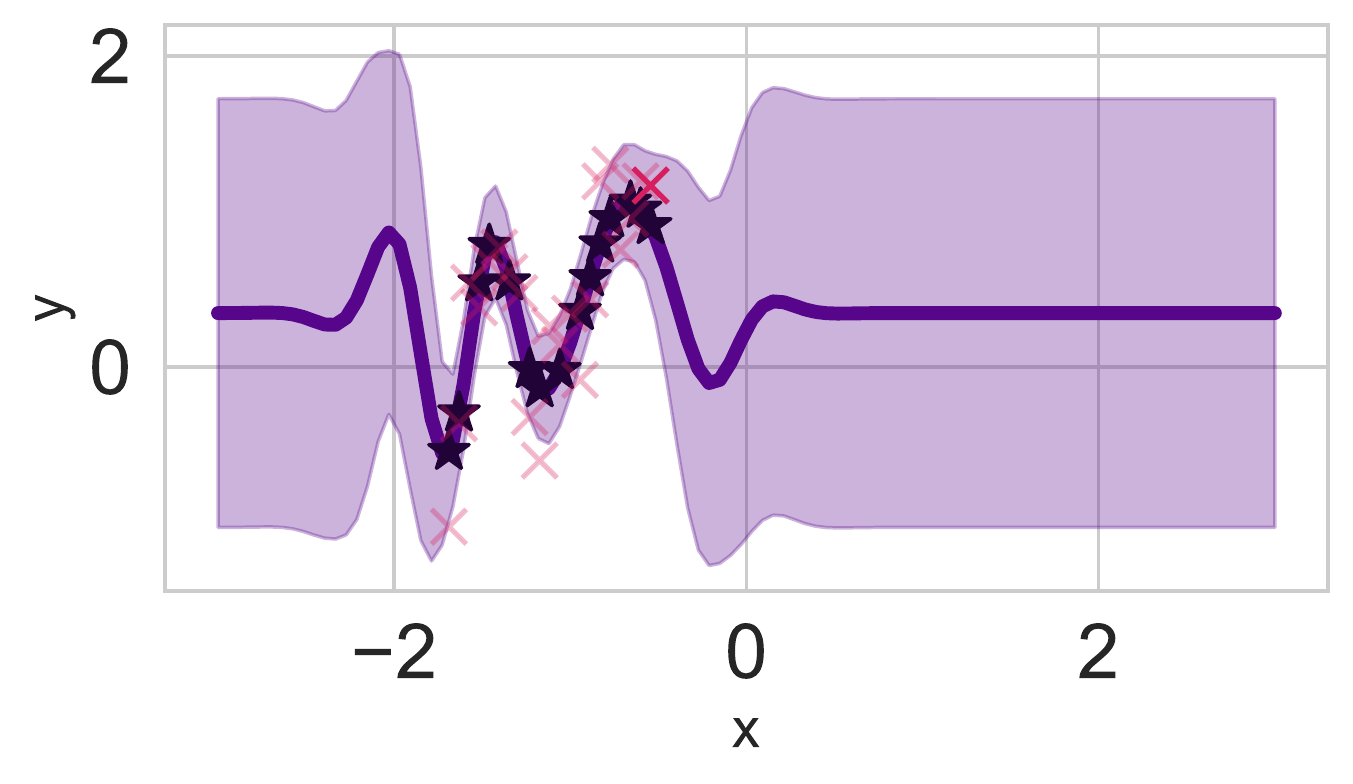}
\caption{$t = 20$}
\end{subfigure}
\begin{subfigure}{0.31\textwidth}
\centering
\includegraphics[width=\linewidth]{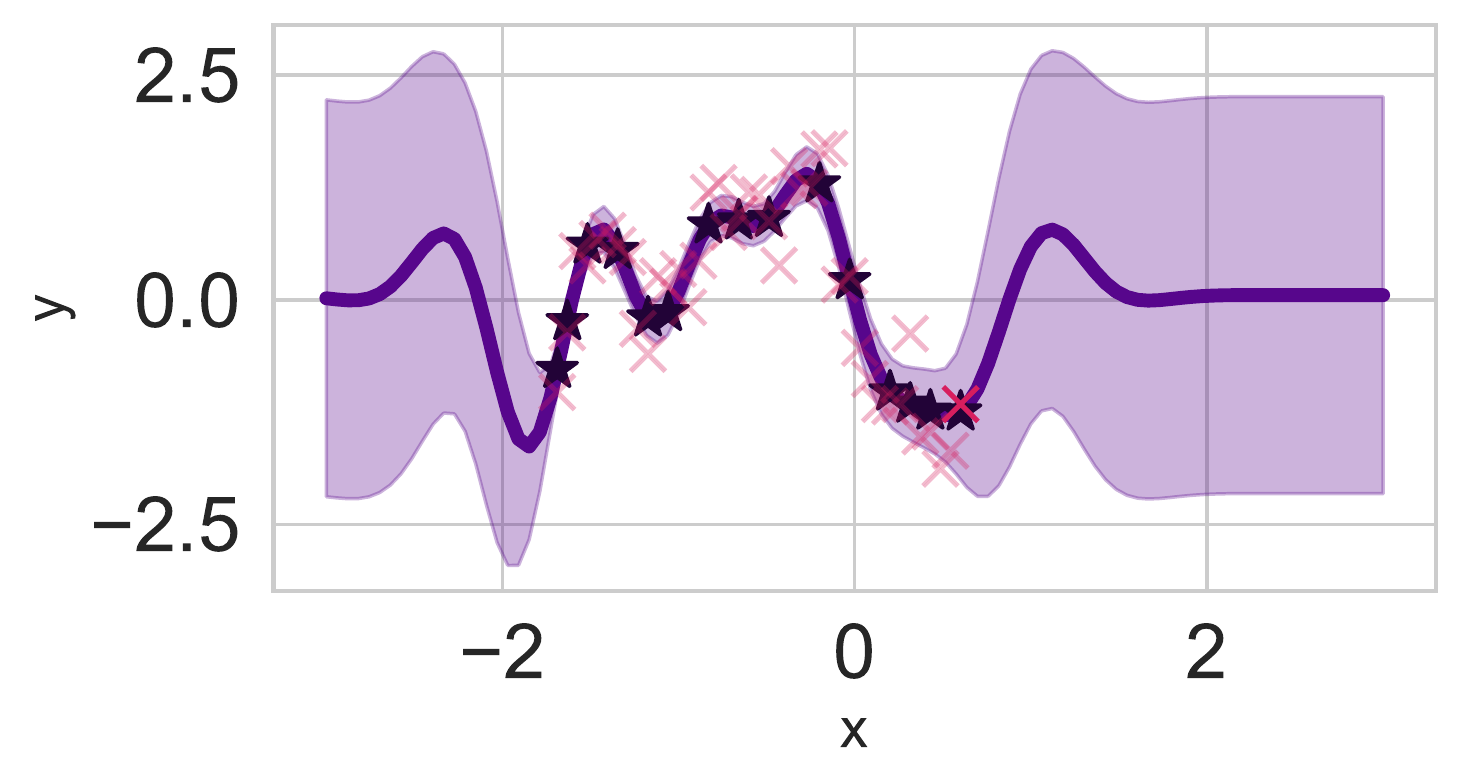}
\caption{$t=40$}
\end{subfigure}
\begin{subfigure}{0.31\textwidth}
\centering
\includegraphics[width=\linewidth]{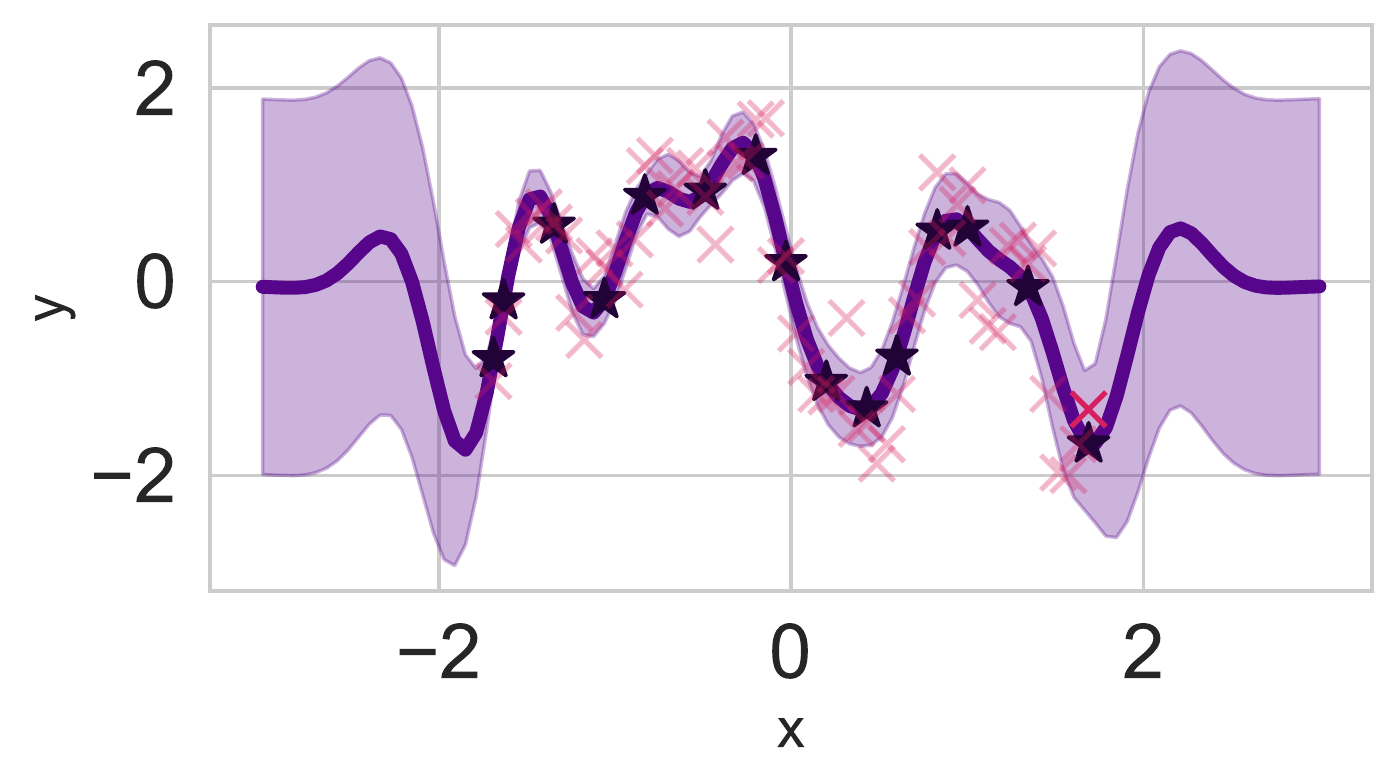}
\caption{$t=60$}
\end{subfigure}
\caption{Online SVGP modelling a time series \textbf{Top row:}  inducing points are updated by replacing an inducing point with the new location as in \citet{bui_streaming_2017}. \textbf{Bottom row:} Inducing points are updated by re-running a pivoted cholesky on both the new data point and the current inducing points. The recursive refitting procedure of the pivoted cholesky placement remedies the catastrophic overfitting and forgetting of merely resampling the inducing points.}
\label{fig:svgp_time_series_conditioning}
\end{figure*}

We illustrate the efficacy of this choice of new inducing points in Figure \ref{fig:svgp_time_series_conditioning} using the same time series data as in \citet{stanton_kernel_2021} originally from \url{https://raw.githubusercontent.com/trungngv/cogp/master/data/fx/fx2007-processed.csv} (that repo uses BSD License). 
Re-sampling the old inducing points is shown in the top row, and tends to first perform well, but then begins to catastrophically forget by $t = 40$ and dramatically so by $t = 60,$ as all of the inducing points have moved over to the right.
By comparison, our approach of iteratively running a pivoted cholesky on the current inducing points and the new data point, prevents catastrophic forgetting, while also enabling the model to learn on the new data stream.

\begin{figure}[h!]
\begin{subfigure}{\textwidth}
\raggedright
\includegraphics[width=0.93\linewidth]{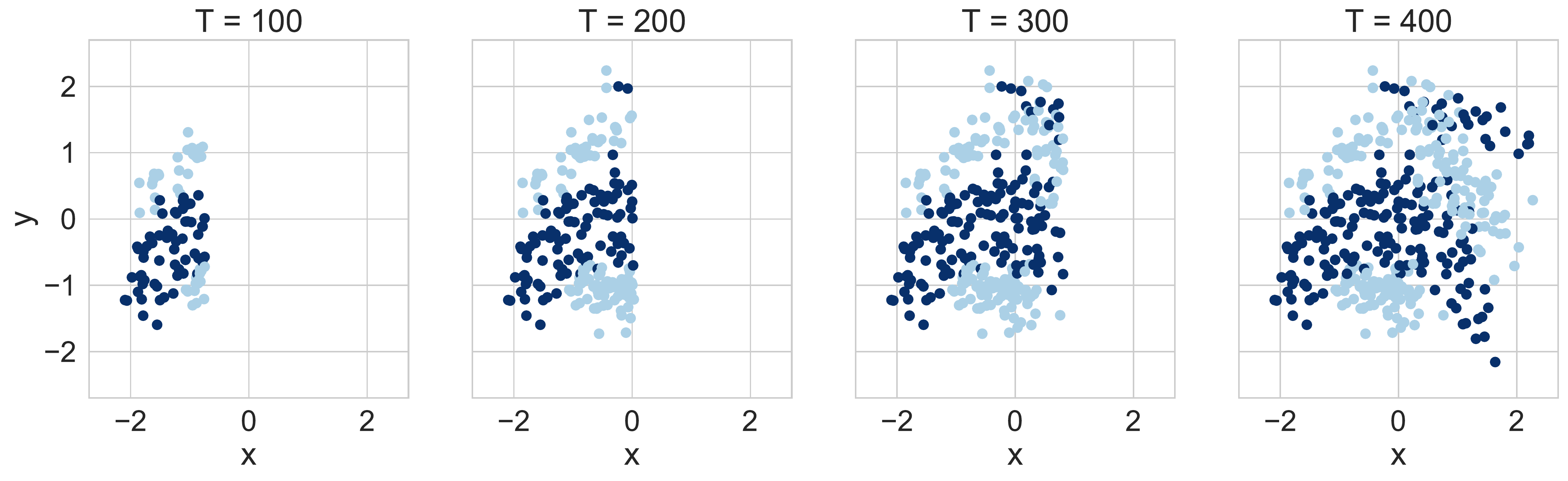}
\end{subfigure}
\begin{subfigure}{\textwidth}
\centering
\includegraphics[width=\linewidth]{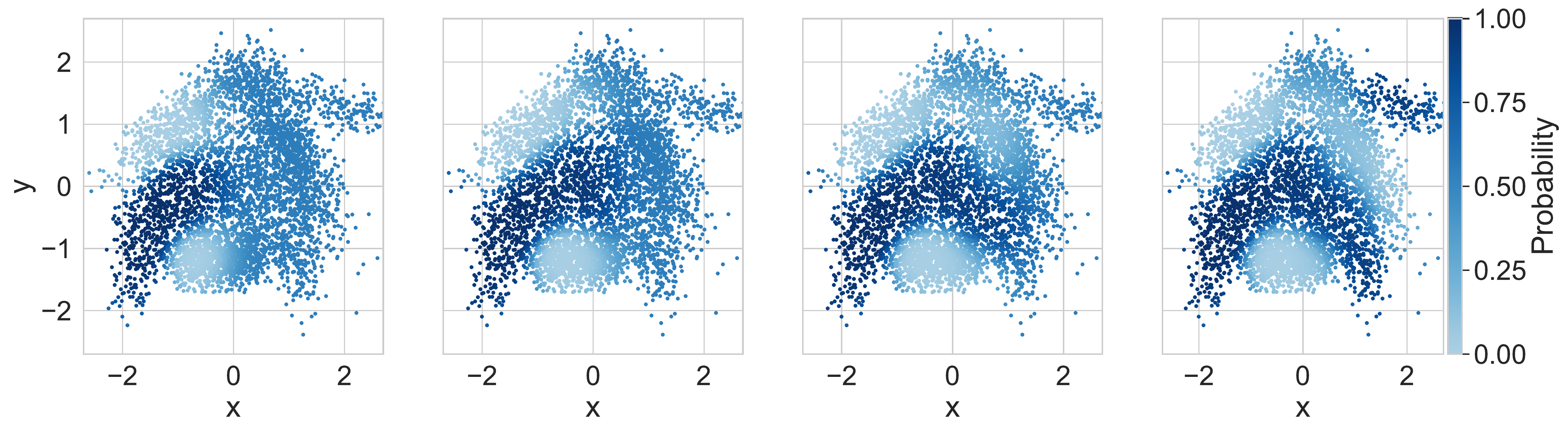}
\end{subfigure}
\begin{subfigure}{\textwidth}
\centering
\includegraphics[width=\linewidth]{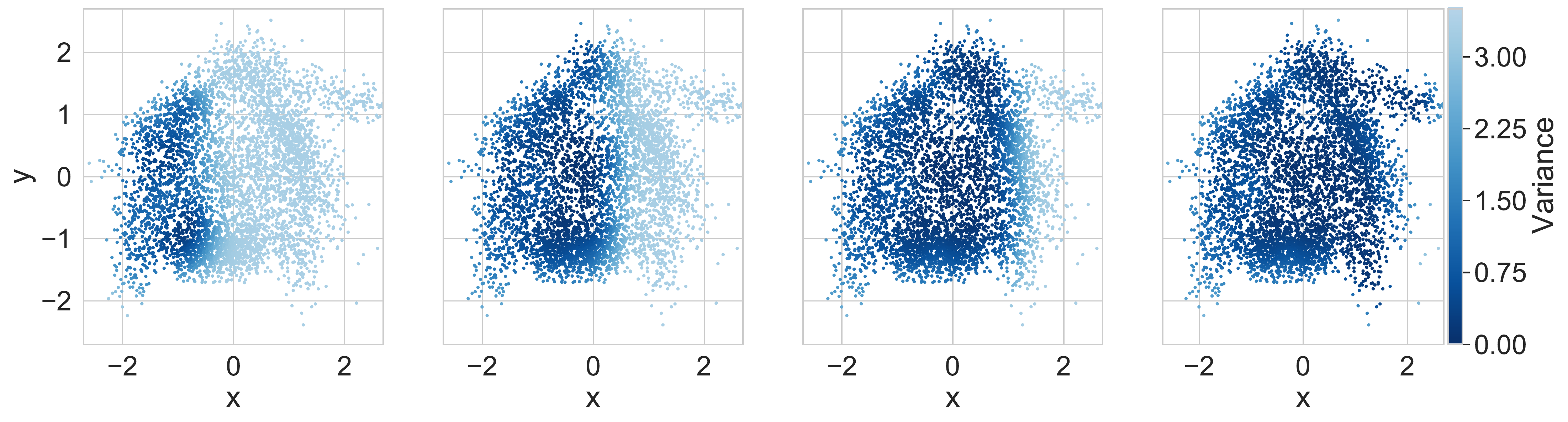}
\end{subfigure}
\caption{\textbf{Top row:} Data from the bananas dataset arriving in a non-i.i.d fashion in four sucessive batches. 
\textbf{Middle row:} Predictive probabilities of a SVGP with OVC to rollout conditional on these batches. Even in the non-Gaussian setting, OVC is able to adapt to new data without catastrophic forgetting.
\textbf{Bottom row:} Variance of the latent function during the OVC rollout. The variances decay as we observe new data.
}
\label{fig:bananas_rollout}
\end{figure}

In Figure \ref{fig:bananas_rollout}, we illustrate the effect of Laplace approximations during a rollout following the streaming classification example of \citet{bui_streaming_2017} as we use OVC.
We first trained a SVGP model with $25$ inducing points on $100$ data points, as shown in the first rows, then performed three steps of rollouts each with $100$ data points each as we observe progressively more and more of the dataset. 
In the middle row, we show the predicted probability on a held-out test set; as we observe more data, the predictions become more and more confident throughout the entire region, and are un-confident in the regions where we have not observed any data.
This effect is similarly observed by the predictive variances, which are high in regions where we have not seen any data, but decay as we observe each region successively.
Data from \url{https://github.com/thangbui/streaming_sparse_gp/tree/master/data} (Apache 2.0 License).

\subsection{Experimental and Data Details}\label{app:data}

Unless otherwise specified, all data is simulated.
The code primarily relies on PyTorch \citep{paske2019pytorch} (MIT License), BoTorch \citep{balandat_botorch_2020} (MIT License), GPyTorch \citep{gardner_gpytorch:_2018} (MIT License).
All GPs (variational and exact) used a constant mean, scaled Matern-$5/2$ kernels with ARD with lengthscale priors of $\texttt{Gamma}(3,6)$ and outputscale priors of $\texttt{Gamma}(2, 0.15),$ which are the current BoTorch defaults for single task GPs.
All variational GPs used GPyTorch's whitened variational strategy.
Unless otherwise specified, we normalized all inputs to $[0,1]^d$ and standardized outputs to have zero mean and standard deviation one during the model fitting stage.
We trained all models to convergence with an exponential moving average stopping rule using Adam with a learning rate of $0.1.$
We re-fit each model independently at each iteration.
When plotting the median, we plot the $95\%$ confidence interval around it following \url{https://www-users.york.ac.uk/\~mb55/intro/cicent.htm}.
Unless otherwise specified, all acquisitions were optimized with multi-start L-BFGS-B with $10$ random restarts and $512$ samples for initialization for up to $200$ iterations with a batch limit of $5$ following \citet{balandat_botorch_2020}.

\subsubsection*{Understanding Experiments}
All understanding experiments, e.g. Figures \ref{fig:svgp_conditioning}, \ref{supp/fig:osgpr_sine}, \ref{fig:svgp_time_series_conditioning}, \ref{fig:bananas_rollout} were run on CPUs with Intel i5 processors.
Computational costs were negligible to the cost of writing this paper.

\textbf{Figure \ref{fig:svgp_conditioning}:} 
We fit each non-Gaussian model using the ELBO. The Gaussian function is $f(x) = \sin(2 |x| + x^2 / 2)$ with $n=100$ and $n_{\text{test}} = 25.$
For the GPCV model, we follow the model definition of \citet{wilson_copula_2010} and parameterize the scale of the Gaussian as a linear softplus transform, but we implemented a variational version, rather than the Laplace or MCMC implementations that are considered in the original paper.
The data itself is a forward simulation of the well known SABR volatility model \citep{hagan2002managing} with parameters $F_0 = 10,$ $V_0=0.2,$ $\mu = 0.2,$ $\alpha=1.5,$ $\beta = 0.9,$ $\rho = -0.2.$
We model the scaled log returns and plot volatility rather than the latent function.
We use $n=250$ and $n_{\text{test}} = 150,$ so that $T = 400;$
we standardize the inputs.
Here, to perform Laplace approximations, we used PyTorch's higher order AD software as deriving the gradients and Hessians would be tedious.

\textbf{Knowledge Gradient on Branin: }
We used the Branin test function as implemented in BoTorch \citep{balandat_botorch_2020} with $n=50$ and $25$ inducing points, $8$ fantasies per data point, $250$ candidate points and a grid of size $15 \times 15.$
These were run on a single Nvidia Titan $24$GB RTX.
Computation took several minutes.

\textbf{Incremental Learning on Protein: }
We followed the experimental protocol of \citet{stanton_kernel_2021} but substituted in Matern-$5/2$ kernels with ARD instead of linear projections.
The data comes from \citet{Dua:2019}.
The experiment is run over $10$ random seeds and we show the mean and two standard deviations of the mean.
Computation took several hours per trial.

\subsubsection*{Batch Knowledge Gradient Experiments}
qEI and qNEI optimization used quasi Monte Carlo (QMC) integration with $256$ random samples, while qKG optimization used QMC integration with $64$ random samples (BoTorch defaults).

\textbf{Hartmann6: }
Data comes from the Hartmann6 test function from \url{https://github.com/pytorch/botorch/blob/master/botorch/test_functions/synthetic.py} and the experiment is inspired by \url{https://botorch.org/tutorials/closed_loop_botorch_only}.
These were run on CPUs on an internal cluster.
Computation took several hours per trial. 
We used  and $1000$ randomly sampled candidate points to estimate the knowledge gradient.

\textbf{Laser: }
Comparison scripts are from \url{https://github.com/ermongroup/bayes-opt/}. No license was provided.
These were run on CPUs on an internal cluster.
Computation took several hours per trial.

\textbf{Preference Learning: }
Function is inspired by \url{https://botorch.org/tutorials/preference_bo}; we used noise of $\sigma = 0.1$ to make the function more difficult.
The Laplace implementation comes from \url{https://github.com/pytorch/botorch/blob/master/botorch/models/pairwise_gp.py}.
These were run on CPUs on an internal cluster.
Computation took several hours per trial.
qNEI optimization used QMC integration with $128$ random samples, while qKG optimization used QMC integration with $64$ random samples.
Here, we used $3$ random restarts and $128$ raw samples for acquisition function optimization.

\subsubsection*{Active Learning Experiments}
\textbf{Malaria: }
Data is originally from \citet{weiss2019mapping} under a creative commons 3 license, \url{https://malariaatlas.org/malaria-burden-data-download/#FAQ}.
From the reference, the data is modelling predictions off of survey data and thus not human responses.
For all models, we used Matern-$1/2$ kernels due to the lower smoothness and fixed noise models as variance is known.
Comparison is to WISKI \citep{stanton_kernel_2021}, with their code \url{https://github.com/wjmaddox/online_gp} which uses Apache License 2.0.
These were run on a combination of Nvidia $32$GB V100s and $48$GB RTXes on an internal cluster.
Here, we used $4$ random restarts with $64$ base samples to optimize the aquisition.
Computation took several hours per trial.

\textbf{Hotspot Modelling: }
Simulated data and comparison data is from \url{https://github.com/disarm-platform/adaptive_sampling_simulation_r_functions}. No license was provided for either.
Our trials were run on AMD $32$GB Mi50 GPUs on an internal cluster.
Computation took close to eight hours per trial.
We used tempering with $\beta = 0.1$ \citep{jankowiak2020parametric} and Matern-$3/2$ kernels for these models following the kriging setup in \citet{andrade2020finding}.
Overall, we used $16$ inner and outer samples for the entropy search objective, enumerating over all remaining test points to select a new point to query.

We note that the model fitting procedure of \citet{andrade2020finding} seems to possibly encourage test-set leakage as they seemingly use a random forest trained on all of the data, rather than on simply the first $n$ observations.
See the third from final paragraph in their description of spatial methods in that section.
We do not follow this as we do not use any random forests.

\subsubsection*{TurBO Experiments}

\textbf{Rover: }
We use the opensource Turbo implementation from \url{https://botorch.org/tutorials/turbo_1} and the rover function setup code from \url{https://github.com/zi-w/Ensemble-Bayesian-Optimization}.
Both are licensed under the MIT License.
These were run on Nvidia $24GB$ RTXes on an eight GPU server.
We repeated experiments $24$ times. 
See the wall-clock time panels for estimates of computational budgets (about an hour).
As not all trials reached exactly $40,000$ iterations (due to cholesky decomposition errors, memory errors, and TurBO not restarting after $190$ steps), we assume that the maximum achieved value was the best evaluation out to $200$ steps to mimic the performance that one would see if using the method in practice.
Early failures were only an issue for the exact GPs and then due to numerical instability, actually inspiring Figure \ref{fig:tree_depth}.
As this truncation wreaked a bit of havoc with our timings, we only report the first $150$ step timings in the main text ($170$ for Figure \ref{fig:app:svgp_inducing_ablation_rover}).

We used $500$ data points and the same model classes for the conditioning experiment (Figure \ref{fig:tree_depth}). 
Error bars are two standard deviations of the mean over the $10$ paths used.

For the global models experiment (Figure \ref{fig:rover_global}), we used a combination strategy that first used half the batch with Thompson sampling (TS) to select the points and then used qGIBBON \citep{moss2021gibbon} to select the other half of the batch by setting the TS half of the batch as pending points.
Performance using qGIBBON alone was about twice as slow and was slightly worse due to the lack of exploration that TS provides.
This strategy also enforces that qGIBBON's implementation actually uses fantasization and OVC, which would not natively have been the case without using the pending points.
For the timings, we report the first $106$ steps over $8$ seeds.

\textbf{MuJoCo: }
We use the codebase of \citet{wang2020learning} including their patched TurBO implementation with an optimization loop. 
This codebase is available from \url{https://github.com/facebookresearch/LaMCTS/tree/master/LA-MCTS} with a Creative Commons $4.0$ License with the included, modified TurBO implementation following a non-commercial license.
The MuJoCo experiments use mujoco-py (\url{https://github.com/openai/mujoco-py}, MIT License) and an institutional license key for MuJoCo itself \citep{todorov2012mujoco}.
These were run on a combination of Nvidia $32$GB V100s and $48$GB RTXes on an internal cluster.
We repeated these experiments over $10$ trials and computation took close to $14$ hours for hopper (where one trial failed to reach $4000$ samples for all methods but exact LTS), and several hours for swimmer.

On hopper, we struggled with wide variation in model fits, so we changed the regularization strategy on the base GP models to account for the high dimensional feature space.
Inspired by \citet{eriksson2021high}, we continued using ARD Matern-$5/2$ kernels but placed \texttt{HalfCauchy}($\tau$) priors on the inverse lengthscales and then placed a \texttt{HalfCauchy}($1.0$) prior on $\tau$ itself.
Rather than using MCMC as \citet{eriksson2021high} did, we used MAP to estimate both the lengthscales and $\tau.$

\subsection{Bayesian Optimization with the Knowledge Gradient}

\begin{figure*}[h!]
\centering
\begin{subfigure}{0.32\textwidth}
\centering
\includegraphics[width=\linewidth]{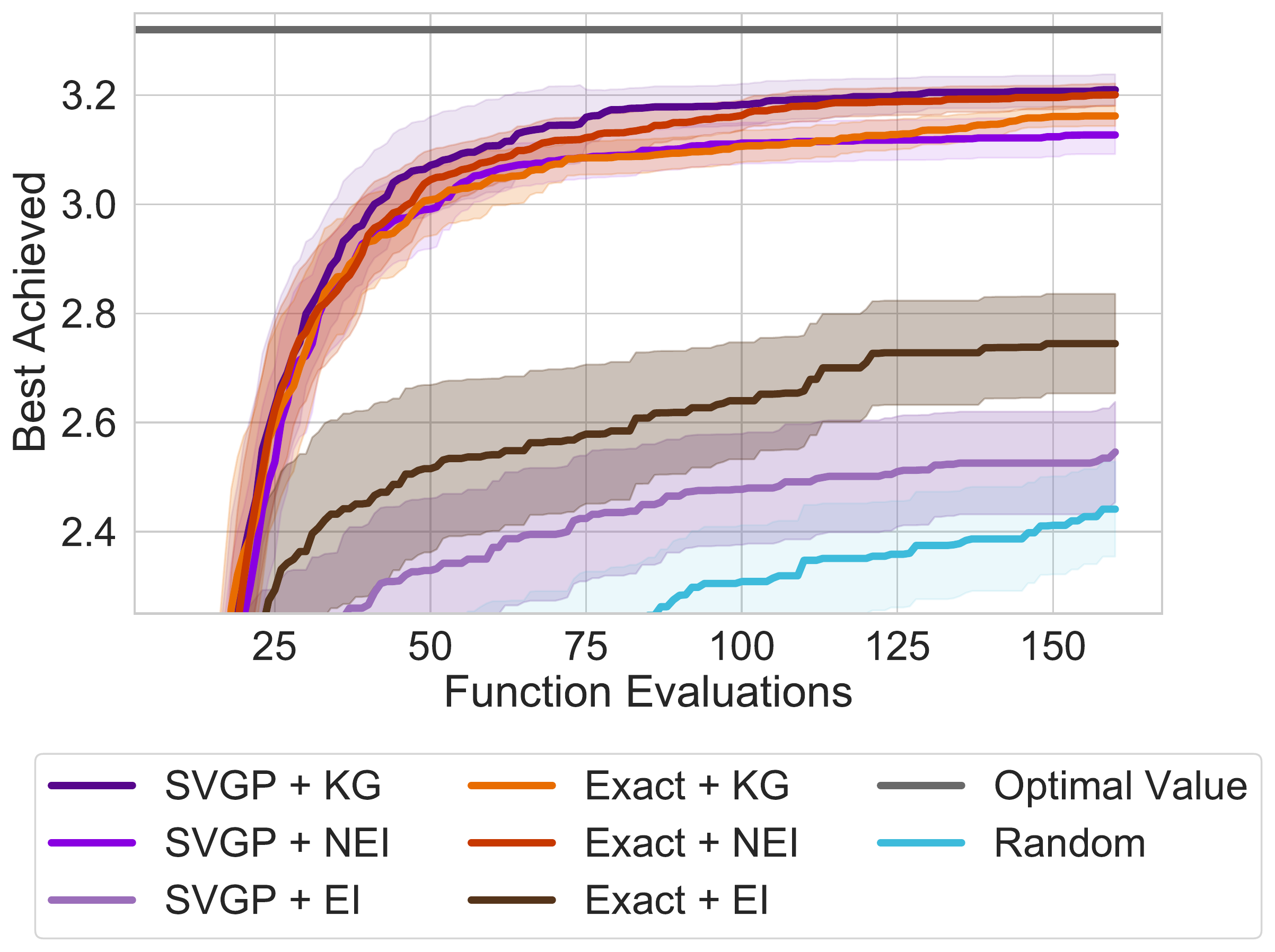}
\caption{Hartmann6.}
\label{fig:app:kg_hartmann}
\end{subfigure}
\begin{subfigure}{0.32\textwidth}
\centering
\includegraphics[width=\linewidth]{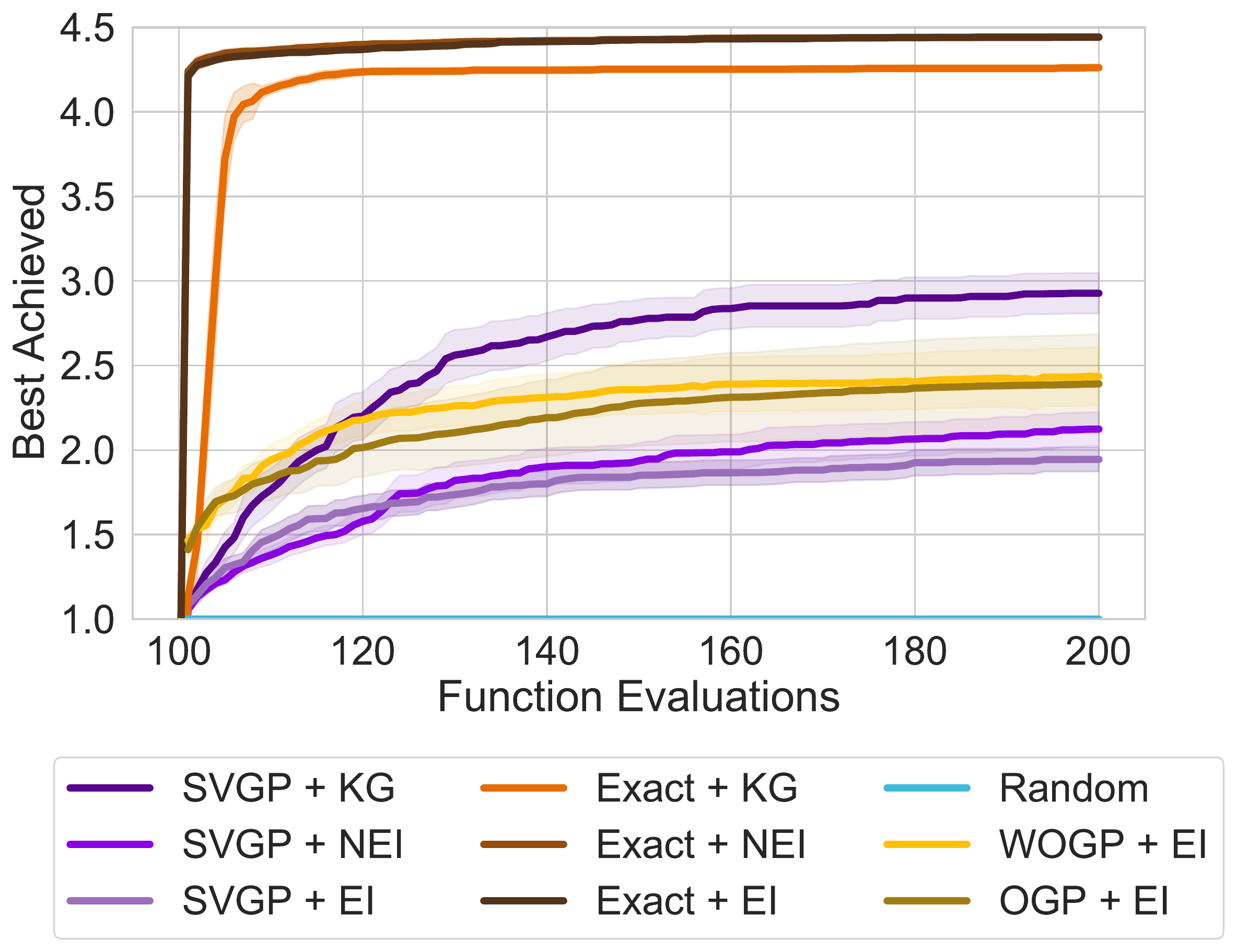}
\caption{Laser.}
\label{fig:app:kg_laser}
\end{subfigure}
\begin{subfigure}{0.32\textwidth}
\centering
\includegraphics[width=\linewidth]{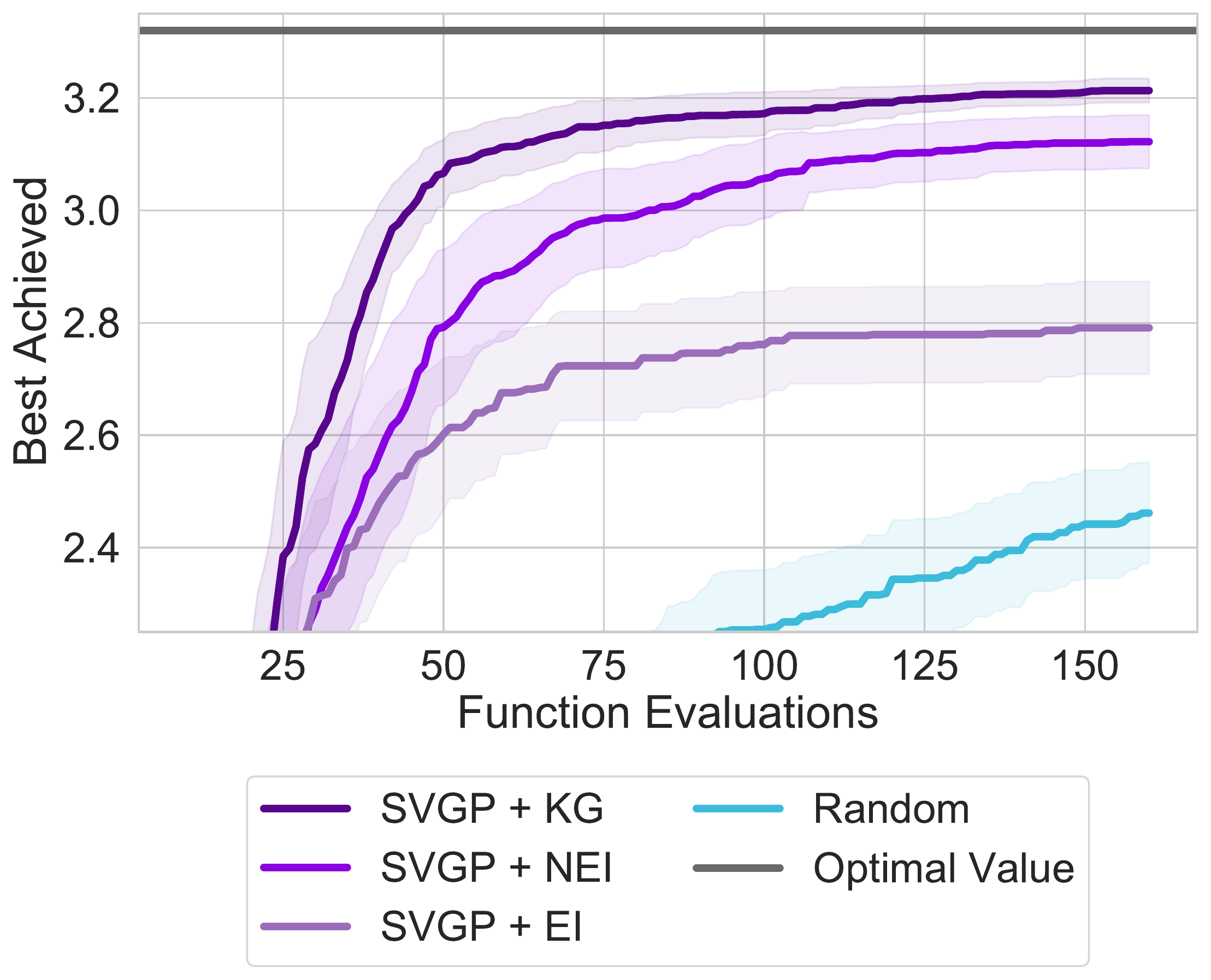}
\caption{Poisson-Hartmann6.}
\label{fig:app:kg_poisson}
\end{subfigure}
\caption{\textbf{(a)} Comparison to a broader suite of methods on Hartmann6, 1 constraint. \textbf{(b)} Comparison includng BoTorch exact GPs and their acquisitions on the free electron laser problem. Comparing to the results in Figure 3 of \citet{mcintire2016sparse}, these exact GPs vastly outperform their implementation, presumably due to advances in acquisition function optimization. \textbf{(c)} Hartmann6 test problem with count responses (Poisson likelihood). Only approximate inference can be used here, and qKG vastly outperforms qNEI. }
\label{fig:app:svgp_kb_bo_gaussian}
\end{figure*}

In Figure \ref{fig:app:svgp_kb_bo_gaussian}, we present the results for a wider set of acquisition functions using the one-shot knowledge gradient on three test functions.
These results complement Figure \ref{fig:svgp_kg_bo} and only include these additional methods.
Overall, qKG (with either an exact GP or a SVGP) generally performs best, followed by qNEI and then qEI.

In Figure \ref{fig:app:kg_hartmann}, we include results on the constrained Hartmann6 problem with random baselines as well as exact and SVGPs with expected improvement (EI). 
Exact and SVGPs with EI are signficantly outperformed by the other, more advanced acquisitions, but do outperform a random baseline.

\begin{wrapfigure}{l}{0.5\textwidth}
\centering
\centering
\caption{Best achieved values on Hartmann-$6$ for batch size, $q = 3$ for both NEI and KG for exact and SVGPs.}
\label{tab:hartmann3}
\small
\begin{tabular}{llll}
\toprule
\textbf{Noise Level} & \textbf{Acqf} & \textbf{Method} & \textbf{Best Value}  \\ \midrule
0.1         & NEI  & Exact  & 3.19 (0.05) \\ 
0.1         & NEI  & SVGP   & 3.08 (0.06) \\ 
0.1         & KG   & Exact  & 3.17 (0.03) \\ 
0.1         & KG   & SVGP   & 3.13 (0.04) \\ \hline
0.5         & NEI  & Exact  & 3.20 (0.03) \\ 
0.5         & NEI  & SVGP   & 3.12 (0.06) \\ 
0.5         & KG   & Exact  & 3.15 (0.03) \\ 
0.5         & KG   & SVGP   & 3.21 (0.05) \\ \bottomrule
\end{tabular}
\end{wrapfigure}

In Figure \ref{fig:app:kg_laser}, we also show the results of exact GPs as well as an online GP \citep[OGP,][]{csato2002sparse} with EI using the implementation of \citet{mcintire2016sparse}.
Interestingly, all of the exact GPs significantly outperform their variational counterparts.
This is quite surprising in some sense, as the true simulator is a weighted OGP with fixed hyper-parameters, and this method (WOGP + EI) performs much worse.
Note that the random baseline makes no progress.

Finally, in Figure \ref{fig:app:kg_poisson}, we display the results on constrained Hartmann-6 with Poisson observations, where each method outperforms random querying, but as expected SVGP + qEI, performs worse than qNEI and qKG.

\textbf{Batch Size and Noise Level Ablation:}
In Tables \ref{tab:hartmann3} and \ref{tab:hartmann1}, we display the final optimization results after $150$ function evaluations on the Hartmann-$6$ test problem for varying levels of noise and for each acquisition. 
These results are over $20$ trials and we display the mean maximum achieved value. 

Overall, KG tends to outperform NEI at both low and noise levels, with both exact and SVGP models performing very similarly overall with the SVGPs getting a slight edge in the high noise setting.
Furthermore, larger batch sizes tend to perform slightly better as the mean maximum achieved tends to have lower variation.
Finally, higher noise levels tend to be somewhat harder to optimize as expected. 

As we add more Gaussian noise into the function, we might a priori expect that qNEI should outperform qKG given the class of models. 
However, all things being held equal, a model that is more robust to the observed noise should tend to perform better, particularly if we are using more than just its mean and variance. 
Thus, SVGP models, by virtue of having more parameters to tune, tend to be more robust to the observed noise than the exact GPs.

\begin{wrapfigure}{l}{0.5\textwidth}
\caption{Best achieved values on Hartmann-$6$ for batch size, $q = 1$ for both NEI and KG for exact and SVGPs.}
\label{tab:hartmann1}
\small
\begin{tabular}{llll}
\toprule
\textbf{Noise Level} & \textbf{Acqf} & \textbf{Method} & \textbf{Best Value}  \\ \midrule
0.1                                        & NEI                                & Exact                                & 3.14 (0.13)                              \\ 
0.1                                        & NEI                                & SVGP                                 & 3.10 (0.10)                             \\ 
0.1                                        & KG                                 & Exact                                & 3.20 (0.02)                              \\ 
0.1                                        & KG                                 & SVGP                                 & 3.18 (0.03)                              \\ \hline
0.5                                        & NEI                                & Exact                                & 3.10 (0.10)                              \\ 
0.5                                        & NEI                                & SVGP                                 & 3.10 (0.14)                              \\ 
0.5                                        & KG                                 & Exact                                & 3.12 (0.04)                              \\ 
0.5                                        & KG                                 & SVGP                                 & 3.16 (0.05)                              \\ \bottomrule
\end{tabular}
\end{wrapfigure}

\subsection{Ablations on Rover}

\begin{figure*}[t!]
\centering
\captionsetup[subfigure]{justification=centering}
	\begin{subfigure}{0.23\textwidth}
		\centering
		\includegraphics[width=\linewidth]{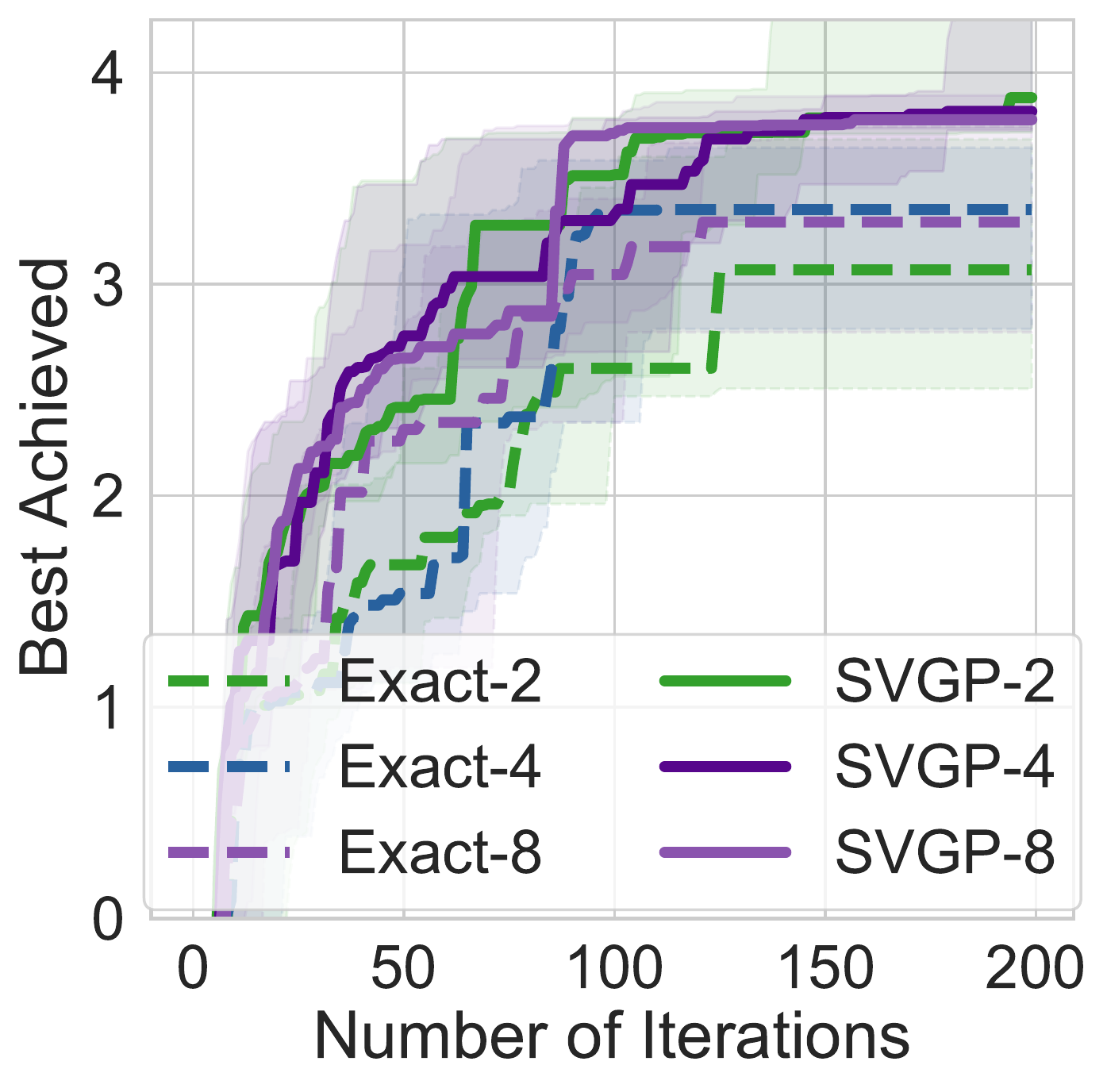}
		\caption{LTS tree depths.}
		\label{fig:tree_depth}
	\end{subfigure}
		\begin{subfigure}{0.23\textwidth}
		\centering
		\includegraphics[width=\linewidth]{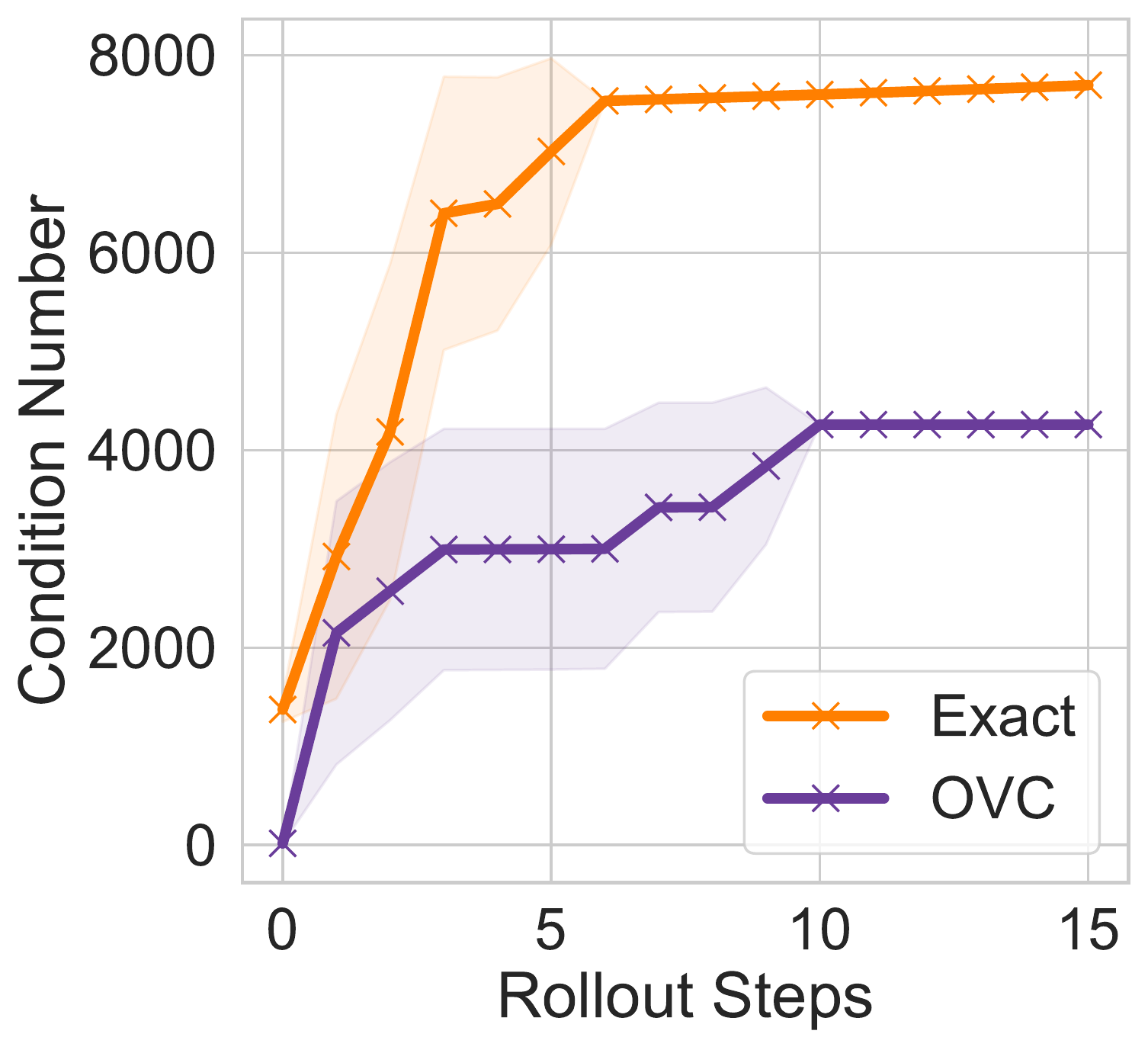}
		\caption{Conditioning.}
		\label{fig:trbo_conditioning}
	\end{subfigure}
\begin{subfigure}{0.23\textwidth}
\centering
\includegraphics[width=\linewidth]{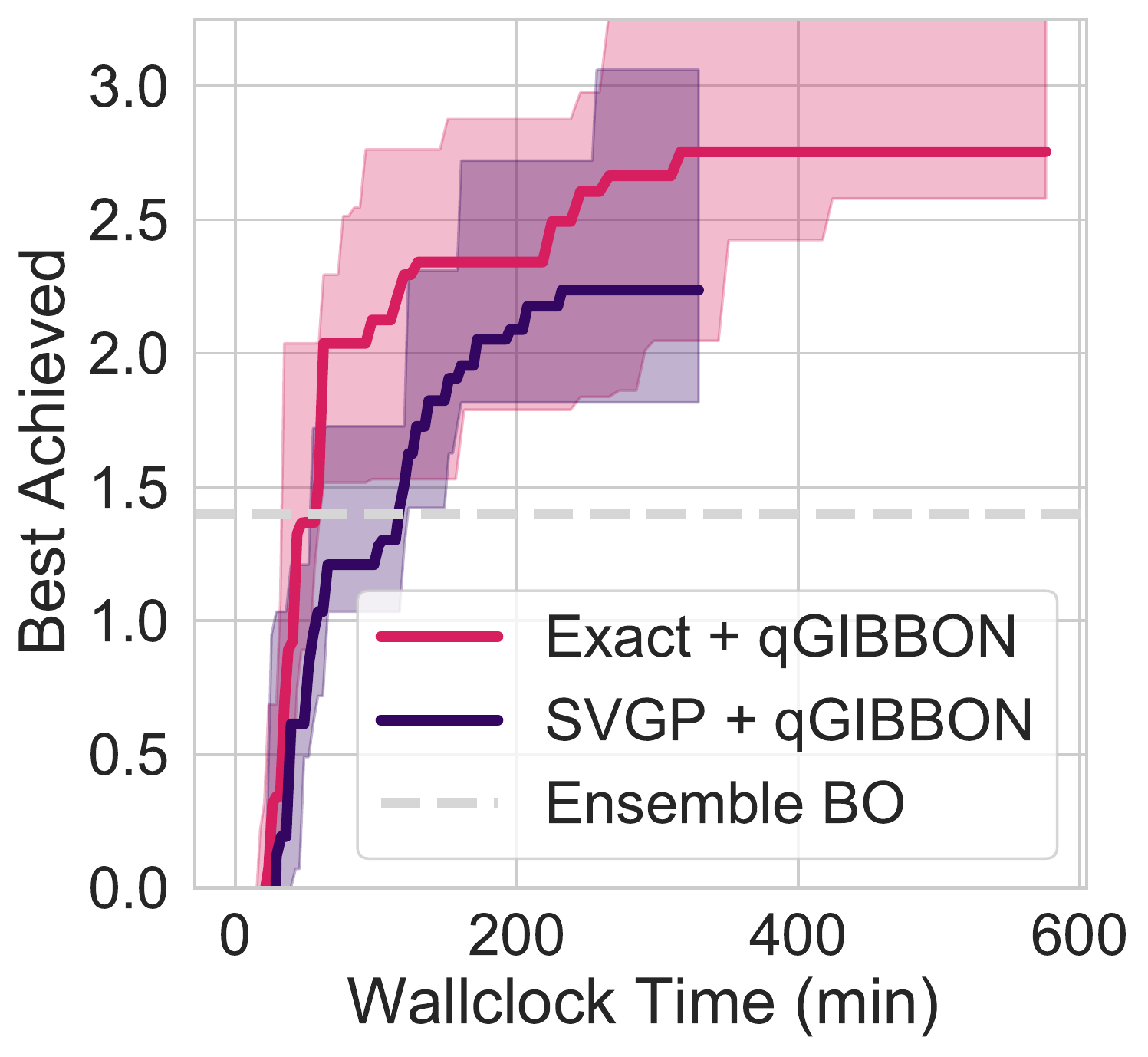}
\caption{Wall-Clock time.}
\label{fig:rover_global}
\end{subfigure}
\begin{subfigure}{0.23\textwidth}
\centering
\includegraphics[width=\linewidth]{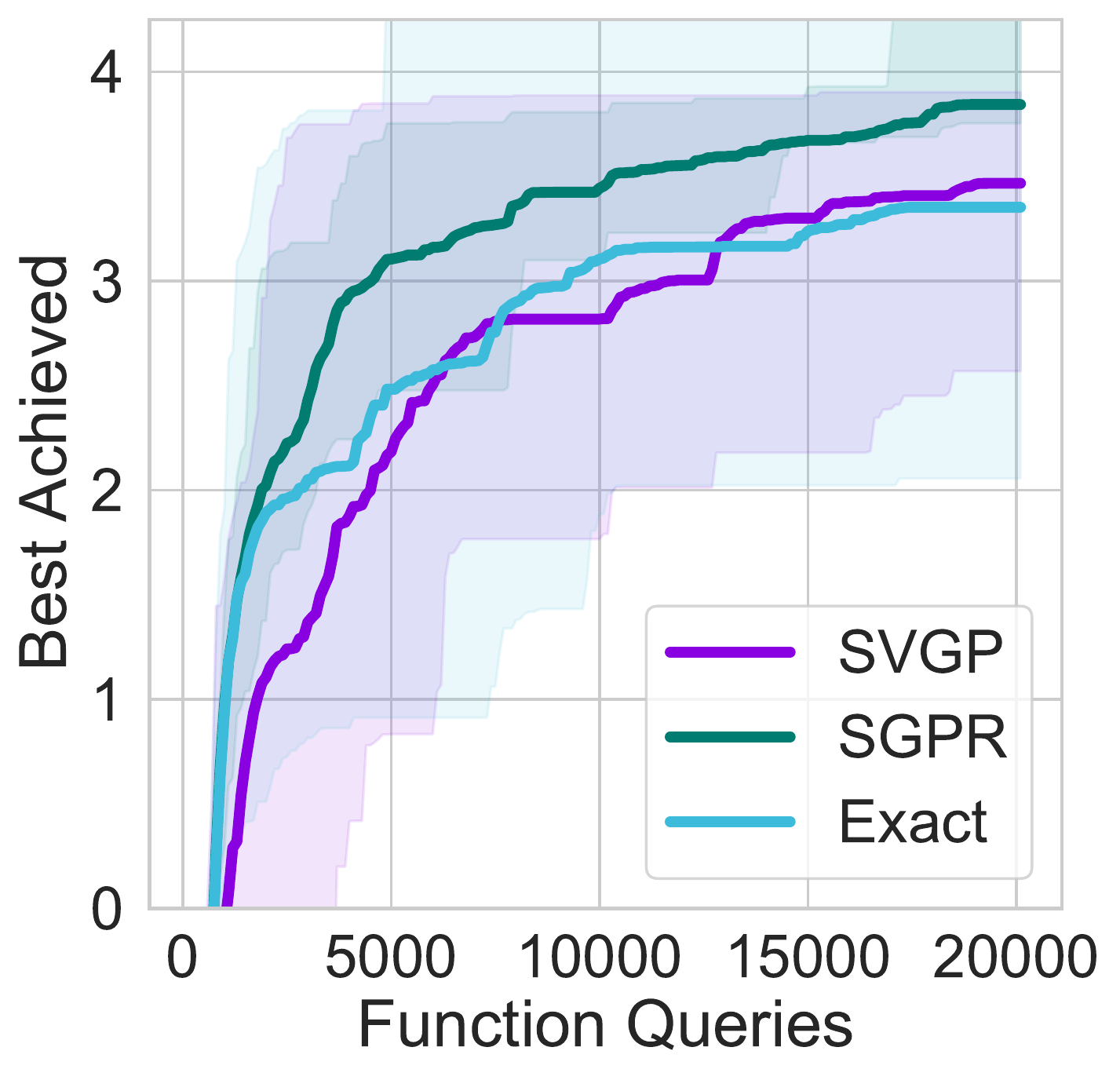}
\caption{SGPR ablation.}
\label{fig:trbo_60}
\end{subfigure}
\caption{
\textbf{(a)} Conditioning of exact and OVC conditioning across LTS tree depths on rover. \textbf{(b)} Performance of different rollout depths is similar, as the bigger gains are conditioning.
\textbf{(c)} Time efficiency using global models. SVGPs are still a strong baseline. \textbf{(d)} Comparison with SGPR using pivoted cholesky initializaiton on rover. SGPR is competitive. 
}
\label{fig:trbo_appendix}
\end{figure*}

\textbf{Effects of LTS:} To ablate the effects of rollouts and improved conditioning, we consider several step rollouts on the rover function as shown in Figure \ref{fig:tree_depth}.
We find that performance is similar across depths.
However, the conditioning of the resulting training data covariance is vastly improved when using OVC, as shown in Figure \ref{fig:trbo_conditioning}.
Taking these two results together, we see that in most cases, a tree depth of $4$ should be enough to gain the improvements from rollouts without increasing the conditioning of the system too much.

\textbf{Global models and SGPR: } To further demonstrate time efficiency of using OVC in the context of even global models, we perform large batch BO with the recently introduced qGIBBON acquisition (a max value entropy search variant) but using half the batch with TS to enforce fantasization over the TS-acquired batch \citep{moss2021gibbon}. 
This strategy is significantly slower than TurBO; however, even in this setting using SVGPs is twice as fast, and achieves a similar result to the exact model, as shown in Figure \ref{fig:rover_global}.
Both are orders of magnitude faster than Ensemble BO, which uses batch max value entropy search and exact GPs with addtive kernels \citep{wang2017max,wang2018batched}.
Ensemble BO takes at least several days of compute time \citep{eriksson2019scalable}.
Our result here compares very favorably to the (un-timed) results using SVGPs as well as exact GPs with ARD kernels that \citet{wang2018batched} also compared to, as neither of those methods reached reward values $\geq 1$ on this problem, even after $35,000$ steps.

\begin{figure}[h!]
\centering
\begin{subfigure}{0.23\textwidth}
\centering
\includegraphics[width=\linewidth]{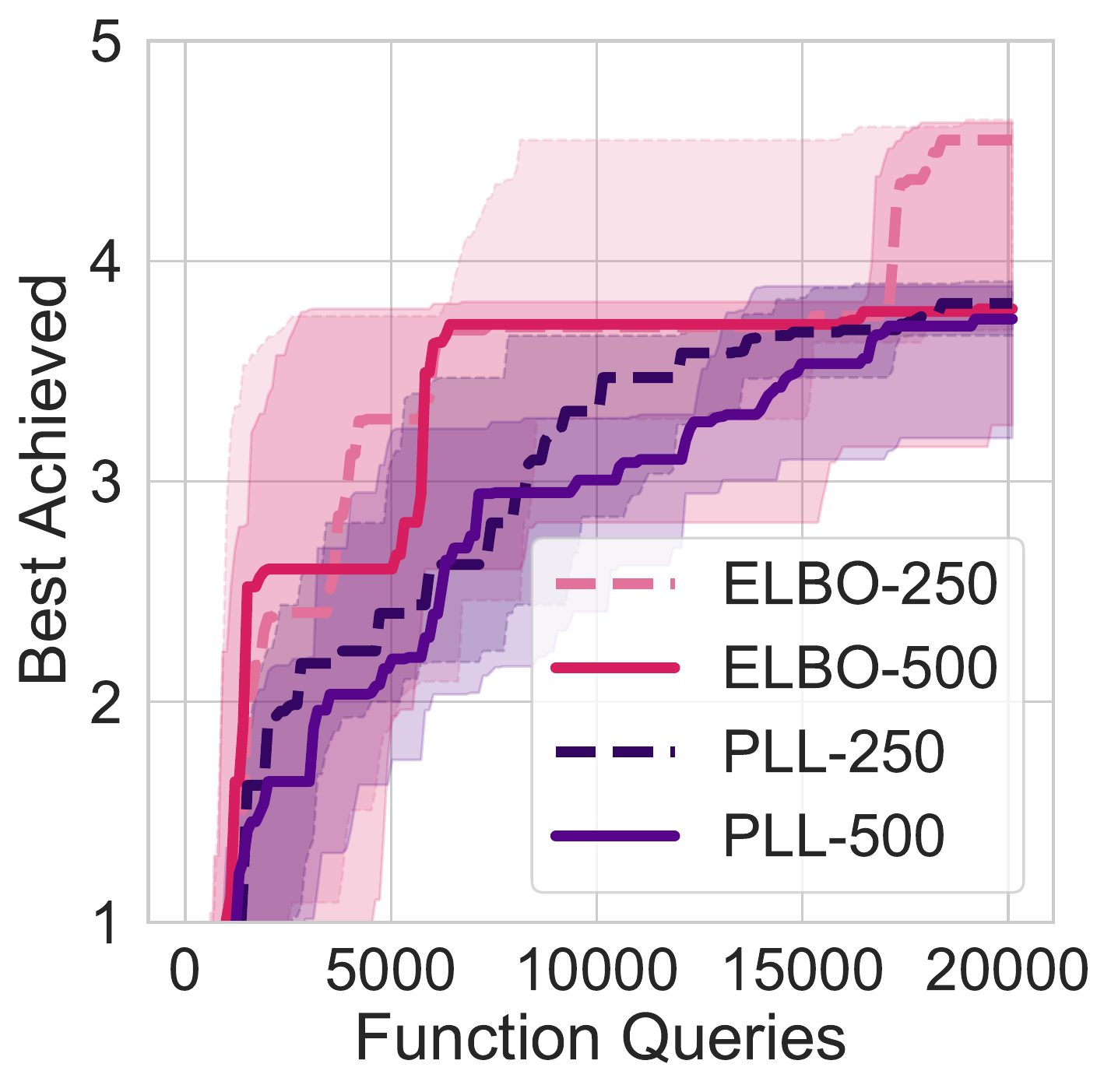}
\caption{TS data efficiency.}
\label{fig:app:rover_methods_data}
\end{subfigure}
\begin{subfigure}{0.23\textwidth}
\centering
\includegraphics[width=\linewidth]{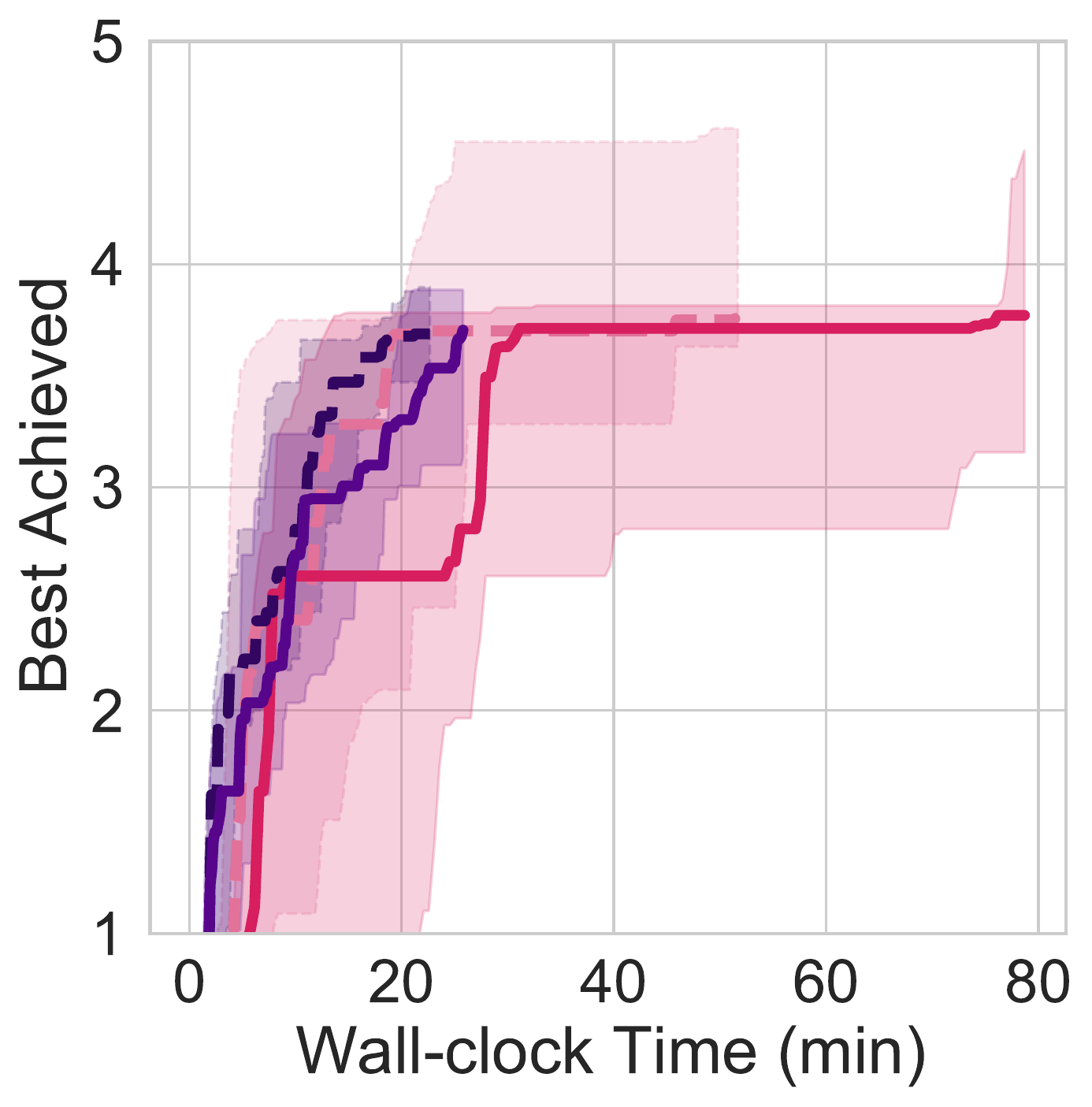}
\caption{TS time.}
\label{fig:app:rover_methods_time}
\end{subfigure}
\begin{subfigure}{0.23\textwidth}
\centering
\includegraphics[width=\linewidth]{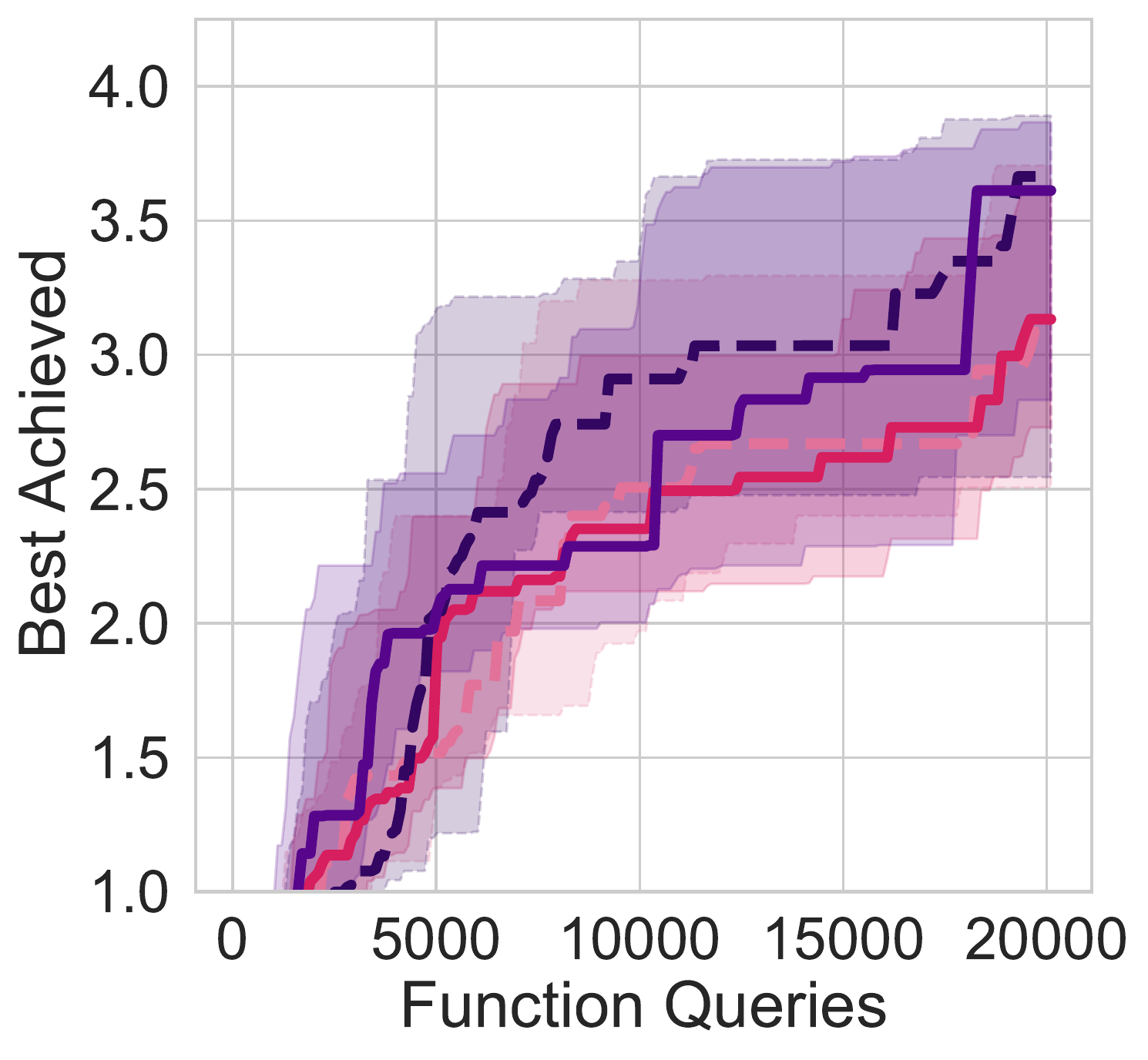}
\caption{LTS data efficiency.}
\label{fig:app:rover_methods_rollout_data}
\end{subfigure}
\begin{subfigure}{0.23\textwidth}
\centering
\includegraphics[width=\linewidth]{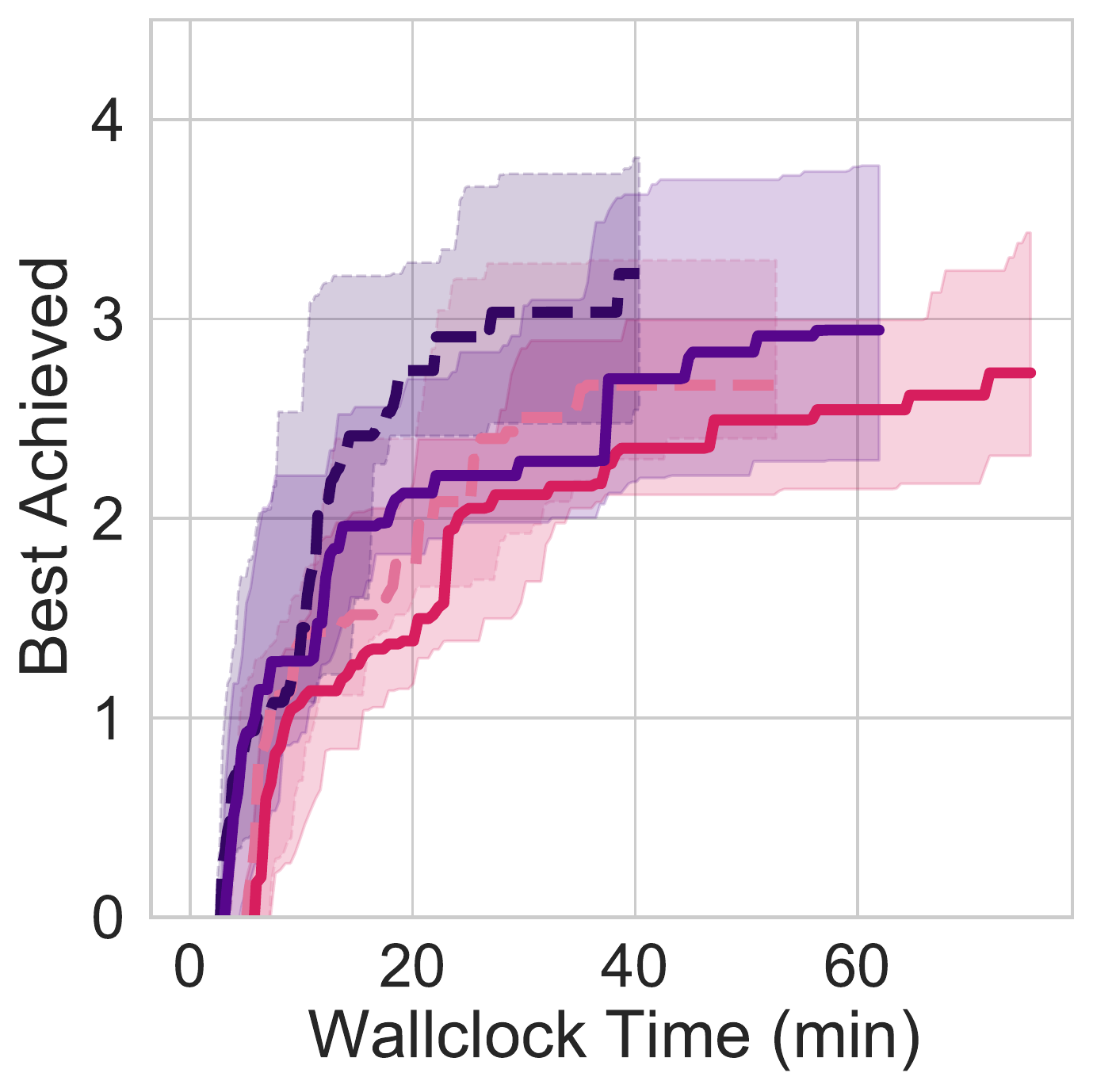}
\caption{LTS time.}
\label{fig:app:rover_methods_rollout_time}
\end{subfigure}
\caption{Inducing point ablation on rover \textbf{(a,b)} use Thompson sampling and \textbf{(c,d)} use rollouts.}
\label{fig:app:svgp_inducing_ablation_rover}
\end{figure}

\textbf{Number of Inducing Points and Training Loss: } Finally, in Figure \ref{fig:app:svgp_inducing_ablation_rover} we ablate between training SVGPs with either $250$ or $500$ inducing points as well as the training loss, either the evidence lower bound (ELBO) or the predictive log likelihood (PLL), (see Appendix \ref{app:training_losses} for further descriptions) on the $d=60$ rover problem. 
We find that there is not a significant amount of difference between any of the four approaches whether using the ELBO or the PLL. 
In general, the PLL approaches are somewhat more quick to train (Figures \ref{fig:app:rover_methods_time} and \ref{fig:app:rover_methods_time}) in comparison to the ELBO models.
They also tend to slightly outperform the ELBO-trained SVGPs when using LTS (Figure \ref{fig:app:rover_methods_rollout_data}) in terms of function efficiency, but perform similarly for standard Thompson sampling (Figure \ref{fig:app:rover_methods_data}).
We leave a detailed benchmarking of these methods in the context of downstream tasks for future work.
\end{document}